\pdfoutput=1
\documentclass[10pt,twocolumn,letterpaper]{article}
\usepackage[pagenumbers]{cvpr}
\usepackage{cuted}      %
\usepackage{caption}    %
\usepackage{iftex}
\ifPDFTeX\usepackage[T1]{fontenc}\fi
\usepackage{booktabs}
\usepackage{multirow}
\usepackage{amsmath}
\usepackage{amssymb}
\usepackage{pifont}
\usepackage{wasysym}    %
\usepackage{xspace}
\usepackage{microtype}

\newcommand{\bench}{GridProbe\xspace}
\definecolor{okc}{HTML}{1A9E6E}    %
\definecolor{noc}{HTML}{D8383A}    %
\definecolor{parc}{HTML}{D99400}   %
\newcommand{\cmark}{\textcolor{okc}{\ding{51}}}
\newcommand{\xmark}{\textcolor{noc}{\ding{55}}}

\newcommand{\best}[1]{\textbf{#1}}          %
\newcommand{\second}[1]{\underline{#1}}     %
\newcommand{\tablemarks}{\textbf{Bold}: best in each group; \underline{underlined}: second-best open-source model.}
\definecolor{imgc}{HTML}{C9501C}   %
\definecolor{txtc}{HTML}{1F62B5}   %
\newcommand{\Img}{\textcolor{imgc}{Image}}
\newcommand{\Txt}{\textcolor{txtc}{Text}}
\newcommand{\readset}{Read\xspace}
\newcommand{\ruleset}{Rule\xspace}
\newcommand{\solveset}{Solve\xspace}
\newcommand{\qT}{Qwen3-VL-8B-Thinking\xspace}
\newcommand{\qI}{Qwen3-VL-8B-Instruct\xspace}
\newcommand{\qL}{Qwen3-VL-32B-Instruct\xspace}
\newcommand{\para}[1]{\vspace{1pt}\noindent\textbf{#1}}
\newcommand{\dvis}{\Delta_{\text{vis}}}

\usepackage{placeins}
\usepackage[most]{tcolorbox}
\definecolor{cvprblue}{rgb}{0.21,0.49,0.74}
\usepackage[pagebackref,breaklinks,colorlinks,allcolors=cvprblue]{hyperref}
\usepackage[capitalize]{cleveref}
\hypersetup{
  pdftitle={Lost in the Grid: Where Multimodal LLMs Fail Between Seeing and Planning},
  pdfauthor={Luoxi Zhang, Tianyi Tan, Jiawei Liu, Xingyu Zeng, Min Yang, Shiwen Ni}
}
\def\confName{arXiv}
\def\confYear{2026}

\definecolor{promptbg}{HTML}{FAF8F2}
\definecolor{promptfr}{HTML}{8A8F9C}
\newtcolorbox{promptbox}[2][]{enhanced, colback=promptbg, colframe=promptfr, boxrule=0.45pt, arc=1.4pt,
  left=3.5pt, right=3.5pt, top=2.5pt, bottom=2.5pt, boxsep=0pt, before skip=3.5pt, after skip=3.5pt,
  fonttitle=\sffamily\bfseries\scriptsize, coltitle=black, colbacktitle=black!9, title={#2},
  toptitle=1.6pt, bottomtitle=1.6pt, fontupper=\ttfamily\fontsize{6.3}{7.5}\selectfont\raggedright\frenchspacing, #1}
\newcommand{\pimg}[1]{\textit{#1}}
\newcommand{\ptxt}[1]{\textit{#1}}
\newcommand{\vimg}[1]{\slot{image}~#1}
\newcommand{\vtxt}[1]{\slot{text}~#1}
\newcommand{\gup}[1]{\textcolor{noc}{#1}}

\newcommand{\fence}{{\char96}{\char96}{\char96}}
\newcommand{\slot}[1]{{\normalfont\sffamily\scriptsize\itshape$\langle$#1$\rangle$}}
\newsavebox{\fitbox}
\newcommand{\fittab}[1]{\sbox{\fitbox}{#1}\ifdim\wd\fitbox>\linewidth\resizebox{\linewidth}{!}{\usebox{\fitbox}}\else\usebox{\fitbox}\fi}

\title{Lost in the Grid: Where Multimodal LLMs Fail\\ Between Seeing and Planning}
\author{
Luoxi Zhang$^{1,*}$\quad
Tianyi Tan$^{2,*}$\quad
Jiawei Liu$^{1,*}$\quad
Xingyu Zeng$^{1}$\quad
Min Yang$^{3}$\quad
Shiwen Ni$^{1,\dagger}$\\[0.2em]
{\small
$^{1}$Shenzhen University of Advanced Technology\quad
$^{2}$Dalian University of Technology\quad
$^{3}$Shenzhen Institutes of Advanced Technology}\\
{\small $^{*}$Equal contribution\quad $^{\dagger}$Corresponding author}
}

\begin{document}
\maketitle
\begin{abstract}
Multimodal large language models (MLLMs) excel at static visual understanding and are increasingly applied to spatial planning, where a model interprets a visual layout and outputs an action sequence to reach a goal.
Grid games are a common testbed for this ability because their rules are explicit and a solver can verify every move.
Yet existing benchmarks mostly report end-to-end success rates, which hide whether a failure stems from visual perception, rule application, or multi-step planning.
We introduce \bench, a diagnostic benchmark that 1) locates the failing stage between visual perception and planning and 2) verifies the diagnosis by targeted fine-tuning.
On Sokoban, \bench pairs end-to-end solving with four probes (recognition, localization, single-step prediction, and state tracking) and gives every board both as an image and as an equivalent text grid.
Evaluating seven MLLMs on Sokoban and a maze game reveals distinct bottlenecks behind similar success rates: proprietary models are limited by multi-step planning, whereas open-source models are limited by rule application or by visual grid indexing, \ie, counting rows and columns.
Planning failures also hide perception failures: when planning is easier, the visual cost of the same model grows from 7.6 to 37.7 points.
The diagnosis identifies what to train: fine-tuning only the vision module of the perception-limited model on board reading raises its interactive Sokoban success 3.5-fold (5.8\% to 20.4\%) and transfers zero-shot to the unseen maze game (single-turn success 13.8\% to 67.1\%), whereas the same data does not help a rule-limited model and training on solver solutions brings no gain.
We will release the benchmark and code at \url{https://github.com/suat-llm/GridProbe}.
\end{abstract}

\section{Introduction}
\label{sec:intro}

Multimodal large language models (MLLMs) have advanced considerably on static image understanding tasks such as visual question answering and image captioning~\cite{qwen3vl,internvl35}.
As these models are increasingly used in interactive settings such as robot manipulation and user-interface automation~\cite{openvla,uitars}, they must go beyond describing a scene and perform spatial planning, where a model interprets a visual layout and generates an action sequence to reach a goal~\cite{vsp2025,viplan2025,ivispar2025}.
In such sequential decisions, a small error in perception or rule application can compound over later steps, so knowing where spatial planning breaks down is important for building reliable multimodal agents.
Grid games are a common testbed for this capability~\cite{balrog,lmgamebench,vgrpbench2025,seeingwithout2026}: their rules are concise and explicit, state transitions are deterministic, and an algorithmic solver yields the exact solution for every board, so every intermediate step of a model can be checked, not only its final outcome.

\begin{figure}[t]
\centering
\includegraphics[width=\linewidth]{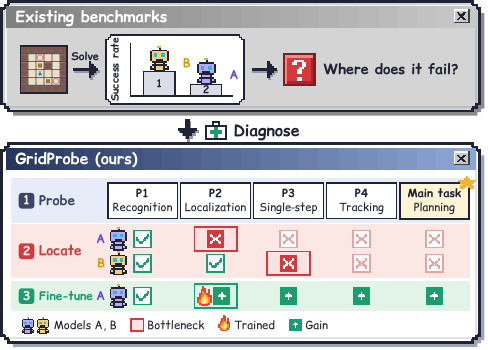}
\caption{\textbf{Where does a model fail?} Existing benchmarks rank models by success rate. \bench probes every stage on the same boards, locates each model's bottleneck (A: localization; B: single-step prediction), and fine-tunes it; training A's bottleneck improves every later stage, including the main task (star).}
\label{fig:teaser}
\end{figure}

Yet current MLLMs struggle in these environments.
In the Sokoban grid game, a player pushes boxes onto target locations, and a single mistaken push can render a level unsolvable~\cite{culberson,sokobench2026}.
Indeed, the open-source models we evaluate solve almost no Sokoban levels from visual inputs.
Such failures can stem from distinct causes: incorrect object recognition, inaccurate coordinate localization, erroneous single-step rule prediction, failed multi-step state tracking, or insufficient planning.
Each cause requires a different remedy, yet a single success rate cannot tell them apart.

\begin{table}[t]
\centering
\footnotesize
\setlength{\tabcolsep}{2.2pt}
\newcommand{\hd}[1]{\scriptsize\shortstack[c]{#1}}
\begin{tabular}{@{}l ccccccc@{}}
\toprule
 & \multicolumn{2}{c}{\scriptsize Solving} & \multicolumn{2}{c}{\scriptsize Probes} & & \multicolumn{2}{c}{\scriptsize Fine-tuning} \\
\cmidrule(lr){2-3}\cmidrule(lr){4-5}\cmidrule(l){7-8}
Benchmark & \hd{Main\\task} & \hd{Inter-\\active} & \hd{Perception} & \hd{Rules} & \hd{Paired\\inputs} & \hd{Used} & \hd{Tests\\diagnosis} \\
\midrule
BALROG~\cite{balrog} & \cmark & \cmark & \xmark & \xmark & \cmark & \xmark & \xmark \\
lmgame-Bench~\cite{lmgamebench} & \cmark & \cmark & \xmark & \xmark & \cmark & \cmark & \xmark \\
iVISPAR~\cite{ivispar2025} & \cmark & \cmark & \xmark & \xmark & \cmark & \xmark & \xmark \\
VGRP-Bench~\cite{vgrpbench2025} & \cmark & \xmark & \cmark & \cmark & \cmark & \cmark & \xmark \\
VSP~\cite{vsp2025} & \cmark & \xmark & \cmark & \cmark & \cmark & \cmark & \xmark \\
Jin et al.~\cite{seeingwithout2026} & \xmark & \xmark & \cmark & \cmark & \cmark & \xmark & \xmark \\
\midrule
\textbf{\bench (ours)} & \cmark & \cmark & \cmark & \cmark & \cmark & \cmark & \cmark \\
\bottomrule
\end{tabular}
\caption{\textbf{Comparison with related benchmarks.} \emph{Paired inputs}: tasks given as images and as text. \emph{Tests diagnosis}: fine-tuning verifies a diagnosed bottleneck rather than only raising scores.}
\label{tab:compare}
\end{table}

Existing benchmarks address only isolated facets of this problem (\cref{tab:compare}).
Game-based benchmarks evaluate end-to-end play through overall success rates or task progress~\cite{balrog,lmgamebench,videogamebench,korgym}, whereas perception-focused benchmarks expose basic visual errors without tracing how they propagate into downstream planning~\cite{vlmsareblind,blink}.
Other studies compare image and text inputs on single tasks~\cite{isobench,spatialeval,mazebench2026}, or add sub-tasks that evaluate perception and reasoning separately~\cite{vsp2025,vgrpbench2025,seeingwithout2026}.
However, existing evaluations rarely trace a failure on the same board back to its originating stage, and they typically fine-tune on the benchmark tasks themselves, which does not show whether fixing a diagnosed bottleneck improves downstream planning.
Two research questions therefore remain open: 1)~[Diagnosis] At which stage between visual perception and planning does each model fail? 2)~[Validation] Does fine-tuning the diagnosed bottleneck improve subsequent stages?

To answer these questions, we introduce \bench, a diagnostic benchmark for MLLMs in grid games (\cref{fig:teaser}).
On Sokoban, \bench pairs the main solving task, evaluated both in a single turn (the full plan in one response) and interactively (a few moves per turn with board feedback), with four progressive probe tasks: recognition, localization, single-step prediction, and state tracking.
Every task is also given as a symbolic text grid on the same boards.
Comparing text and image performance along this chain shows where visual errors first emerge and how severe they are, enabling stage-wise error attribution (\cref{sec:metrics}).
We further apply the same probes to a maze grid game to test generalization.

Evaluating seven MLLMs (two proprietary and five open-source) reveals distinct bottlenecks in spatial planning.
The proprietary models read boards and apply rules accurately with both inputs, leaving multi-step planning as their primary limitation.
The open-source models all solve almost no levels from images, but for different reasons.
\qT is bottlenecked by visual perception: its localization accuracy drops from 83.3\% on text to 23.3\% on images, and adding coordinate labels raises it to 75.4\%, pointing to grid coordinate indexing as its main visual deficit.
In contrast, the remaining open-source models perform near chance on single-step prediction even from text, a bottleneck in rule application: three of them apply the rules correctly when told what the two cells ahead of the player contain (89--98\%), yet fail to apply them to a board.
The same patterns hold on the maze, where easier planning exposes what Sokoban success rates hide: the visual cost of \qT rises from 7.6 to 37.7 points.

Targeted fine-tuning confirms this diagnosis.
Fine-tuning only the vision module of \qT on board-reading data raises image localization to 90.8\%, boosts interactive Sokoban solving from 5.8\% to 20.4\%, and transfers zero-shot to the unseen maze grid game (single-turn success from 13.8\% to 67.1\%).
Conversely, the same data does not help a model limited by rule application, and training on solver solutions does not improve solving.

By combining stage-wise diagnosis with targeted fine-tuning, \bench offers a systematic framework for diagnosing and improving spatial planning in MLLMs.
Our main contributions are threefold:
\begin{itemize}[leftmargin=*,itemsep=1pt,topsep=2pt]
\item \textbf{Diagnostic Benchmark.} We introduce \bench, which pairs a main solving task with four progressive probe tasks and symbolic text inputs on the same boards to trace downstream planning failures back to earlier stages.
\item \textbf{Stage-Wise Diagnosis.} Evaluating seven MLLMs identifies three distinct bottlenecks (visual perception, rule application, and multi-step planning), shows that grid coordinate indexing is the main visual deficit of the model limited by visual perception, and shows that planning failures hide this deficit in end-to-end success rates.
\item \textbf{Diagnosis-Guided Fine-Tuning.} Comparing on-target, off-target, and direct main-task training, we show that \bench pinpoints actionable training targets: training the diagnosed stage of the perception-limited model improves every later stage, including the main task, whereas training on solver solutions does not.
\end{itemize}

\begin{figure*}[t]
\centering
\includegraphics[width=\linewidth]{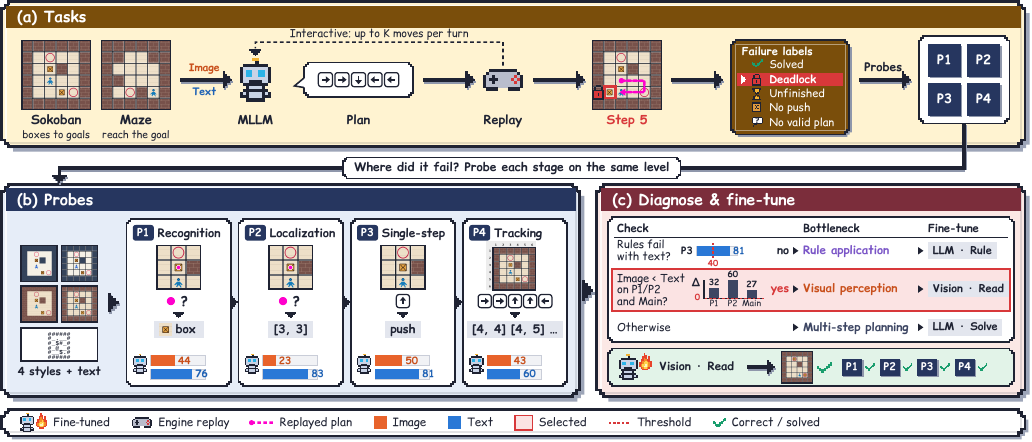}
\caption{\textbf{Overview of \bench.} (a) \emph{Tasks}: an MLLM plans a level in one response (single-turn) or up to $K$ moves per turn; the engine replays it and labels the failure. (b) \emph{Probes}: four probes test each stage on this level (four styles or text; bars: Qwen3-VL-8B-Thinking). (c) \emph{Diagnose and fine-tune}: the checks (\cref{sec:metrics}) pick what to train for this model, improving all later stages (\cref{tab:ft}; solved level illustrative).}
\label{fig:overview}
\end{figure*}

\section{Related Work}
\label{sec:related}

\subsection{Game and Puzzle Benchmarks for MLLMs}
Games are widely used to evaluate MLLMs as interactive agents.
BALROG~\cite{balrog}, lmgame-Bench~\cite{lmgamebench}, VideoGameBench~\cite{videogamebench}, KORGym~\cite{korgym}, and V-MAGE~\cite{vmage} place models in text or video games and report overall success rates or task progress.
Grid and puzzle benchmarks further leverage explicit rules and exact solutions, spanning grid puzzles (VGRP-Bench~\cite{vgrpbench2025}), sliding tiles (iVISPAR~\cite{ivispar2025}), multi-step assembly (LEGO-Puzzles~\cite{legopuzzles2025}), and symbolic visual planning (ViPlan~\cite{viplan2025}).
Specifically on Sokoban, lmgame-Bench notes that models struggle to parse visual boards, whereas SokoBench~\cite{sokobench2026} examines long-horizon planning from text-only inputs.
While these benchmarks measure end-to-end gameplay performance, \bench pairs a solvable main task with progressive diagnostic probes to trace each failure back to its underlying stage.

\subsection{Diagnosing Perception and Reasoning in MLLMs}
Perception benchmarks such as BLINK~\cite{blink}, \emph{VLMs Are Blind}~\cite{vlmsareblind}, and MMVP~\cite{eyeswideshut} reveal that MLLMs often struggle with basic visual properties, including object counting and relative spatial relations.
Meanwhile, IsoBench~\cite{isobench} and SpatialEval~\cite{spatialeval} present identical problems across visual and textual modalities, showing that models often perform better on text.
Closer to our work, VSP~\cite{vsp2025} decomposes spatial planning into perception and reasoning sub-tasks, VGRP-Bench~\cite{vgrpbench2025} pairs puzzle solving with cell-level perception and step-level rule following, and Jin et al.~\cite{seeingwithout2026} separately evaluate board perception and rule following in chess and Gomoku.
However, these studies primarily characterize aggregate capability gaps across models.
In contrast, \bench tracks failures on the exact same boards across modalities and sequential stages, isolating the primary bottleneck for each individual model.

\subsection{Fine-Tuning MLLMs for Spatial Perception and Planning}
Recent studies enhance spatial capabilities through targeted post-training.
SpatialVLM~\cite{spatialvlm} fine-tunes on large-scale synthetic spatial QA data, G1~\cite{g1vlmgym2025} and Game-RL~\cite{gamerl2026} apply reinforcement learning in game environments, and lmgame-Bench~\cite{lmgamebench} trains on Sokoban and Tetris and reports cross-task gains.
Similarly, VSP and VGRP-Bench fine-tune directly on their benchmark tasks to improve overall scores.
Notably, Wu et al.~\cite{wu2026asymmetric} find that post-training benefits reasoning more than perception, suggesting that perceptual deficits may need dedicated visual supervision.
Prior work fine-tunes to raise benchmark scores; we fine-tune to test a diagnosis: training a model's diagnosed bottleneck should improve its later stages but not help a model limited elsewhere.

\section{\bench Benchmark}
\label{sec:bench}

\subsection{Benchmark Overview}
\bench has two objectives: (1) to evaluate each stage from visual perception to planning and locate the failing stage, and (2) to validate this diagnosis by testing whether training that stage improves later stages.
Both require exact answers at every intermediate step, a way to separate image-reading errors from reasoning errors, and an automated protocol to compare models before and after training.

The Sokoban grid game meets these requirements.
A player pushes boxes onto goals, and one wrong push can make a level unsolvable, so solving requires looking ahead (\cref{fig:overview}).
Its explicit rules and deterministic transitions let a solver answer every question exactly.
To separate image-reading errors from reasoning errors, every board is given both as an image and as a symbolic text grid.
The two inputs carry the same information and use identical prompts except for the sentence describing the input format, so the gap between them measures the cost of reading the board from pixels.
Each task thus has three reference points: the solver upper bound, text-input accuracy, and image-input accuracy.
To test whether the diagnosis generalizes beyond one game, we also build a maze grid game with the same sprites and task formats but without boxes (\cref{sec:maze}).

\subsection{Diagnostic Probes and Main Task}
\label{sec:probes}
Solving a Sokoban level requires a chain of abilities: reading what each cell contains and where it is, predicting how one move changes the board, keeping track of the state over many moves, and searching for a plan.
\bench tests each link of this chain with one probe on the same boards as the main task, so that a main-task failure can be traced back to the abilities it depends on.
\begin{itemize}[leftmargin=*,itemsep=1pt,topsep=2pt]
\item \textbf{P1. Recognition.} Name the object in a marked cell. This tests basic visual object recognition without game rules.
\item \textbf{P2. Localization.} Report the \texttt{[row, column]} coordinate of a marked cell (evaluated with and without border coordinate rulers). This tests spatial grid indexing.
\item \textbf{P3. Single-Step Prediction.} Given a board and one action, predict the player's start position, end position, and outcome type (walk, push, or three blocked-move types). This tests whether the model applies the move rules.
\item \textbf{P4. State Tracking.} Given a board and $H \in \{5, 10, 15, 20\}$ moves, report the player's position after each step (shown with coordinate rulers, so that P4 isolates tracking from the indexing that P2 measures). This tests long-horizon state updating.
\item \textbf{Main Task: Spatial Planning.} Push all boxes onto goals within a move budget, in two settings: \emph{single-turn}, where the model outputs the full plan in a single response, and \emph{interactive}, where it outputs up to $K$ moves per turn and then receives the updated board. This tests end-to-end search and planning.
\end{itemize}
Visual styles, move budgets, and termination rules are detailed in \cref{sec:setup} and Supp.~A.

\subsection{Benchmark Creation and Quality Control}
We generate all levels with our own engine, which controls board size and difficulty, computes every answer exactly, and reduces the risk of training data contamination.

\para{Level synthesis.} Sokoban levels are built by pulling boxes backward from a solved state, which guarantees that every level is solvable.
A breadth-first solver then finds a push-optimal reference solution, whose length sets the move budget.
Maze boards are sampled with random walls and kept only if the exit is reachable.

\para{Balancing and verification.} Items are balanced over board size, number of boxes, and outcome type, and duplicate layouts are removed.
A self-test confirms that the scorer gives 100\% to the reference answers, so every lost point is a model error.

\begin{table*}[t]
\centering
\footnotesize
\setlength{\tabcolsep}{3.1pt}
\renewcommand{\arraystretch}{0.94}
\newcommand{\grp}[1]{\multicolumn{15}{@{}l}{\textit{#1}}}
\begin{tabular}{@{}l cc ccc cc cc c cc cc c l@{}}
\toprule
 & \multicolumn{9}{c}{Stage-wise probes} & & \multicolumn{4}{c}{Main task} & & \\
\cmidrule(lr){2-10}\cmidrule(lr){12-15}
 & \multicolumn{2}{c}{P1 Recognition} & \multicolumn{3}{c}{P2 Localization} & \multicolumn{2}{c}{P3 Single-step} & \multicolumn{2}{c}{P4 Tracking} & & \multicolumn{2}{c}{Single-turn} & \multicolumn{2}{c}{Interactive} & & \\
\cmidrule(lr){2-3}\cmidrule(lr){4-6}\cmidrule(lr){7-8}\cmidrule(lr){9-10}\cmidrule(lr){12-13}\cmidrule(lr){14-15}
Model & \Img & \Txt & \Img & \textcolor{imgc}{+ ruler} & \Txt & \Img & \Txt & \Img & \Txt & & \Img & \Txt & \Img & \Txt & & Bottleneck \\
\midrule
\grp{Sokoban} \\
Random guess & 20 & 20 & $<$3 & $<$3 & $<$3 & 20 & 20 & $<$3 & $<$3 & & 0.5 & 0.5 & 0.5 & 0.5 & & -- \\
GPT-6.1 & 94.3$^{*}$ & 98.3 & 89.7$^{*}$ & 97.9$^{*}$ & 99.2 & 99.4$^{\dagger}$ & 98.2$^{\dagger}$ & 87.9$^{\dagger}$ & 86.7$^{\dagger}$ & & 61.8 & 66.2 & 81.3 & \best{85.3} & & Planning \\
Gemini 3.8 Flash & \best{100.0}$^{*}$ & \best{99.6} & \best{97.0}$^{*}$ & \best{99.9}$^{*}$ & \best{100.0} & \best{100.0}$^{\dagger}$ & \best{100.0}$^{\dagger}$ & \best{100.0}$^{\dagger}$ & \best{99.2}$^{\dagger}$ & & \best{82.2} & \best{80.0} & \best{87.1} & 84.0 & & Planning \\
\cmidrule(l){1-17}
Qwen3-VL-32B-Instruct & \best{62.1} & \second{27.5} & \best{29.6} & \best{88.8} & \second{81.2} & \second{28.7} & 28.7 & 1.7 & 2.9 & & 0.0 & 0.0 & \second{0.4} & 0.4 & & Rules \\
Qwen3-VL-8B-Thinking & 44.2 & \best{75.8} & \second{23.3} & 75.4 & \best{83.3} & \best{50.0} & \best{81.0} & \best{43.1} & \best{60.4} & & \best{1.3} & \best{8.9} & \best{5.8} & \best{32.4} & & Perception \\
Qwen3-VL-8B-Instruct & 24.2 & 24.2 & 11.2 & \second{82.9} & 24.6 & 20.7 & 22.0 & 0.8 & 2.5 & & \second{0.9} & 0.0 & 0.0 & 0.0 & & Rules \\
InternVL3.5-8B & 20.0 & 26.7 & 5.8 & 28.7 & 24.6 & 23.7 & \second{30.0} & \second{1.9} & \second{30.0} & & 0.0 & 0.0 & \second{0.4} & \second{5.3} & & Rules \\
MiniCPM-V 4.6 & \second{45.4} & 19.2 & 0.8 & 13.3 & 4.2 & 19.7 & 20.3 & 0.2 & 0.0 & & 0.0 & 0.0 & 0.0 & 0.0 & & Rules \\
\midrule
\grp{Maze} \\
Random guess & 25 & 25 & $<$3 & $<$3 & $<$3 & 33 & 33 & $<$3 & $<$3 & & 7.4 & 7.4 & 7.1 & 7.1 & & -- \\
GPT-6.1 & 95.4 & 92.5 & 97.5 & 99.6 & 97.1 & 98.0 & 87.0 & 87.5 & 82.1 & & 89.3 & 82.2 & 99.6 & 96.4 & & Planning \\
Gemini 3.8 Flash & \best{99.6} & \best{100.0} & \best{98.3} & \best{100.0} & \best{100.0} & \best{100.0} & \best{100.0} & \best{100.0} & \best{99.8} & & \best{97.8} & \best{98.7} & \best{100.0} & \best{100.0} & & Planning \\
\cmidrule(l){1-17}
Qwen3-VL-32B-Instruct & \second{63.3} & 32.1 & \best{34.2} & \second{84.6} & \best{82.5} & \second{41.0} & \second{39.7} & \second{6.0} & 6.5 & & \best{15.6} & 8.0 & \second{32.0} & 14.2 & & Rules \\
Qwen3-VL-8B-Thinking & 56.2 & \best{79.2} & \second{27.1} & 77.9 & \second{77.5} & \best{71.0} & \best{88.0} & \best{59.8} & \best{72.3} & & \second{13.8} & \best{48.9} & \best{46.7} & \best{84.4} & & Perception \\
Qwen3-VL-8B-Instruct & 45.0 & \second{35.4} & 11.7 & \best{92.9} & 26.2 & 33.7 & 35.7 & 5.8 & 4.0 & & 0.9 & 7.6 & 9.3 & 5.8 & & Rules \\
InternVL3.5-8B & 27.1 & 25.0 & 9.2 & 44.2 & 31.2 & 34.3 & 39.0 & 5.0 & \second{36.0} & & 10.2 & \second{28.4} & 8.9 & \second{65.8} & & Rules \\
MiniCPM-V 4.6 & \best{77.5} & 23.8 & 2.9 & 15.8 & 6.2 & 38.7 & 20.0 & 0.0 & 0.8 & & 0.0 & 0.4 & 2.7 & 4.0 & & Rules \\
\bottomrule
\end{tabular}
\caption{\textbf{Main results.} Probe accuracy and main-task success rate (\%) with image and text inputs on the same boards, for Sokoban (top) and the maze (bottom); within each pair, Text $-$ Image (right minus left) is the visual cost (\cref{eq:viscost}). \emph{Random guess}: score of random answers (probes) and random moves (main task); the solver scores 100. P3: outcome type; P4: final position; + ruler: image with coordinate labels; Single-turn: full plan in one response; Interactive: up to $K{=}3$ moves per turn. \tablemarks{} $^{*}$Four styles pooled. $^{\dagger}$6$\times$6 and 8$\times$8 boards.}
\label{tab:main}
\end{table*}

\subsection{Metrics and Diagnostic Criteria}
\label{sec:metrics}
\para{Visual cost.} For each task $s$, we measure text-input accuracy and image-input accuracy on the same boards and define the visual cost as
\begin{equation}
  \dvis(s) = \text{Acc}_{\text{text}}(s) - \text{Acc}_{\text{img}}(s).
  \label{eq:viscost}
\end{equation}
Because the two inputs carry the same information, $\dvis(s)$ is the accuracy lost when the board is read from pixels.
The text-input error, $100\% - \text{Acc}_{\text{text}}(s)$, remains even when the board is given exactly.
On P3, this error reflects rule application; on the main task, it reflects both rule application and planning.

\para{Failure labels.} Every answer is also labeled by how it fails.
P3 answers carry the five outcome types, and every main-task plan is replayed with the engine and ends as \emph{solved}, \emph{deadlock} (a push after which the solver proves the level unsolvable), \emph{unfinished} (the moves run out first), \emph{no push} (no box is ever moved), or \emph{no valid plan} (no parsable action sequence).

\para{Item-level attribution.} The visual cost shows how much accuracy is lost, and item-level attribution shows where.
In P3, an item with both a wrong outcome and a wrong start position indicates that the rule was applied to a misread board.
In P4, the new-error rate counts the steps at which the reported position is not the correct successor of the model's own previous answer, so each tracking error is counted once.

\para{Diagnostic criteria.} To turn these measurements into a diagnosis, we assign each model one primary bottleneck, the stage expected to limit its main-task performance, and test this expectation by fine-tuning in \cref{sec:ft}.
Rule application is checked first.
With text input, no visual perception is required, so a model that cannot apply the rules from text would remain limited by them even with perfect perception.
Significance is judged by whether the 95\% confidence interval excludes the threshold (\cref{sec:setup}).
\begin{enumerate}[leftmargin=*,itemsep=1pt,topsep=2pt]
\item \textbf{Rule application.} The model cannot apply the game rules to a board even when it is given as text. Three of the five outcome types in P3 are blocked moves, so a model that never predicts a blocked move scores at most 40\%, from the remaining two types (walk and push). A model whose P3 text accuracy is not significantly above 40\% is assigned this bottleneck.
\item \textbf{Visual perception.} The model applies the rules from text but loses accuracy when it reads the board from the image, and the loss carries over to the main task. A model whose visual cost is significantly above zero on P1 or P2 and on the interactive main task is assigned this bottleneck. We use the interactive setting because single-turn success is near zero for many models.
\item \textbf{Multi-step planning.} The model applies the rules and shows no visual cost that carries over to the main task, yet it still fails some levels from text, which places the problem in searching for a plan. Models that meet neither criterion above are assigned this bottleneck.
\end{enumerate}

\section{Experiments}
\label{sec:exp}

We evaluate seven MLLMs on \bench to locate where each model fails between visual perception and planning.

\subsection{Experimental Setup}
\label{sec:setup}

\para{Models.} We evaluate two proprietary models (GPT-6.1~\cite{gpt61sol} and Gemini 3.8 Flash~\cite{gemini38flash}) and five open-source models (Qwen3-VL-32B-Instruct, Qwen3-VL-8B-Instruct, Qwen3-VL-8B-Thinking~\cite{qwen3vl}, InternVL3.5-8B~\cite{internvl35}, and MiniCPM-V 4.6~\cite{minicpmv46}).
The two Qwen3-VL-8B models share one architecture but are post-trained separately.

\para{Protocol.} All models are evaluated without in-context examples.
The main task uses 225 Sokoban levels with one to three boxes, and each probe uses 240 to 480 items.
Every item is evaluated with both image and text inputs.
In the main task, the move budget is three times the reference solution length, and an interactive episode ends when the level is solved, when the solver confirms a deadlock, or when the move or turn limit is reached.
We report 95\% bootstrap confidence intervals over boards and pair all differences by board.
The proprietary models are evaluated on P3 and P4 with 6$\times$6 and 8$\times$8 boards.
Each open-source model answers 5,085 Sokoban queries for the probes and the single-turn task alone; decoding settings, prompts, and item counts are given in Supp.~B and C.

\subsection{Main Task Results}
\label{sec:main}

\Cref{tab:main} reports all probe and main-task results; we first discuss the main task (right columns).

\para{Success rates alone do not reveal why a model fails.} In the single-turn setting, the five open-source models solve at most 1.3\% of levels from images, and four of them remain below 1\% even in the interactive setting, close to the 0.5\% of random moves.
These scores are nearly identical, yet the probes in \cref{sec:probe-results} show different failing stages.

\para{Proprietary models are limited by planning.} GPT-6.1 and Gemini 3.8 Flash solve 61.8\% and 82.2\% of levels from images in a single turn, and replacing the image with text changes these rates by less than 5 points.
In the interactive setting, their visual costs are $+$4.0 and $-$3.1 points, with confidence intervals that include zero, so reading the board does not explain their failures.
Instead, success falls as plans become longer with both inputs: from one-box to three-box levels, the single-turn success of GPT-6.1 drops from 78.7\% to 45.3\% with images and from 77.3\% to 53.3\% with text.

\begin{figure}[t]
\centering
\includegraphics[width=\linewidth]{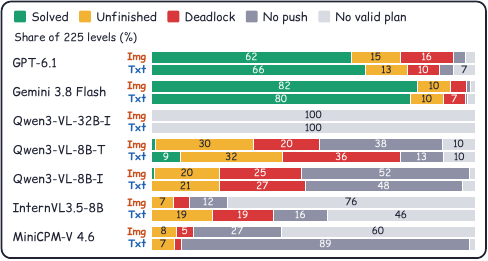}
\caption{\textbf{How single-turn plans end.} Outcome of every replayed plan (225 levels per input). No push: no box is ever moved. -T: Thinking; -I: Instruct.}
\label{fig:outcomes}
\end{figure}

\para{Replaying the plans shows how models fail.} We replay every single-turn plan and sort the outcomes into success and four failure types (\cref{fig:outcomes}; breakdowns in Supp.~E).
With images, \qT never pushes a box in 38\% of levels, whereas with text it most often pushes a box into a deadlock (36\%), which matches a reading failure on images and a planning failure on text.
\qI fails in the same way with both inputs, and on average 34--37\% of the moves in its plans walk into a wall or an immovable box, in line with its single-step errors in \cref{sec:probe-results}.
\qL never produces a valid plan, because every answer repeats a short action cycle until the token limit.

\subsection{Stage-Wise Probe Results}
\label{sec:probe-results}

Reading the probe columns of \cref{tab:main} along the chain reveals three failure profiles.
The proprietary models score above 86\% on every probe with both inputs, so their main-task failures arise after these stages.
The probes also explain a difference between the two models: GPT-6.1 tracks states less accurately than Gemini 3.8 Flash (87.9\% vs.\ 100.0\% from images), and it benefits more from the interactive setting, where the board is shown again after at most three moves (19.5 points vs.\ 4.9).

\begin{figure}[t]
\centering
\includegraphics[width=\linewidth]{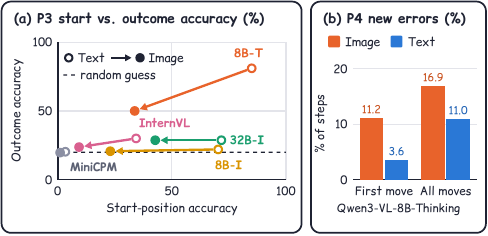}
\caption{\textbf{Error propagation.} (a) P3 accuracy on the start position and on the outcome; each arrow goes from text to image input. (b) Share of P4 steps with a new error (Qwen3-VL-8B-Thinking).}
\label{fig:propagation}
\end{figure}

For \qT, in contrast, the visual cost propagates from localization to later stages.
Its localization accuracy drops from 83.3\% with text to 23.3\% with images, the largest visual cost among its probes.
In P3, outcome accuracy drops from 81.0\% to 50.0\% together with start-position accuracy (\cref{fig:propagation}a), and 78\% of the wrong image answers also report a wrong start position, which indicates that the model applies the rule to a board it has misread.
The confusion matrices agree: most extra image errors predict a plain walk even though a wall or a box lies ahead (Supp.~E).
In P4, the errors come early: only 56.5\% of the 20-move image sequences are still fully correct after five moves (83.7\% for text), and a move into a wall causes a new error in 22.7\% vs.\ 7.3\% of steps.
Item-level attribution thus traces its downstream failures back to board reading.

The remaining open-source models fail single-step prediction even with text input.
Their P3 text accuracy lies between 20.3\% and 30.0\%, close to the 20\% chance level, and the three direct-answer models predict a blocked move for at most 10.3\% of text items, even though 60\% of the items are blocked.
To test whether these models lack the rules themselves, we repeat the text items without the board and state only what the two cells ahead of the player contain.
\qL, \qI, and InternVL3.5-8B then reach 98.3\%, 96.3\%, and 89.0\%, so they know the rules but cannot apply them to a board, where the two cells must first be found from the player's position and the move direction.
MiniCPM-V 4.6 stays at 22.7\% and never predicts a blocked move, so it alone lacks the rules themselves (Supp.~D).
Within the Qwen3 family, the split follows the reasoning mode: the Thinking model reaches 81.0\% and the Instruct model 22.0\%, as does the text-only Qwen3-8B~\cite{qwen3} with reasoning on or off (85.7\% vs.\ 22.3\%).

\subsection{Ablation Studies}
\label{sec:ablation}

\begin{figure}[t]
\centering
\includegraphics[width=\linewidth]{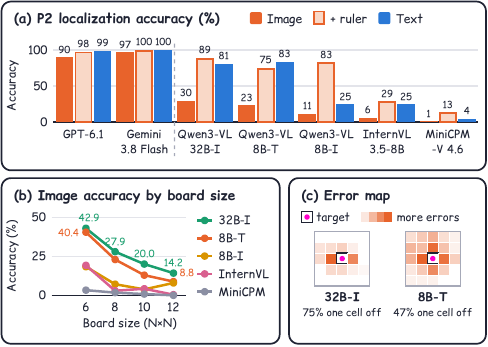}
\caption{\textbf{Localization (P2) ablations.} (a) Accuracy with image, image plus coordinate ruler, and text inputs. (b) Image accuracy by board size, four visual styles pooled. (c) Error map: where wrong answers land around the target cell (center); darker cells receive more errors.}
\label{fig:reading}
\end{figure}

Three controlled input changes pinpoint the missing visual skill: grid coordinate indexing, \ie, counting rows and columns from pixels (\cref{fig:reading}).
First, a coordinate ruler adds row and column labels at the border without changing the cells, yet it raises image localization from 23.3\% to 75.4\% for \qT and from 29.6\% to 88.8\% for \qL.
These models thus read printed labels but fail to count rows and columns from pixels, a weakness also reported on plain grids~\cite{vlmsareblind}.
Second, accuracy falls as there are more cells to count, from 40.4\% on 6$\times$6 to 8.8\% on 12$\times$12 boards for \qT.
Third, errors cluster around the target: one-cell offsets account for 75\% of the errors of \qL and 47\% of those of \qT, as expected from miscounting.

The ruler also carries over to the main task: with the same labels on the image, \qT solves 38.7\% of mazes in a single turn instead of 13.8\%, and its Sokoban success rises from 1.3\% to 7.6\% in a single turn and from 5.8\% to 18.7\% in the interactive setting.
These significant gains close 48--82\% of the gap to text input without training (Supp.~D): grid indexing drives much of the visual cost.

Other settings do not change these conclusions (all ten input conditions and interactive solving with $K{=}5$ and $7$ are in Supp.~D).
Grid lines raise the localization accuracy of \qT only from 13.8\% to 23.3\%, and the sprite set has little effect.
Even GPT-6.1 falls to 43.3\% on maze boards drawn with flat colors and no grid lines, and the ruler restores it to 88.8\%.

\subsection{Generalization to the Maze Grid Game}
\label{sec:maze}

To test whether the diagnosis depends on Sokoban, we repeat all tasks on the maze grid game, which keeps the sprites and task formats but removes the boxes (bottom of \cref{tab:main}).

The same pattern holds on the maze.
The proprietary models solve at least 82.2\% of mazes in a single turn and 96.4\% interactively, with either input.
\qT again loses accuracy at localization (27.1\% with images vs.\ 77.5\% with text), and the ruler restores it (77.9\%).
Meanwhile, the four models that fail single-step prediction on Sokoban stay near the 33\% chance level on maze single-step prediction with text (between 20.0\% and 39.7\%).
With a rule threshold of 66.7\% (one of three maze outcome types is blocked), the criteria of \cref{sec:metrics} assign every model the same bottleneck as on Sokoban.

Because maze moves are reversible, planning is easier, which exposes a larger visual cost: for \qT it grows from 7.6 points in single-turn Sokoban to 26.6 in interactive Sokoban, 35.1 in the single-turn maze, and 37.7 in the interactive maze (84.4\% with text vs.\ 46.7\% with images).
Similarly, InternVL3.5-8B solves at most 5.3\% of Sokoban levels in any setting, yet with board feedback it solves 65.8\% of mazes with text and only 8.9\% with images.
When planning fails first, the success rate thus masks an underlying perception deficit.

\subsection{Bottleneck Analysis}
\label{sec:bottleneck}

The criteria of \cref{sec:metrics} separate the seven models (all quantities with confidence intervals are in Supp.~D).
Four open-source models are limited by rule application: the upper confidence bounds of their P3 text accuracy (23.0--36.2\%) lie below the 40\% threshold.
\qT passes this check (81.0\%, lower bound 76.5\%) and is limited by visual perception, with significant visual costs on P2 (60.0 points) and on the interactive main task (26.6 points).
The two proprietary models pass both checks and are assigned multi-step planning.
GPT-6.1 has a visual cost of 9.5 points on P2, but it does not carry over to the main task.
State tracking does not explain their failures: in the interactive setting the board is shown again after at most three moves, which leaves little to track, yet both models still fail 13--19\% of the levels, and 131 of the 140 failed episodes end in a deadlock, an irreversible push.

Each label marks the primary bottleneck, not every weakness: \qL also has a large visual cost on P2, but it fails P3 even from text.
Each label is thus a testable prediction of which training should help (\cref{sec:ft}).

\section{Diagnosis-Guided Fine-Tuning}
\label{sec:ft}

We now ask whether fine-tuning the diagnosed bottleneck improves later stages, which would confirm the diagnosis and yield a better solver.

\subsection{Setup and Hypotheses}
\label{sec:ft-setup}

\para{Models and data.} We fine-tune the two Qwen3-VL-8B models, which share one architecture and differ in their bottleneck: visual perception for the Thinking model and rule application for the Instruct model.
All training boards are newly generated and share no layout with the benchmark.
\readset contains 20,000 board-reading items with image input, which ask for the content and position of cells and for a transcription of the board.
\ruleset contains 20,000 single-step items in the P3 format with text input, plus 8,000 with image input.
\solveset contains 8,000 levels with solver solutions.

\para{Training.} For the Thinking model, \readset updates only the vision encoder and the projector through low-rank adapters~\cite{lora}, while the language model stays frozen.
Any downstream gain must therefore come from better board reading, since the module that reasons is unchanged; text inputs never pass through the trained modules, so text scores differ only by decoding variation.
All other runs train adapters on the language model (Supp.~F).

\para{Hypotheses.} H1 (on-target training): \readset should raise the image accuracy of the Thinking model toward its text accuracy on all later stages and improve solving from images, while text accuracy stays unchanged.
H2 (off-target training): for the Instruct model, \readset should improve localization but leave single-step prediction at chance, which should improve only with \ruleset.
As a diagnosis-free reference, we train the language model of the Thinking model on \solveset, the common recipe of training on the end task.

\subsection{Results}
\label{sec:ft-results}

\begin{table}[t]
\centering
\resizebox{\linewidth}{!}{%
\footnotesize
\setlength{\tabcolsep}{1.6pt}
\begin{tabular}{@{}l cc cc cc cc cc@{}}
\toprule
 & \multicolumn{2}{c}{P2} & \multicolumn{2}{c}{P3} & \multicolumn{2}{c}{P4} & \multicolumn{2}{c}{Single-turn} & \multicolumn{2}{c}{Interactive} \\
\cmidrule(lr){2-3}\cmidrule(lr){4-5}\cmidrule(lr){6-7}\cmidrule(lr){8-9}\cmidrule(l){10-11}
Training & \Img & \Txt & \Img & \Txt & \Img & \Txt & \Img & \Txt & \Img & \Txt \\
\midrule
\multicolumn{11}{@{}l}{\emph{Qwen3-VL-8B-Thinking (visual perception)}} \\
\quad Base & 23.3 & 83.3 & 50.0 & 81.0 & 43.1 & \best{60.4} & 1.3 & 8.9 & 5.8 & \best{32.4} \\
\quad + \readset & \best{90.8} & \best{84.6} & \best{89.3} & \best{84.0} & \best{67.9} & 57.5 & \best{10.2} & \best{10.7} & \best{20.4} & 28.4 \\
\quad + \solveset & -- & -- & 56.0 & 77.3 & -- & -- & 1.8 & 5.8 & 5.3 & 18.7 \\
\midrule
\multicolumn{11}{@{}l}{\emph{Qwen3-VL-8B-Instruct (rule application)}} \\
\quad Base & 11.2 & 24.6 & 20.7 & 22.0 & 0.8 & 2.5 & 0.9 & 0.0 & 0.0 & 0.0 \\
\quad + \readset & \best{100.0} & 25.8 & 20.7 & 25.0 & 0.0 & 2.1 & 0.0 & 0.0 & 0.4 & 0.0 \\
\quad + \ruleset & 16.2 & \best{47.1} & 44.7 & \best{98.7} & \best{3.3} & \best{4.4} & 0.0 & 0.4 & 0.9 & 0.0 \\
\quad + \readset + \ruleset & \best{100.0} & 25.0 & \best{84.3} & 43.0 & 1.5 & 2.5 & 0.0 & 0.0 & 0.0 & 0.0 \\
\bottomrule
\end{tabular}}
\caption{\textbf{Fine-tuning results (\%).} \readset: board-reading data; \ruleset: single-step data (text items; image items when combined with \readset); \solveset: solver solutions. Bold: best per model (unmarked if all runs are below 1\%). --: not evaluated.}
\label{tab:ft}
\end{table}

\para{H1 holds: training the diagnosed stage improves the whole chain.} After \readset, the image localization accuracy of \qT rises from 23.3\% to 90.8\% (\cref{tab:ft}).
More importantly, the gain extends to stages that receive no training: image accuracy rises from 50.0\% to 89.3\% on P3 and from 43.1\% to 67.9\% on P4, while text accuracy stays within 4 points of the base model.
This follows the propagation path identified in \cref{sec:probe-results}, and among wrong P3 outcomes, the share with a wrong start position correspondingly falls from 78\% to 6\%.
Because the language model is frozen, these gains confirm that the earlier failures on P3 and P4 came from misreading the board.

\begin{figure}[t]
\centering
\includegraphics[width=\linewidth]{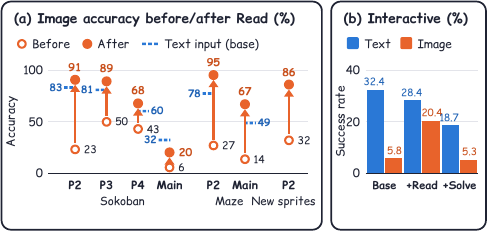}
\caption{\textbf{Fine-tuning \qT on \readset.} (a) Image accuracy (success rate for Main) before and after training, rounded; dashed ticks and blue values: text accuracy of the base model. (b) Interactive success rate.}
\label{fig:ft}
\end{figure}

The gain also carries over to the main task but stops at planning. Single-turn success with images rises from 1.3\% to 10.2\%, which matches the text level (10.7\%).
Interactive success with images rises from 5.8\% to 20.4\%, a 3.5-fold increase (paired gain of 14.7 points [9.3, 20.0]) obtained without any training on the main task, and the visual cost shrinks from 26.6 to 8.0 points (\cref{fig:ft}). Success rises on every board size, from 0.0\% to 16.0\% on $10{\times}10$ boards.
A coordinate ruler at test time gives a similar gain (18.7\%, \cref{sec:ablation}), pointing to grid indexing as the skill that \readset improves.
The remaining failures lie at the next stage, planning: deadlocks end about 155 of the 225 interactive episodes, as before training.

\para{H2 holds: the same data does not help a model with a different bottleneck.} \readset brings \qI to 100\% image localization, yet its P3 accuracy with images stays at 20.7\%.
The model now reads the start position correctly in 98.3\% of items but still predicts the wrong outcome, so localization was not its limiting factor.
\ruleset raises its P3 accuracy with text from 22.0\% to 98.7\%, but with images only to 44.7\%, since the board is still misread.
With both sets (\ruleset as image items), image P3 accuracy reaches 84.3\%.
The bottleneck then moves to the next stage, which P4 locates: state tracking stays at 1.5\%, so no level is solved.

\para{Training on solver solutions does not help.} After training on \solveset, interactive success stays at 5.3\% with images and falls from 32.4\% to 18.7\% with text, a paired change of $-$13.8 points [$-$20.0, $-$7.6].
During training, the validation loss keeps decreasing while the validation success rate falls from 7.0\% to 3.0\%: fitting solution strings does not teach the model to solve levels.
Instead, the trained model commits more moves per turn (2.76 vs.\ 2.29 with text) and ends more episodes in a deadlock (159 vs.\ 135).

\subsection{Zero-Shot Cross-Game Transfer}
\label{sec:ft-transfer}

The maze never appears in training, yet after \readset, \qT solves 67.1\% of mazes with images in a single turn, up from 13.8\% and above the 52.0\% it reaches from text, and its maze localization rises from 27.1\% to 95.4\%.
On a sprite set unseen in training, image localization also rises from 32.1\% to 86.2\%.
The gain holds even for the longest mazes (8\% to 51\%). These results indicate that the trained skill is grid indexing, which both games share and which carries over to unseen sprites.

Supp.~G covers VSP mazes~\cite{vsp2025} and general benchmarks.

\section{Conclusion}
\label{sec:conclusion}

We present \bench, a diagnostic benchmark that locates where multimodal LLMs fail between visual perception and planning.
Across seven MLLMs, \bench reveals differences that success rates alone do not show: proprietary models are limited by multi-step planning, whereas open-source models are limited by either visual perception or rule application, with grid coordinate indexing as the main visual deficit.
The diagnosis is also actionable.
Fine-tuning only the vision module of the perception-limited model on board reading improves every later stage, raises interactive solving from 5.8\% to 20.4\%, and transfers zero-shot to an unseen game (single-turn, 13.8\% to 67.1\%), all without main-task training, whereas the same data does not help a rule-limited model and training on solver solutions brings no gain.
\bench thus offers a systematic way to find why planning fails and to choose what to train: the diagnosed bottleneck, not the end task.
We will release the benchmark and code and plan to extend the diagnosis to embodied and interface tasks.

\section*{LLM Usage Statement}
We used the LLM assistants Claude (Anthropic), Gemini (Google), and ChatGPT (OpenAI) to draft and edit the text of all sections, to write the code that draws the figures and builds the tables from our result files, and to cross-check the numbers in the text against those files.
The benchmark design, the experiments, the analyses, and the conclusions are the authors' own, and the authors have checked all content and take full responsibility for it.

{
    \small
    \bibliographystyle{ieeenat_fullname}
    \bibliography{main}
}

\clearpage
\maketitlesupplementary
\appendix
\renewcommand{\thesection}{\Alph{section}}
\setcounter{figure}{0}
\setcounter{table}{0}
\renewcommand{\thefigure}{S\arabic{figure}}
\renewcommand{\thetable}{S\arabic{table}}
\renewcommand{\theHfigure}{S\arabic{figure}}
\renewcommand{\theHtable}{S\arabic{table}}
\renewcommand{\topfraction}{0.85}
\renewcommand{\bottomfraction}{0.70}
\renewcommand{\textfraction}{0.12}
\renewcommand{\floatpagefraction}{0.75}
\renewcommand{\dbltopfraction}{0.86}
\renewcommand{\dblfloatpagefraction}{0.80}
\setcounter{topnumber}{3}
\setcounter{bottomnumber}{2}
\setcounter{totalnumber}{5}
\setcounter{dbltopnumber}{3}
\raggedbottom

\newcommand{\FloatFigPipeline}{%
\begin{figure*}[t]
\centering
\includegraphics[width=\linewidth]{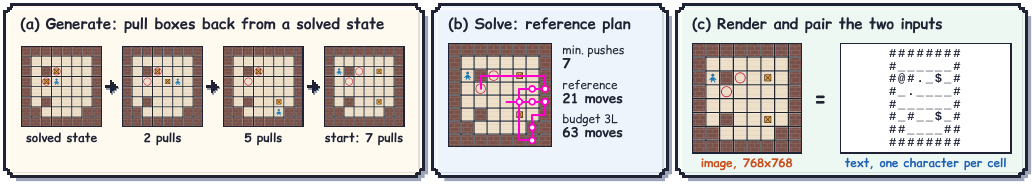}
\caption{\textbf{From a solved state to a paired benchmark item.} (a) A level is generated by pulling boxes backward from a solved state, so it is solvable by construction. (b) A push-optimal solver returns the reference plan; its length $L$ sets the move budget of $3L$. (c) The level is rendered as an image and as a text grid that carry the same information.}
\label{fig:pipeline}
\end{figure*}%
}

\newcommand{\FloatFigAtlas}{%
\begin{figure*}[t]
\centering
\includegraphics[width=\linewidth]{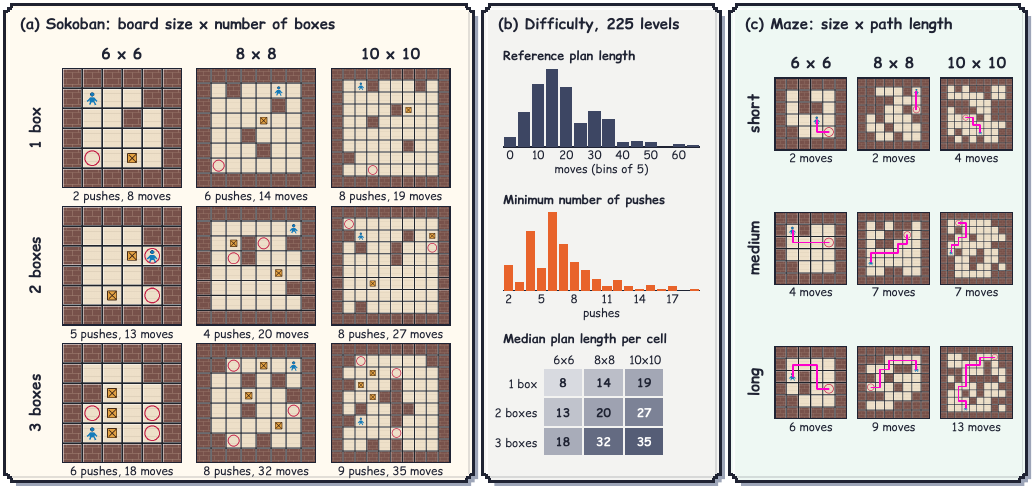}
\caption{\textbf{The levels of the main task.} (a) One Sokoban level for each board size and number of boxes, with the pushes and moves of its reference plan. (b) Reference plan length and minimum number of pushes over the 225 levels, and the median plan length per cell. (c) One maze for each board size and path-length group, with its shortest path in magenta.}
\label{fig:atlas}
\end{figure*}%
}

\newcommand{\FloatFigViews}{%
\begin{figure*}[t]
\centering
\includegraphics[width=\linewidth]{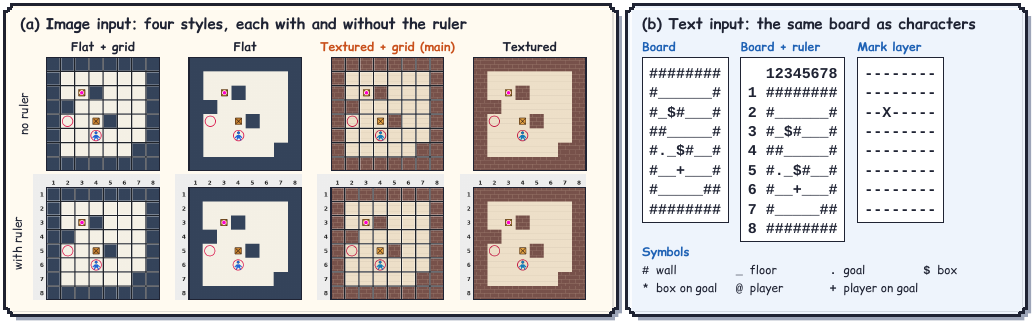}
\caption{\textbf{One board under all input conditions.} (a) The eight image conditions of P1 and P2: four visual styles, each without (top) and with (bottom) the coordinate ruler. (b) The text input: the board, the board with the ruler, and the mark layer that carries the marked cell.}
\label{fig:views}
\end{figure*}%
}

\newcommand{\FloatFigEndings}{%
\begin{figure*}[t]
\centering
\includegraphics[width=\linewidth]{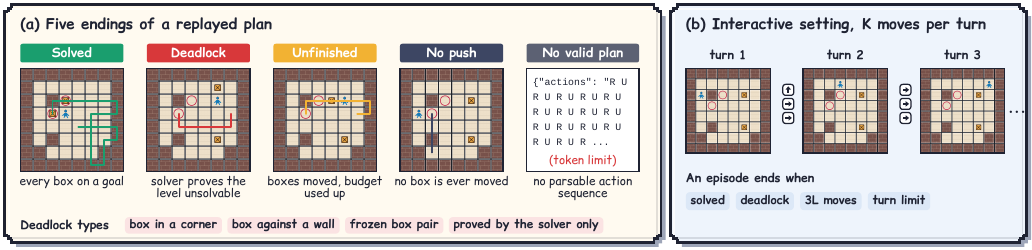}
\caption{\textbf{How a plan is replayed and ended.} (a) The five endings of a single-turn plan with one example each (the line is the path of the player), and the four deadlock types. (b) The interactive setting: the model sees the board, commits up to $K$ moves, and sees the new board.}
\label{fig:endings}
\end{figure*}%
}

\newcommand{\FloatFigCardsA}{%
\begin{figure*}[t]
\centering
\includegraphics[width=\linewidth]{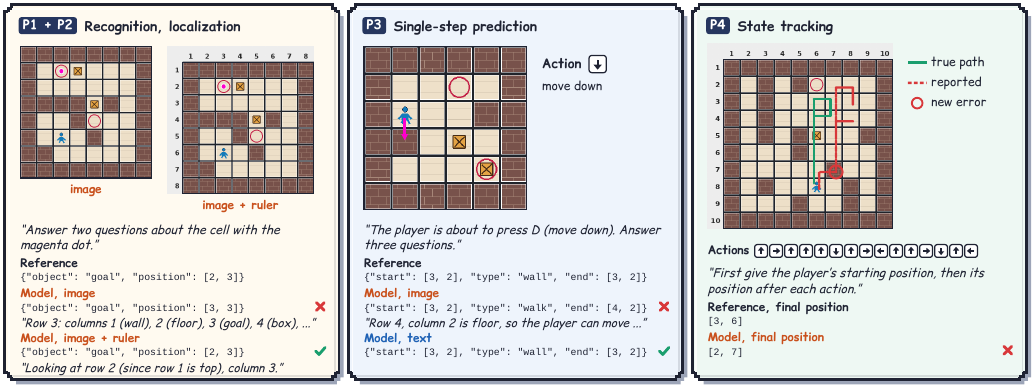}
\caption{\textbf{The probes P1--P4 on real items, with answers of \qT.} P1 + P2: the model names the correct object and reports a position one row off; with the ruler it gives the correct cell. P3: from the image the model reads the wall below the player as floor and predicts a walk; from text it answers correctly. P4: one early error shifts all later positions; the new-error rate counts it once.}
\label{fig:cards1}
\end{figure*}%
}

\newcommand{\FloatFigCardsB}{%
\begin{figure*}[t]
\centering
\includegraphics[width=\linewidth]{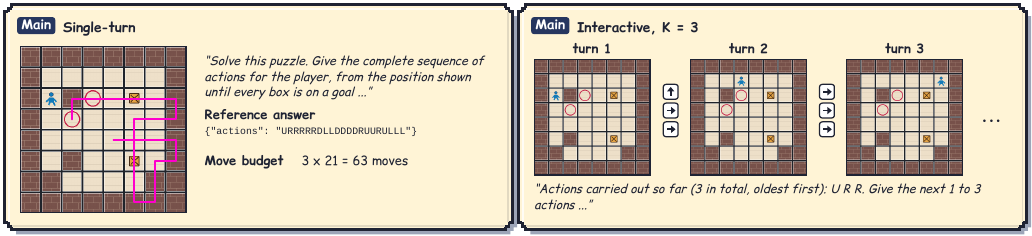}
\caption{\textbf{The main task of Sokoban.} Single-turn: the whole plan in one response (reference plan in magenta). Interactive: up to $K{=}3$ moves per turn, with the board shown again after each turn.}
\label{fig:cards2}
\end{figure*}%
}

\newcommand{\FloatFigMazecards}{%
\begin{figure*}[t]
\centering
\includegraphics[width=\linewidth]{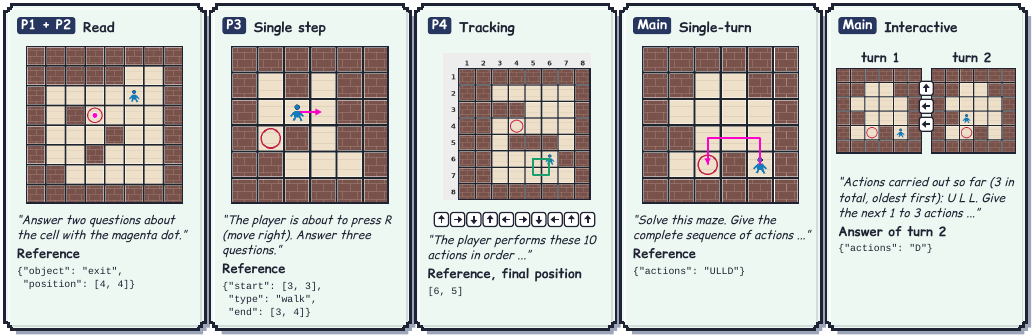}
\caption{\textbf{Every task on the maze, with real items.} The maze keeps the sprites, the coordinate convention, and the answer formats of Sokoban and removes the boxes. P1 has four classes (wall, floor, exit, player) and P3 three outcome types (walk, wall, exit). P4 uses the coordinate ruler, as in Sokoban; the green line is the true path.}
\label{fig:mazecards}
\end{figure*}%
}

\newcommand{\FloatFigPromptPair}{%
\begin{figure*}[t]
\begin{minipage}[t]{0.495\linewidth}\vspace{0pt}
\begin{promptbox}{P3 Single-step, image input \quad (the board image is attached)}
This is a Sokoban puzzle. The player moves one cell per action: U = up, D = down, L = left, R = right. The player cannot walk into a wall. When the player walks into a box, the box is pushed one cell in the same direction, but only if the cell beyond the box is floor or a goal; a box cannot be pushed into a wall or into another box. If a move is blocked, nothing changes. Boxes cannot be pulled, and the player cannot push two boxes at once. The puzzle is solved when every box is on a goal.\\[2pt]
\pimg{The image shows a Sokoban board. Red-brown brick cells are walls, beige plank cells are floor, red rings are goals, orange crates marked with an X are boxes, and the blue figure is the player. A red ring drawn around a box or the player means it is standing on a goal. Ignore the gray background outside the board.}\\[2pt]
Positions are written as [row, column]. \pimg{Row 1 is the top row of the board and column 1 is the leftmost column. The outer wall is included, so the top-left wall cell is [1, 1].} Rows increase downward and columns increase to the right. The board has 6 rows and 6 columns.\\[2pt]
The player is about to press D (move down). Answer three questions.\\
"start": the player's position now.\\
"type": what this action does. Choose one:\\
\hspace*{0.6em}"walk": the player moves into an empty floor or goal cell;\\
\hspace*{0.6em}"wall": the player walks into a wall and does not move;\\
\hspace*{0.6em}"push": the player pushes a box one cell and moves into the box's old cell;\\
\hspace*{0.6em}"box\_wall": the player walks into a box, but the cell beyond the box is a wall, so nothing moves;\\
\hspace*{0.6em}"box\_box": the player walks into a box, but the cell beyond the box holds another box, so nothing moves.\\
"end": the player's position after the action.\\[2pt]
Return ONLY a JSON object, with no explanation, in this format: \{"start": [row, column], "type": "<walk|wall|push|box\_wall|box\_box>", "end": [row, column]\}
\end{promptbox}
\end{minipage}\hfill
\begin{minipage}[t]{0.495\linewidth}\vspace{0pt}
\begin{promptbox}{P3 Single-step, text input}
This is a Sokoban puzzle. The player moves one cell per action: U = up, D = down, L = left, R = right. The player cannot walk into a wall. When the player walks into a box, the box is pushed one cell in the same direction, but only if the cell beyond the box is floor or a goal; a box cannot be pushed into a wall or into another box. If a move is blocked, nothing changes. Boxes cannot be pulled, and the player cannot push two boxes at once. The puzzle is solved when every box is on a goal.\\[2pt]
\ptxt{The Sokoban board is given as text, one line per row: \# wall, \_ floor, . goal, \$ box, * box on a goal, @ player, + player on a goal.}\\[2pt]
Positions are written as [row, column]. \ptxt{Row 1 is the first line of the board and column 1 is the first character of each line. The outer wall is included, so the top-left \# is [1, 1].} Rows increase downward and columns increase to the right. The board has 6 rows and 6 columns.\\[2pt]
\ptxt{Board:\\
\fence\\
\#\#\#\#\#\#\\
\#\_\_.\_\#\\
\#@\_\_\#\#\\
\#\#\_\$\_\#\\
\#\_\_\_*\#\\
\#\#\#\#\#\#\\
\fence}\\[2pt]
The player is about to press D (move down). Answer three questions.\\
"start": the player's position now.\\
"type": what this action does. Choose one:\\
\hspace*{0.6em}"walk": \ldots{} "wall": \ldots{} "push": \ldots{} "box\_wall": \ldots{} "box\_box": \ldots{}\quad\slot{the five definitions, as on the left}\\
"end": the player's position after the action.\\[2pt]
Return ONLY a JSON object, with no explanation, in this format: \{"start": [row, column], "type": "<walk|wall|push|box\_wall|box\_box>", "end": [row, column]\}
\end{promptbox}
\end{minipage}
\caption{\textbf{The same question with two inputs.} Complete P3 prompt of the item in \cref{fig:cards1}. Regular text is identical in the two versions. The sentences in italics describe the input: sprites and image coordinates on the left; characters, text coordinates, and the board itself on the right. The legend sentence changes with the sprite set; the one shown is for the textured set.}
\label{fig:prompt-pair}
\end{figure*}%
}

\newcommand{\FloatFigPromptTasks}{%
\begin{figure*}[t]
\begin{minipage}[t]{0.495\linewidth}\vspace{0pt}
\begin{promptbox}{P1 + P2 \quad order: legend, mark, [ruler], coordinates, [board, mark layer], question}
\vimg{A large magenta dot with a white border is drawn in the centre of one cell.}
\vtxt{The mark layer below has the same shape as the board: X marks one cell, and - means no mark.}\\
\slot{with ruler}
\vimg{Row numbers are printed to the left of the board and column numbers above it.}
\vtxt{Row numbers are written to the left of each line and column numbers above the board; they are not part of the board.}\\[2pt]
Answer two questions about the cell \vimg{with the magenta dot.} \vtxt{marked X in the mark layer.}\\
"object": what the cell contains\vimg{, ignoring the dot itself}. Choose one of: wall, floor, goal, box, player.\\
"position": the position of that cell.\\[2pt]
Return ONLY a JSON object, with no explanation, in this format: \{"object": "<wall|floor|goal|box|player>", "position": [row, column]\}
\end{promptbox}
\begin{promptbox}{P4 State tracking \quad order: rules, legend, ruler, coordinates, [board with ruler], question}
The player performs these 15 actions in order: U R U U U D U R L U U R D U L\\
First give the player's starting position, then its position after each action. If an action is blocked, the position stays the same.\\
The list must contain exactly 16 positions: the start, then one after each action.\\[2pt]
Return ONLY a JSON object, with no explanation, in this format: \{"trajectory": [[row, column], ...]\}
\end{promptbox}
\begin{promptbox}{Main task, single-turn \quad order: rules, legend, [board], question}
Solve this puzzle. Give the complete sequence of actions for the player, from the position shown until every box is on a goal, using only the letters U, D, L and R.\\[2pt]
Return ONLY a JSON object, with no explanation, in this format: \{"actions": "<a string of U, D, L, R>"\}
\end{promptbox}
\end{minipage}\hfill
\begin{minipage}[t]{0.495\linewidth}\vspace{0pt}
\begin{promptbox}{Main task, interactive ($K{=}3$) \quad order: rules, legend, current board, question}
\vtxt{Current board (all the actions listed below have already been carried out; this is the position now):} \slot{board}\\
This puzzle is played in turns. In each turn you give the next 1 to 3 actions; they are carried out, and then you are shown the new board.\\
Actions carried out so far (1 in total, oldest first): R\\
Give the next 1 to 3 actions for the player, using only the letters U, D, L and R.\\[2pt]
Return ONLY a JSON object, with no explanation, in this format: \{"actions": "<1 to 3 letters>"\}
\end{promptbox}
\begin{promptbox}{Maze \quad the same templates with these sentences}
\slot{rules} This is a maze. The player moves one cell per action: U = up, D = down, L = left, R = right. The player cannot walk into a wall. If a move is blocked, nothing changes. The maze is solved when the player stands on the exit.\\[2pt]
\slot{legend} \vimg{The image shows a maze. Red-brown brick cells are walls, beige plank cells are floor, the red ring is the exit, and the blue figure is the player. A red ring drawn around the player means it is standing on the exit. Ignore the gray background outside the board.}
\vtxt{The maze is given as text, one line per row: \# wall, \_ floor, . exit, @ player, + player on the exit.}\\[2pt]
\slot{P1 classes} wall, floor, exit, player\\
\slot{P3 types} "walk": the player moves into an empty floor cell; "wall": the player walks into a wall and does not move; "exit": the player moves onto the exit cell.\\
\slot{P4, added} If the player steps onto the exit, it simply continues with the next action.\\
\slot{main task} Solve this maze. Give the complete sequence of actions that takes the player from the position shown to the exit, using only the letters U, D, L and R.
\end{promptbox}
\end{minipage}
\caption{\textbf{Questions and answer formats of all tasks.} Each box gives the task-specific part of the prompt and, in its title, the order of the shared sentences (rules, legend, and coordinate convention as in \cref{fig:prompt-pair}; parts in square brackets are present with text input or with the ruler). Parts after $\langle$image$\rangle$ or $\langle$text$\rangle$ are used with that input, up to the end of the sentence; all other text is common to both inputs. The interactive prompt is shown for text input; with image input the current board is the attached image.}
\label{fig:prompt-tasks}
\end{figure*}%
}

\newcommand{\FloatTabStats}{%
\begin{table*}[t]
\centering
\footnotesize
\setlength{\tabcolsep}{5pt}
\fittab{%
\begin{tabular}{@{}l l r r l r@{}}
\toprule
Task & Boards & Items & Sizes & Input conditions & Queries \\
\midrule
\multicolumn{6}{@{}l}{\textit{Sokoban}} \\
P1 + P2 (one query) & 120 & 240 & 6, 8, 10, 12 & 4 styles $\times$ ruler on/off; text $\times$ ruler on/off & 2,400 \\
P3 Single-step & 47 & 300 & 6, 8, 10 & main style; text & 600 \\
P4 Tracking & 120 & 480 & 6, 8, 10 & main style + ruler; text + ruler & 960 \\
Main task, single-turn & 225 & 225 & 6, 8, 10 & 4 styles; text & 1,125 \\
Main task, interactive & 225 & 225 & 6, 8, 10 & main style; text; $K \in \{3, 5, 7\}$ & episodes \\
\midrule
\multicolumn{6}{@{}l}{\textit{Maze}} \\
P1 + P2 (one query) & 120 & 240 & 6, 8, 10, 12 & 4 styles $\times$ ruler on/off; text $\times$ ruler on/off & 2,400 \\
P3 Single-step & -- & 300 & 6, 8, 10 & main style; text & 600 \\
P4 Tracking & 120 & 480 & 6, 8, 10 & main style + ruler; text + ruler & 960 \\
Main task, single-turn & 225 & 225 & 6, 8, 10 & 4 styles; text & 1,125 \\
Main task, interactive & 225 & 225 & 6, 8, 10 & main style; text; $K{=}3$ & episodes \\
\bottomrule
\end{tabular}
}
\caption{\textbf{Composition of \bench.} \emph{Items}: distinct questions; \emph{Queries}: items $\times$ input conditions for one model. P3 has 60 items per outcome type on Sokoban (five types) and 100 per type on the maze (three types); P4 has 120 items for each length $H \in \{5, 10, 15, 20\}$. The main style is the textured sprite set with grid lines.}
\label{tab:stats}
\end{table*}
}

\newcommand{\FloatFigFactors}{%
\begin{figure*}[t]
\centering
\includegraphics[width=\linewidth]{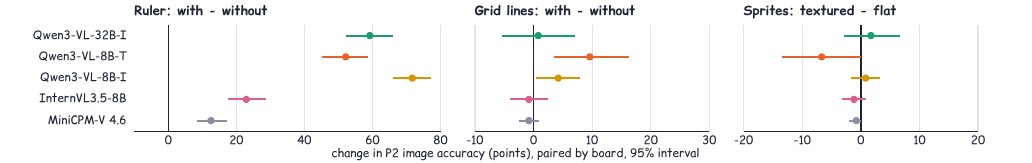}
\caption{\textbf{Effect of each rendering factor on image localization (P2).} Paired change in accuracy when the coordinate ruler is added (main style), when grid lines are added (textured sprites), and when flat sprites are replaced by textured ones (with grid lines).}
\label{fig:factors}
\end{figure*}%
}

\newcommand{\FloatFigDiag}{%
\begin{figure*}[t]
\centering
\includegraphics[width=\linewidth]{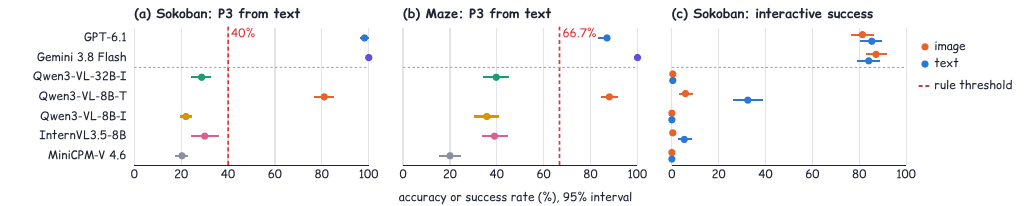}
\caption{\textbf{The two checks of the diagnostic criteria, with 95\% intervals.} (a, b) Rule check: P3 accuracy from text against the threshold of each game (dashed). No interval touches a threshold. (c) Perception check: interactive success from image and from text. For \qT the two intervals are far apart; for the proprietary models they overlap. Models above the dotted line are proprietary.}
\label{fig:diag}
\end{figure*}%
}

\newcommand{\FloatTabConditions}{%
\begin{table*}[t]
\centering
\footnotesize
\setlength{\tabcolsep}{4.2pt}
\fittab{%
\begin{tabular}{@{}l cccc cccc cc@{}}
\toprule
 & \multicolumn{4}{c}{\Img, no ruler} & \multicolumn{4}{c}{\Img, with ruler} & \multicolumn{2}{c}{\Txt} \\
\cmidrule(lr){2-5}\cmidrule(lr){6-9}\cmidrule(l){10-11}
Model & Flat+G & Flat & Tex+G & Tex & Flat+G & Flat & Tex+G & Tex & no ruler & ruler \\
\midrule
\multicolumn{11}{@{}l}{\textit{P1 Recognition}} \\
GPT-6.1 & \multicolumn{4}{c}{94.3 (pooled)} & \multicolumn{4}{c}{92.8 (pooled)} & 98.3 & 95.4 \\
Gemini 3.8 Flash & \multicolumn{4}{c}{100.0 (pooled)} & \multicolumn{4}{c}{99.9 (pooled)} & \best{99.6} & \best{99.2} \\
\cmidrule(l){1-11}
Qwen3-VL-32B-Instruct & \best{57.5} & \best{63.7} & \best{62.1} & \best{71.2} & \best{60.4} & \best{67.5} & \best{64.6} & \best{68.8} & \second{27.5} & \second{53.8} \\
Qwen3-VL-8B-Thinking & \second{48.8} & \second{51.7} & 44.2 & \second{49.2} & \second{49.6} & \second{55.4} & \second{50.4} & \second{59.2} & \best{75.8} & \best{84.6} \\
Qwen3-VL-8B-Instruct & 27.1 & 29.2 & 24.2 & 31.7 & 29.6 & 26.7 & 32.1 & 34.6 & 24.2 & 44.6 \\
InternVL3.5-8B & 20.0 & 20.0 & 20.0 & 20.0 & 20.0 & 20.0 & 20.0 & 20.0 & 26.7 & 35.0 \\
MiniCPM-V 4.6 & 38.8 & 35.4 & \second{45.4} & 40.4 & 40.0 & 37.1 & 46.2 & 42.5 & 19.2 & 25.4 \\
\midrule
\multicolumn{11}{@{}l}{\textit{P2 Localization}} \\
GPT-6.1 & \multicolumn{4}{c}{89.7 (pooled)} & \multicolumn{4}{c}{97.9 (pooled)} & 99.2 & \best{100.0} \\
Gemini 3.8 Flash & \multicolumn{4}{c}{97.0 (pooled)} & \multicolumn{4}{c}{99.9 (pooled)} & \best{100.0} & \best{100.0} \\
\cmidrule(l){1-11}
Qwen3-VL-32B-Instruct & \second{27.9} & \best{18.8} & \best{29.6} & \best{28.7} & \best{87.1} & \best{89.2} & \best{88.8} & \best{88.3} & \second{81.2} & \best{89.6} \\
Qwen3-VL-8B-Thinking & \best{30.0} & \second{17.9} & \second{23.3} & \second{13.8} & 76.2 & \second{60.8} & 75.4 & \second{67.9} & \best{83.3} & \best{89.6} \\
Qwen3-VL-8B-Instruct & 10.4 & 8.3 & 11.2 & 7.1 & \second{85.4} & 60.0 & \second{82.9} & 62.9 & 24.6 & \second{54.2} \\
InternVL3.5-8B & 7.1 & 7.1 & 5.8 & 6.7 & 35.8 & 31.7 & 28.7 & 32.9 & 24.6 & 44.2 \\
MiniCPM-V 4.6 & 1.7 & 1.7 & 0.8 & 1.7 & 11.7 & 7.9 & 13.3 & 7.5 & 4.2 & 13.8 \\
\bottomrule
\end{tabular}
}
\caption{\textbf{P1 and P2 accuracy (\%) under all ten input conditions.} Flat / Tex: flat-color or textured sprites; +G: with grid lines; Tex+G is the main style of the paper. Each cell has 240 items; proprietary models are pooled over the four styles (960 items). \tablemarks}
\label{tab:conditions}
\end{table*}
}

\newcommand{\FloatTabPSize}{%
\begin{table*}[t]
\centering
\footnotesize
\setlength{\tabcolsep}{3.6pt}
\fittab{%
\begin{tabular}{@{}l cccc cccc cccc c@{}}
\toprule
 & \multicolumn{4}{c}{\Img} & \multicolumn{4}{c}{\Img{} + ruler} & \multicolumn{4}{c}{\Txt} & One-cell \\
\cmidrule(lr){2-5}\cmidrule(lr){6-9}\cmidrule(lr){10-13}
Model & 6 & 8 & 10 & 12 & 6 & 8 & 10 & 12 & 6 & 8 & 10 & 12 & errors \\
\midrule
GPT-6.1 & 95.0 & 95.8 & 88.3 & 79.6 & 98.3 & 98.8 & 97.9 & 96.7 & \best{100.0} & 98.3 & \best{100.0} & \best{98.3} & -- \\
Gemini 3.8 Flash & \best{100.0} & \best{100.0} & \best{89.2} & \best{90.4} & \best{100.0} & \best{100.0} & \best{100.0} & \best{99.2} & \best{100.0} & \best{100.0} & \best{100.0} & \best{98.3} & -- \\
\midrule
Qwen3-VL-32B-Instruct & \best{42.9} & \best{27.9} & \best{20.0} & \best{14.2} & \second{85.8} & \best{94.6} & \best{87.1} & \best{85.8} & \second{83.3} & \second{85.0} & \second{80.0} & \best{76.7} & 75\% \\
Qwen3-VL-8B-Thinking & \second{40.4} & \second{22.9} & \second{12.9} & \second{8.8} & 82.5 & 67.5 & \second{71.2} & 59.2 & \best{96.7} & \best{90.0} & \best{85.0} & \second{61.7} & 47\% \\
Qwen3-VL-8B-Instruct & 18.3 & 7.1 & 3.8 & 7.9 & \best{87.1} & \second{73.8} & 70.0 & \second{60.4} & 41.7 & 30.0 & 16.7 & 10.0 & 33\% \\
InternVL3.5-8B & 19.2 & 2.9 & 4.2 & 0.4 & 40.0 & 21.7 & 32.1 & 35.4 & 43.3 & 25.0 & 20.0 & 10.0 & 21\% \\
MiniCPM-V 4.6 & 3.3 & 1.7 & 0.8 & 0.0 & 18.3 & 12.5 & 3.3 & 6.2 & 6.7 & 3.3 & 3.3 & 3.3 & 7\% \\
\bottomrule
\end{tabular}
}
\caption{\textbf{P2 accuracy (\%) by board size (number of cells per side).} Image and Image + ruler: four styles pooled, 240 items per cell; Text: 60 items per cell. \emph{One-cell errors}: share of the wrong image answers (main style, no ruler) that name a cell directly above, below, left, or right of the target. \tablemarks}
\label{tab:p2size}
\end{table*}
}

\newcommand{\FloatTabDiag}{%
\begin{table*}[t]
\centering
\scriptsize
\setlength{\tabcolsep}{3.4pt}
\fittab{%
\begin{tabular}{@{}l c cc c cc c l@{}}
\toprule
 & P3 text & \multicolumn{3}{c}{P2 Localization} & \multicolumn{3}{c}{Interactive main task} & \\
\cmidrule(lr){3-5}\cmidrule(lr){6-8}
Model & [95\% CI] & \Img & \Txt & $\dvis$ & \Img & \Txt & $\dvis$ [95\% CI] & Bottleneck \\
\midrule
\multicolumn{9}{@{}l}{\textit{Sokoban (rule threshold 40\%)}} \\
GPT-6.1 & 98.2 [96.1, 100.0]$^{\dagger}$ & 89.7 [87.7, 91.6]$^{*}$ & 99.2 [97.9, 100.0] & $+$9.5 & 81.3 [76.4, 86.2] & 85.3 [80.4, 89.8] & $+$4.0 [$-$2.2, 10.2] & Planning \\
Gemini 3.8 Flash & 100.0 [100.0, 100.0]$^{\dagger}$ & 97.0 [95.5, 98.2]$^{*}$ & 100.0 [100.0, 100.0] & $+$3.0 & 87.1 [82.7, 91.6] & 84.0 [79.1, 88.9] & $-$3.1 [$-$8.0, 1.8] & Planning \\
\cmidrule(l){1-9}
Qwen3-VL-32B-Instruct & 28.7 [24.0, 32.8] & 29.6 [23.3, 36.2] & 81.2 [76.2, 85.8] & $+$51.6 & 0.4 [0.0, 1.3] & 0.4 [0.0, 1.3] & $+$0.0 & Rules \\
Qwen3-VL-8B-Thinking & 81.0 [76.5, 85.3] & 23.3 [17.5, 29.6] & 83.3 [78.3, 87.9] & $+$60.0 & 5.8 [3.1, 8.9] & 32.4 [26.2, 38.7] & $+$26.6 & Perception \\
Qwen3-VL-8B-Instruct & 22.0 [19.3, 24.4] & 11.2 [7.1, 15.8] & 24.6 [19.2, 30.4] & $+$13.4 & 0.0 [0.0, 0.0] & 0.0 [0.0, 0.0] & $+$0.0 & Rules \\
InternVL3.5-8B & 30.0 [24.3, 36.2] & 5.8 [2.9, 9.2] & 24.6 [19.2, 30.0] & $+$18.8 & 0.4 [0.0, 1.3] & 5.3 [2.7, 8.4] & $+$4.9 & Rules \\
MiniCPM-V 4.6 & 20.3 [17.5, 23.0] & 0.8 [0.0, 2.1] & 4.2 [2.1, 6.7] & $+$3.4 & 0.0 [0.0, 0.0] & 0.0 [0.0, 0.0] & $+$0.0 & Rules \\
\midrule
\multicolumn{9}{@{}l}{\textit{Maze (rule threshold 66.7\%)}} \\
GPT-6.1 & 87.0$^{\ddagger}$ & 97.5$^{\ddagger}$ & 97.1$^{\ddagger}$ & $-$0.4 & 99.6 [98.7, 100.0] & 96.4 [93.8, 98.7] & $-$3.2 & Planning \\
Gemini 3.8 Flash & 100.0 [100.0, 100.0] & 98.3 [96.7, 99.6] & 100.0 [100.0, 100.0] & $+$1.7 & 100.0 [100.0, 100.0] & 100.0 [100.0, 100.0] & $+$0.0 & Planning \\
\cmidrule(l){1-9}
Qwen3-VL-32B-Instruct & 39.7 [34.0, 45.3] & 34.2 [27.9, 40.0] & 82.5 [77.5, 87.1] & $+$48.3 & 32.0 [26.2, 38.2] & 14.2 [9.8, 18.7] & $-$17.8 & Rules \\
Qwen3-VL-8B-Thinking & 88.0 [84.3, 91.7] & 27.1 [21.7, 32.9] & 77.5 [72.1, 82.5] & $+$50.4 & 46.7 [40.4, 53.3] & 84.4 [79.6, 88.9] & $+$37.7 & Perception \\
Qwen3-VL-8B-Instruct & 35.7 [30.3, 41.0] & 11.7 [7.5, 15.8] & 26.2 [20.8, 32.1] & $+$14.5 & 9.3 [5.8, 13.3] & 5.8 [3.1, 8.9] & $-$3.5 & Rules \\
InternVL3.5-8B & 39.0 [33.7, 44.7] & 9.2 [5.8, 12.9] & 31.2 [25.4, 37.1] & $+$22.0 & 8.9 [5.3, 12.9] & 65.8 [59.6, 71.6] & $+$56.9 & Rules \\
MiniCPM-V 4.6 & 20.0 [15.3, 24.7] & 2.9 [0.8, 5.0] & 6.2 [3.3, 9.6] & $+$3.3 & 2.7 [0.9, 4.9] & 4.0 [1.8, 6.7] & $+$1.3 & Rules \\
\bottomrule
\end{tabular}
}
\caption{\textbf{Quantities used by the diagnostic criteria, with 95\% confidence intervals.} Accuracy and success rate in \%, visual cost $\dvis$ (text $-$ image) in points. A model is assigned \emph{Rules} if its P3 text interval does not lie above the threshold, \emph{Perception} if it passes this check and its visual cost is significant on P2 and on the interactive main task, and \emph{Planning} otherwise. Paired intervals are given for the proprietary models; for the other models the image and text intervals are listed side by side. $^{*}$Four styles pooled. $^{\dagger}$6$\times$6 and 8$\times$8 boards. $^{\ddagger}$Point estimate; the lower bound of the P3 interval is 83.0\%.}
\label{tab:diag}
\end{table*}
}

\newcommand{\FloatTabOutcomes}{%
\begin{table*}[t]
\centering
\footnotesize
\setlength{\tabcolsep}{5.0pt}
\fittab{%
\begin{tabular}{@{}l ccccc c ccccc@{}}
\toprule
 & \multicolumn{5}{c}{\Img} & & \multicolumn{5}{c}{\Txt} \\
\cmidrule(lr){2-6}\cmidrule(l){8-12}
Model & Solved & Unfinished & Deadlock & No push & No valid plan & & Solved & Unfinished & Deadlock & No push & No valid plan \\
\midrule
GPT-6.1 & 139 & 34 & 37 & 8 & 7 & & 149 & 29 & 22 & 10 & 15 \\
Gemini 3.8 Flash & 185 & 23 & 11 & 3 & 3 & & 180 & 23 & 15 & 2 & 5 \\
\midrule
Qwen3-VL-32B-Instruct & 0 & 0 & 0 & 0 & 225 & & 0 & 0 & 0 & 0 & 225 \\
Qwen3-VL-8B-Thinking & 3 & 68 & 46 & 85 & 23 & & 20 & 71 & 82 & 30 & 22 \\
Qwen3-VL-8B-Instruct & 2 & 45 & 57 & 117 & 4 & & 0 & 47 & 60 & 109 & 9 \\
InternVL3.5-8B & 0 & 15 & 11 & 27 & 172 & & 0 & 42 & 43 & 37 & 103 \\
MiniCPM-V 4.6 & 0 & 17 & 12 & 61 & 135 & & 0 & 16 & 5 & 200 & 4 \\
\bottomrule
\end{tabular}
}
\caption{\textbf{How single-turn plans end (number of levels out of 225).} The counts behind the outcome figure of the paper. \emph{Unfinished}: the moves run out first; \emph{Deadlock}: a push after which the solver proves the level unsolvable; \emph{No push}: no box is ever moved; \emph{No valid plan}: no parsable action sequence.}
\label{tab:outcomes}
\end{table*}
}

\newcommand{\FloatTabPType}{%
\begin{table*}[t]
\centering
\footnotesize
\setlength{\tabcolsep}{3.6pt}
\fittab{%
\begin{tabular}{@{}l ccccc c ccccc c cc@{}}
\toprule
 & \multicolumn{5}{c}{\Img: accuracy by true outcome} & & \multicolumn{5}{c}{\Txt: accuracy by true outcome} & & \multicolumn{2}{c}{Predicts blocked} \\
\cmidrule(lr){2-6}\cmidrule(lr){8-12}\cmidrule(l){14-15}
Model & Walk & Wall & Push & Box--wall & Box--box & & Walk & Wall & Push & Box--wall & Box--box & & \Img & \Txt \\
\midrule
GPT-6.1 & \best{100.0} & \best{100.0} & \best{100.0} & 96.3 & \best{100.0} & & 100.0 & 94.1 & 97.1 & \best{100.0} & \best{100.0} & & 57.7 & 57.7 \\
Gemini 3.8 Flash & \best{100.0} & \best{100.0} & \best{100.0} & \best{100.0} & \best{100.0} & & 100.0 & \best{100.0} & \best{100.0} & \best{100.0} & \best{100.0} & & 58.3 & 58.3 \\
\midrule
Qwen3-VL-32B-Instruct & \best{76.7} & 0.0 & 66.7 & 0.0 & 0.0 & & 73.3 & 5.0 & 65.0 & 0.0 & 0.0 & & 0.0 & 2.3 \\
Qwen3-VL-8B-Thinking & \best{76.7} & \best{56.7} & 41.7 & \best{35.0} & \best{40.0} & & 90.0 & \best{93.3} & 73.3 & \best{66.7} & \best{81.7} & & 36.3 & 53.0 \\
Qwen3-VL-8B-Instruct & 13.3 & 0.0 & \best{90.0} & 0.0 & 0.0 & & 13.3 & 0.0 & \best{96.7} & 0.0 & 0.0 & & 0.0 & 0.0 \\
InternVL3.5-8B & \second{26.7} & \second{21.7} & \second{70.0} & 0.0 & 0.0 & & 36.7 & \second{26.7} & 65.0 & \second{8.3} & 13.3 & & 14.3 & 21.0 \\
MiniCPM-V 4.6 & 18.3 & \second{21.7} & 58.3 & 0.0 & 0.0 & & 0.0 & 5.0 & \second{91.7} & 5.0 & 0.0 & & 19.3 & 10.3 \\
\bottomrule
\end{tabular}
}
\caption{\textbf{P3 accuracy (\%) by outcome type.} \emph{Wall}, \emph{Box--wall}, and \emph{Box--box} are the three blocked moves and make up 60\% of the items. \emph{Predicts blocked}: share of answers that name one of these three types. Qwen3-VL-8B-Thinking handles blocked moves from text (67--93\%) and loses 32--42 points on each of them from images. Proprietary models: 6$\times$6 and 8$\times$8 boards. \tablemarks}
\label{tab:p3type}
\end{table*}
}

\newcommand{\FloatFigFtdata}{%
\begin{figure*}[t]
\centering
\includegraphics[width=\linewidth]{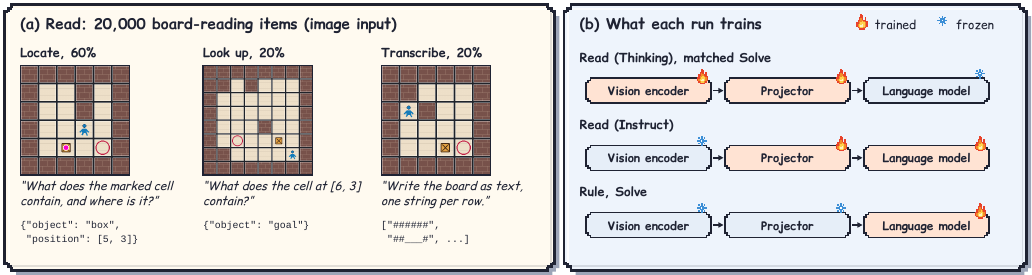}
\caption{\textbf{Training data and trained modules.} (a) The three item types of \readset with one example each (questions abridged). No item contains an action, a rule, or a plan. (b) Modules updated by each run. For \readset on the Thinking model, the language model is frozen, and text inputs do not pass through a trained module.}
\label{fig:ftdata}
\end{figure*}%
}

\newcommand{\FloatTabFtcfg}{%
\begin{table*}[t]
\centering
\footnotesize
\setlength{\tabcolsep}{4.2pt}
\fittab{%
\begin{tabular}{@{}l l l l c l@{}}
\toprule
Run & Start model & Trained modules & Training data & Epochs / steps & Checkpoint \\
\midrule
\readset & Thinking & vision encoder (LoRA) + projector & 20k board-reading items, image & 1 / 625 & last \\
\readset & Instruct & language model (LoRA) + projector & 20k board-reading items, image & 2 / 1,250 & step 1,000 \\
\ruleset & Instruct & language model (LoRA) & 20k single-step items, text & 2 / 1,250 & best val. \\
\readset{} + \ruleset & Instruct + \readset & language model (LoRA) & 8k single-step items, image & 1 / 250 & last \\
\midrule
\solveset & Thinking & language model (LoRA) & 8k levels with solutions (4k image, 4k text) & 1 / 250 & step 100 \\
\solveset, matched & Thinking & vision encoder (LoRA) + projector & 20k levels with solutions, image & 1 / 625 & last \\
\midrule
\readset & Kimi-VL-A3B-Thinking & vision encoder (LoRA) + projector & 20k board-reading items, image & 1 / 625 & last \\
\bottomrule
\end{tabular}
}
\caption{\textbf{Training configurations.} All runs use low-rank adapters of rank 16, learning rate $10^{-4}$, an effective batch size of 32, and bfloat16 weights, and are trained with ms-swift. For reasoning models the loss is computed on the answer only. \emph{Best val.}: highest joint accuracy on 1,000 validation items; step 1,000 and step 100 are the checkpoints with the highest validation accuracy and validation success rate. \emph{\solveset, matched}: the same modules, data size, and schedule as \readset, with solver solutions as the target. Thinking / Instruct: Qwen3-VL-8B.}
\label{tab:ftcfg}
\end{table*}
}

\newcommand{\FloatFigSprites}{%
\begin{figure*}[t]
\centering
\includegraphics[width=\linewidth]{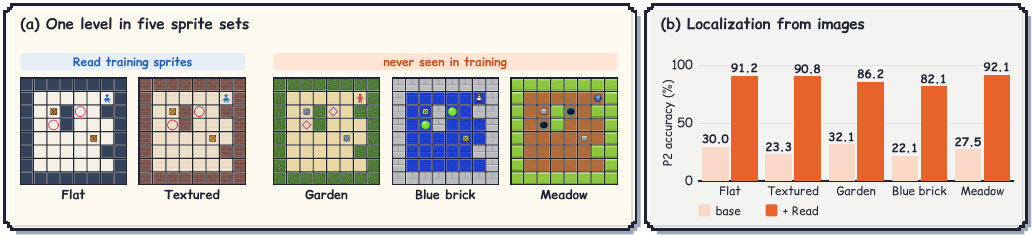}
\caption{\textbf{Localization transfers to sprite sets that were never used in training.} (a) One level drawn with the two sprite sets of the \readset training boards and with three new sets, in which walls, floor, goals, boxes, and the player all look different. (b) Image localization (P2) of \qT before and after \readset on each set.}
\label{fig:sprites}
\end{figure*}%
}

\newcommand{\FloatFigMazeds}{%
\begin{figure*}[t]
\centering
\includegraphics[width=\linewidth]{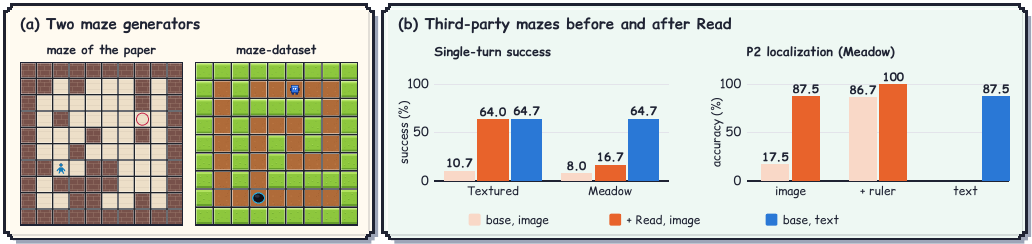}
\caption{\textbf{Diagnosis and transfer on mazes from a third-party generator.} (a) A maze of the paper and a maze from \texttt{maze-dataset}, drawn with the new Meadow sprites. (b) \qT before and after \readset: single-turn success under two sprite sets (left) and localization on the Meadow mazes (right).}
\label{fig:mazeds}
\end{figure*}%
}

\newcommand{\FloatFigMzcases}{%
\begin{figure*}[t]
\centering
\includegraphics[width=\linewidth]{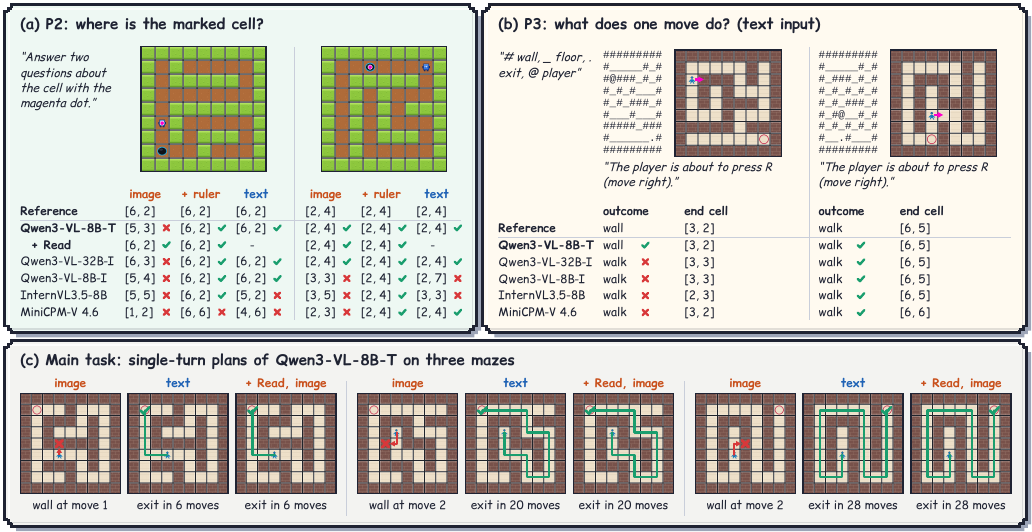}
\caption{\textbf{Real answers on third-party mazes.} (a) Two P2 items on Meadow boards: the position that each model reports for the cell with the magenta dot, from the image, from the image with the coordinate ruler, and from text (+~Read: \qT after \readset). (b) Two P3 items, asked from the text board; the drawn board is added for the reader, and the check or cross shows whether the outcome is correct. (c) Single-turn plans of \qT on three mazes with shortest paths of 6, 20, and 28 moves, from the image, from text, and from the image after \readset. A red cross marks the wall that a plan first walks into, and a green line is a plan that reaches the exit.}
\label{fig:mzcases}
\end{figure*}%
}

\newcommand{\FloatTabMzddiag}{%
\begin{table*}[t]
\centering
\footnotesize
\setlength{\tabcolsep}{4.0pt}
\fittab{%
\begin{tabular}{@{}l c ccc c cc l@{}}
\toprule
 & P3 text & \multicolumn{3}{c}{P2 Localization} & $\dvis$ P2 & \multicolumn{2}{c}{Single-turn} & \\
\cmidrule(lr){3-5}\cmidrule(lr){7-8}
Model & [95\% CI] & \Img & \textcolor{imgc}{+ ruler} & \Txt & [95\% CI] & \Img & \Txt & Bottleneck \\
\midrule
Qwen3-VL-32B-Instruct & 36.0 [28.7, 43.3] & 58.3 & 100.0 & 82.5 & $+$24.2 [12.5, 35.8] & -- & -- & Rules \\
Qwen3-VL-8B-Thinking & 91.3 [86.7, 95.3] & 17.5 & 86.7 & 87.5 & $+$70.0 [60.8, 78.3] & 8.0 & 64.7 & Perception \\
Qwen3-VL-8B-Instruct & 34.0 [26.7, 41.3] & 14.2 & 95.0 & 19.2 & $+$5.0 [$-$4.2, 14.2] & -- & -- & Rules \\
InternVL3.5-8B & 35.3 [28.0, 43.3] & 0.0 & 71.7 & 20.0 & $+$20.0 [13.3, 26.7] & -- & -- & Rules \\
MiniCPM-V 4.6 & 25.3 [18.7, 32.7] & 2.5 & 15.0 & 3.3 & $+$0.8 [$-$3.3, 5.0] & -- & -- & Rules \\
\bottomrule
\end{tabular}
}
\caption{\textbf{The diagnosis on third-party mazes (\%).} 150 mazes of 9$\times$9 cells generated with the \texttt{maze-dataset} library and drawn with a new sprite set (Meadow). P3: 150 text items, rule threshold 66.7\%; P2: 120 items; single-turn main task on all 150 mazes, run for the models that pass the rule check.}
\label{tab:mzddiag}
\end{table*}
}

\newcommand{\FloatFigBlocks}{%
\begin{figure*}[t]
\centering
\includegraphics[width=\linewidth]{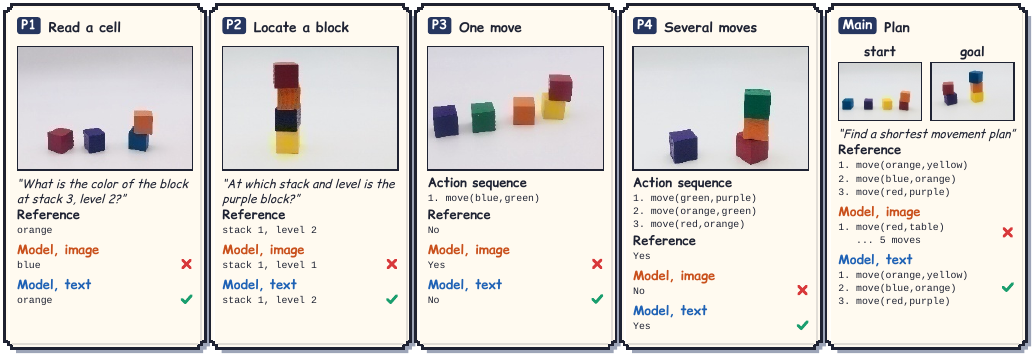}
\caption{\textbf{The five stages on VSP Blocks World, with real items and the answers of \qT.} P1, P3, P4, and the main task use the questions of VSP; P2 asks the reverse of P1, and its wording is paraphrased here. Every item is asked with the image and with the equivalent text state. In all five items the model answers correctly from text and wrongly from the image: it misreads a color (P1), a level (P2), sees a blue block that is not in the scene (P3), rejects a legal sequence (P4), and starts its plan with a block that is not on top of a stack (main task).}
\label{fig:blockcards}
\end{figure*}%
}

\newcommand{\FloatFigBlockcases}{%
\begin{figure*}[t]
\centering
\includegraphics[width=\linewidth]{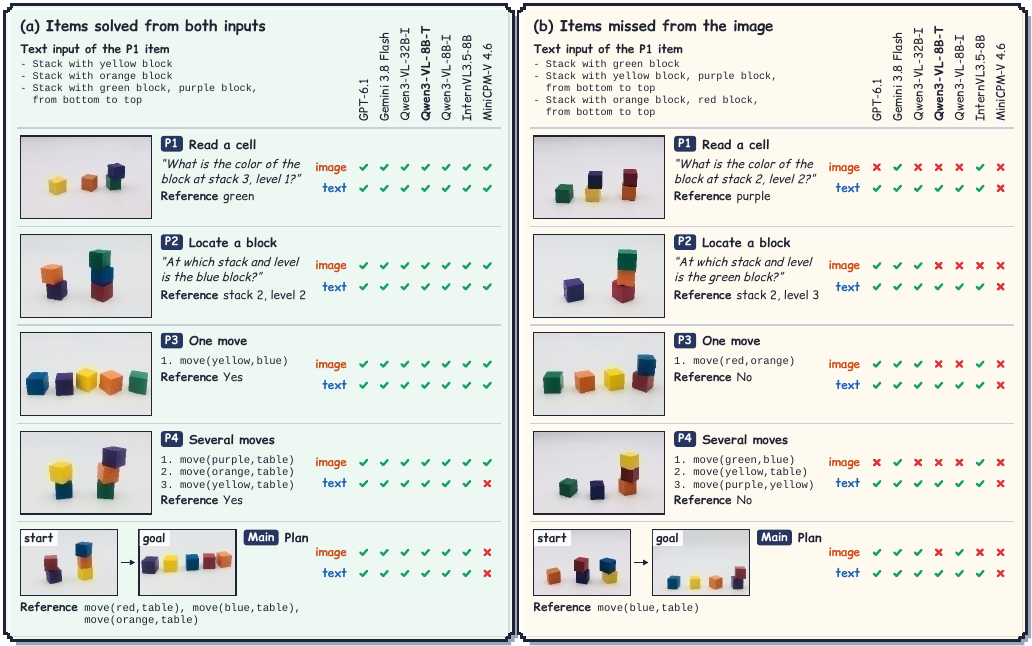}
\caption{\textbf{Real items of VSP Blocks World with the result of every model.} One row per stage. Each item shows the image, the question (P1, P2), the moves to be checked for legality (P3, P4), or the reference plan (main task), and whether each of the seven models answers correctly from the image and from the text state. (a) Items that \qT solves from both inputs. (b) Items that \qT solves from text and misses from the image. The top of each panel shows the text input of its P1 item.}
\label{fig:blockcases}
\end{figure*}%
}

\newcommand{\FloatTabBlocks}{%
\begin{table*}[t]
\centering
\footnotesize
\setlength{\tabcolsep}{4.0pt}
\fittab{%
\begin{tabular}{@{}l cc cc cc cc cc c@{}}
\toprule
 & \multicolumn{2}{c}{P1 Read cell} & \multicolumn{2}{c}{P2 Locate block} & \multicolumn{2}{c}{P3 One move} & \multicolumn{2}{c}{P4 Several moves} & \multicolumn{2}{c}{Main task} & $\dvis$ main \\
\cmidrule(lr){2-3}\cmidrule(lr){4-5}\cmidrule(lr){6-7}\cmidrule(lr){8-9}\cmidrule(lr){10-11}
Model & \Img & \Txt & \Img & \Txt & \Img & \Txt & \Img & \Txt & \Img & \Txt & [95\% CI] \\
\midrule
GPT-6.1 & 96.7 & \best{100.0} & 96.7 & \best{100.0} & 98.0 & 99.0 & 90.7 & 96.7 & 82.0 & 91.5 & $+$9.5 [3.5, 15.5] \\
Gemini 3.8 Flash & \best{99.3} & \best{100.0} & \best{100.0} & \best{100.0} & \best{100.0} & \best{100.0} & \best{100.0} & \best{100.0} & \best{100.0} & \best{99.5} & $-$0.5 [$-$1.5, 0.0] \\
\midrule
Qwen3-VL-32B-Instruct & \best{95.3} & \best{100.0} & \best{96.7} & \best{100.0} & \best{93.0} & \best{99.0} & \best{93.3} & \second{94.0} & \best{63.5} & \best{69.0} & $+$5.5 [$-$2.0, 13.0] \\
Qwen3-VL-8B-Thinking & 87.3 & \best{100.0} & 85.3 & \best{100.0} & 83.0 & \second{98.0} & \second{76.7} & \best{96.0} & \second{32.0} & \second{52.5} & $+$20.5 [14.0, 27.5] \\
Qwen3-VL-8B-Instruct & \second{91.3} & \best{100.0} & 89.3 & \best{100.0} & \second{84.0} & 88.0 & 68.0 & 75.3 & 18.0 & 29.0 & $+$11.0 [5.5, 16.5] \\
InternVL3.5-8B & 79.3 & \second{98.7} & \second{90.0} & \second{98.7} & 82.0 & 87.0 & 60.0 & 59.3 & 6.0 & 5.5 & $-$0.5 [$-$4.5, 4.0] \\
MiniCPM-V 4.6 & 61.3 & 12.0 & 50.7 & 56.0 & 18.0 & 8.0 & 5.3 & 12.7 & 4.5 & 0.0 & $-$4.5 [$-$7.5, $-$2.0] \\
\bottomrule
\end{tabular}
}
\caption{\textbf{The paired protocol on the photo-realistic Blocks World images of VSP (\%).} 750 items per model, each asked with the image and with the equivalent text state. P1: color of the block at a given stack and level (150 items); P2: stack and level of a given block (150); P3: legality of one move (100); P4: legality of a sequence of 3, 5, or 7 moves (150); main task: shortest plan of 1, 3, 5, or 7 moves, replayed by the official evaluator (200). $\dvis$: visual cost (text $-$ image) with its paired 95\% interval. \tablemarks}
\label{tab:blocks}
\end{table*}
}

\newcommand{\FloatFigGalleries}{%
\begin{figure*}[p]
\centering
\includegraphics[width=0.97\linewidth]{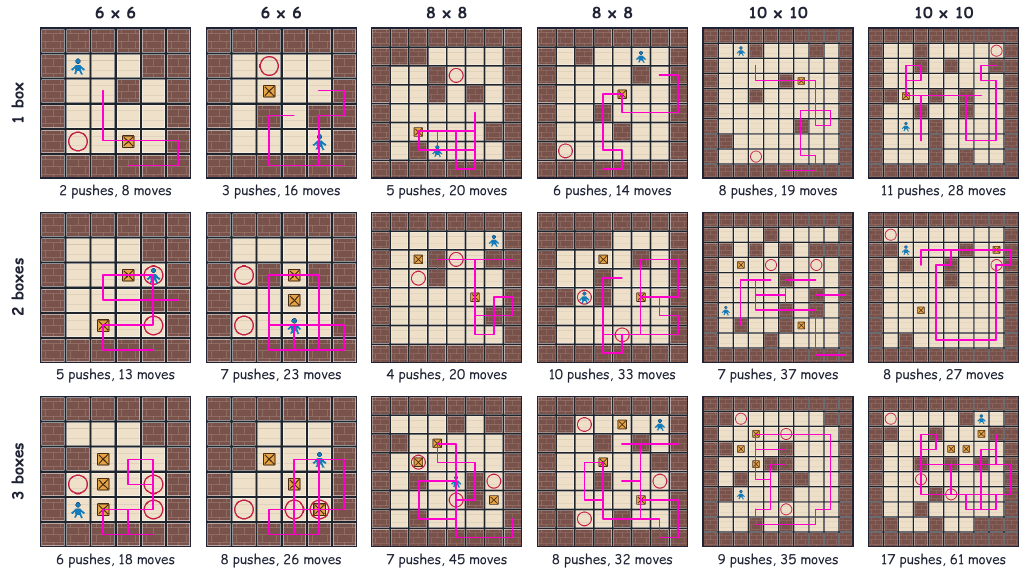}
\caption{\textbf{Sokoban levels of the main task.} Two levels for each board size and number of boxes, in the main style, with the reference plan in magenta. Below each level: pushes and moves of the reference plan.}
\label{fig:gallery-soko}
\vspace{10pt}
\includegraphics[width=0.97\linewidth]{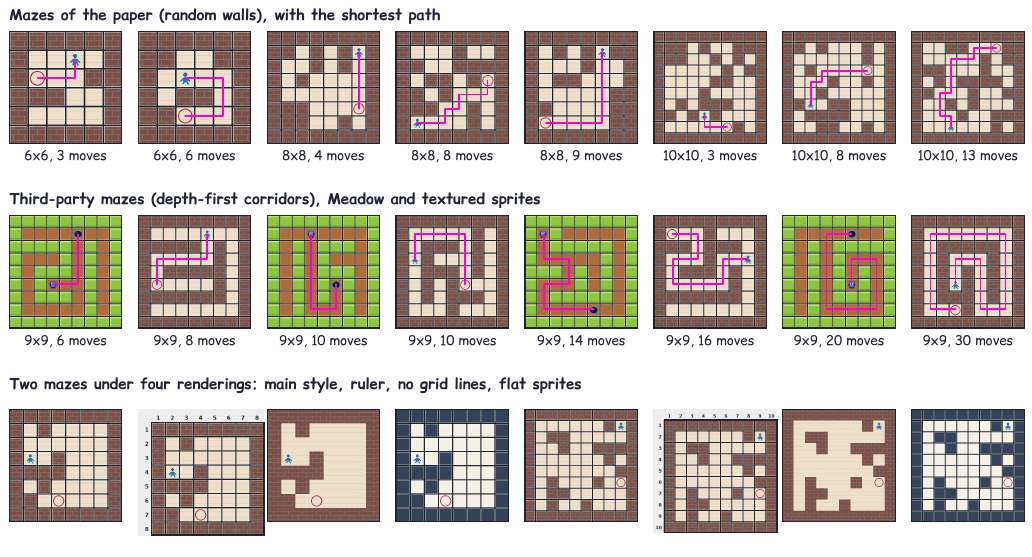}
\caption{\textbf{Mazes.} Top: mazes of the paper. Middle: third-party mazes from \texttt{maze-dataset}, drawn with the Meadow sprites and with the textured sprites of the paper. Shortest paths are in magenta. Bottom: two mazes of the paper, each in the main style, with the coordinate ruler, without grid lines, and with flat sprites.}
\label{fig:gallery-maze}
\end{figure*}%
}

\newcommand{\FloatFigReplay}{%
\begin{figure*}[t]
\centering
\includegraphics[width=\linewidth]{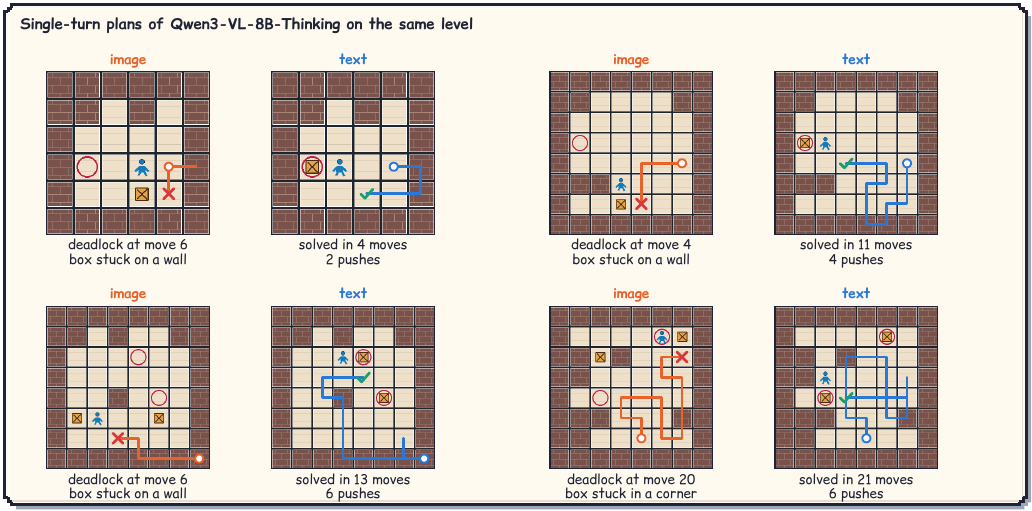}
\caption{\textbf{Replays of real single-turn plans of \qT on the same level.} Each pair shows one level solved from text and failed from the image. Circle: start of the player; line: path of the player; the board shows the final position.}
\label{fig:replay}
\end{figure*}%
}

\newcommand{\FloatFigIareplay}{%
\begin{figure*}[t]
\centering
\includegraphics[width=\linewidth]{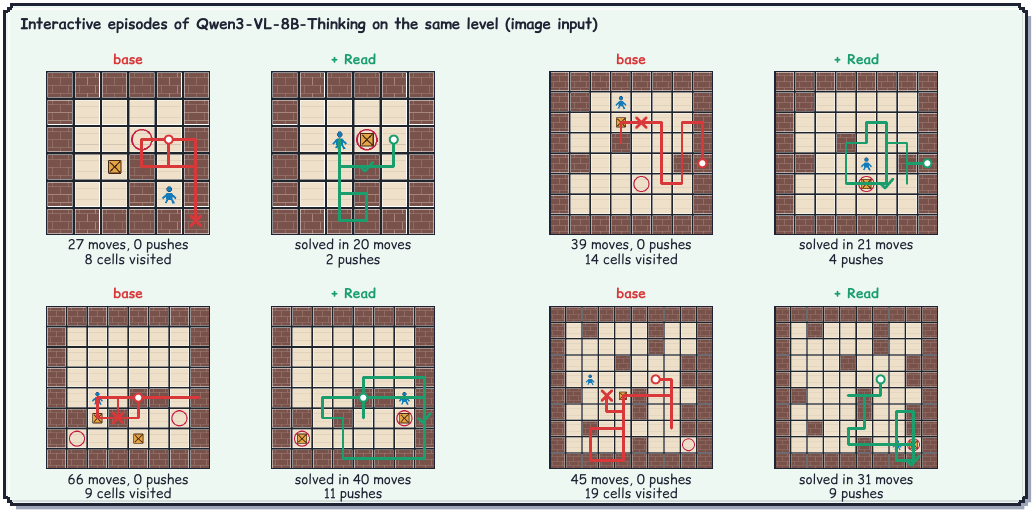}
\caption{\textbf{Replays of real interactive episodes before and after \readset (image input, $K{=}3$).} Each pair shows one level played by the base model and by the model trained on \readset. Circle: start of the player; line: path of the player.}
\label{fig:iareplay}
\end{figure*}%
}

\newcommand{\FloatFigPromptCtrl}{%
\begin{figure*}[t]
\begin{minipage}[t]{0.495\linewidth}\vspace{0pt}
\begin{promptbox}[equal height group=ctrl]{Control for P3: the two cells ahead are stated in words}
\slot{variant 1: two cells, no board} \slot{rules}
The board is not shown. Instead, you are told what the two cells in front of the player contain.
The player is about to press L (move left).
The cell directly in front of the player in that direction contains a box.
The next cell in the same direction contains a box.
Answer one question.
"type": what this action does. Choose one: \slot{the five definitions}
Return ONLY a JSON object, with no explanation, in this format: \{"type": "<walk|wall|push|box\_wall|box\_box>"\}\\[5pt]
\slot{variant 2: board + two cells} \slot{P3 text prompt of \cref{fig:prompt-pair}, with the action sentence extended}
The player is about to press L (move left). The cell directly in front of the player in that direction contains a box, and the next cell in the same direction contains a box. Answer three questions. \slot{as in P3}
\end{promptbox}
\end{minipage}\hfill
\begin{minipage}[t]{0.495\linewidth}\vspace{0pt}
\begin{promptbox}[equal height group=ctrl]{Control for P1: the mark is written inside the board}
\slot{benchmark: mark layer}
The mark layer below has the same shape as the board: X marks one cell, and - means no mark.
\slot{board} \slot{mark layer}
Answer two questions about the cell marked X in the mark layer.\\[5pt]
\slot{control: inline mark}
One cell of the board is written in square brackets: [c] is that cell, and c is what the cell contains. The brackets take no space: every other character is one cell.\\
Board:\\
\#\#\#\#\#\#\\
\#\_[\_].\_\#\\
\#@\$\_\_\#\\
\#\_\#\_\_\#\\
\#\_\_\_\#\#\\
\#\#\#\#\#\#\\
Answer two questions about the cell written in square brackets.
\end{promptbox}
\end{minipage}
\caption{\textbf{Wording of the two control prompts.} Left: the two variants of the P3 control of \cref{sec:p3ctrl}. A cell is described as a wall, an empty floor cell, an empty goal cell, a box, a box standing on a goal, or outside the board. Right: the P1 control of \cref{sec:p1fmt}, next to the benchmark wording that it replaces; the rest of the P1 + P2 prompt is unchanged.}
\label{fig:prompt-ctrl}
\end{figure*}%
}

\noindent This document has three parts.
\Cref{sec:construction,sec:prompts,sec:suppsetup,sec:results,sec:failures,sec:ftdetails,sec:transfer} give the material that the paper points to by letter: how \bench is built, the prompts, the evaluation settings, all results with confidence intervals, the failure breakdowns, and the training details.
\Cref{sec:mazeds,sec:photos} repeat the diagnosis on data that we did not create: mazes from a third-party generator and the photo-realistic Blocks World images of VSP~\cite{vsp2025}.
\Cref{sec:gallery} is a gallery of levels.
\Cref{tab:pointers} maps every pointer in the paper to a table or figure of this document.

\para{Evidence added in this document.}
\begin{itemize}[leftmargin=*,itemsep=1pt,topsep=2pt]
\item \textbf{Third-party mazes} (\cref{sec:mazeds}). On 150 mazes from the \texttt{maze-dataset} library~\cite{mazedataset}, the five open-source models are assigned the same bottleneck as in the paper, and \readset raises the single-turn success of \qT from 10.7\% to 64.0\%, the level it reaches from text (64.7\%).
\item \textbf{Same budget, different target} (\cref{sec:ft-matched}). With the trained modules, the data size, and the schedule held fixed, board-reading supervision brings image localization to 90.8\% and maze solving to 67.1\%; with solver solutions as the target the same budget gives 2.9\% and 11.1\%.
\item \textbf{Unseen sprites} (\cref{sec:sprites}). On three sprite sets that never appear in training, image localization after \readset reaches 82.1--92.1\% (90.8\% on the training sprites).
\item \textbf{A second model family} (\cref{sec:ft-second}). For Kimi-VL-A3B-Thinking~\cite{kimivl} the criteria predict that \readset will leave solving unchanged. Its localization rises from 16.7\% to 92.1\%, and its interactive success stays at the level of its text input, as predicted.
\item \textbf{Photo-realistic images} (\cref{sec:photos}). The paired image/text protocol runs unchanged on the Blocks World scenario of VSP: \qT pays a significant visual cost at all five stages, from 12.7 to 20.5 points.
\end{itemize}

\para{Cases with real answers.} Model answers on single items are shown for all three games: Sokoban plans in \cref{fig:replay,fig:iareplay}, third-party mazes in \cref{fig:mzcases}, and Blocks World in \cref{fig:blockcards,fig:blockcases}.

\begin{table}[h]
\centering
\footnotesize
\setlength{\tabcolsep}{3.0pt}
\renewcommand{\arraystretch}{1.06}
\fittab{%
\begin{tabular}{@{}l p{0.47\linewidth} l@{}}
\toprule
Paper & Content & Here \\
\midrule
Sec.~3.2 & Visual styles, move budgets, termination rules & \cref{sec:render,sec:termination}; \cref{fig:views,fig:endings} \\
Sec.~4.1 & Prompts & \cref{sec:prompts}; \cref{fig:prompt-pair,fig:prompt-tasks} \\
Sec.~4.1 & Decoding settings, item counts & \cref{tab:models,tab:stats} \\
Sec.~4.2 & Breakdown of plan outcomes & \cref{tab:outcomes,tab:grid,tab:iaend}; \cref{fig:replay} \\
Sec.~4.3 & Confusion matrices & \cref{fig:conf}; \cref{tab:p3type} \\
Sec.~4.3 & Control with the two cells ahead & \cref{tab:p3ctrl}; \cref{fig:prompt-ctrl} \\
Sec.~4.4 & Ruler closes 48--82\% of the gap & \cref{tab:rulermain} \\
Sec.~4.4 & Ten input conditions; $K{=}5, 7$ & \cref{tab:conditions,tab:ksteps} \\
Sec.~4.6 & Diagnostic quantities with intervals & \cref{tab:diag}; \cref{fig:diag} \\
Sec.~5.1 & Training configurations & \cref{tab:ftcfg}; \cref{fig:ftdata} \\
Sec.~5.2 & 78\% $\rightarrow$ 6\%; 10$\times$10 boards; \solveset curve & \cref{tab:chain,tab:ftsize}; \cref{fig:solve} \\
Sec.~5.3 & Longest mazes; text score 52.0\% & \cref{tab:ftsize,tab:mazeft} \\
Sec.~5.3 & VSP mazes, general benchmarks & \cref{tab:scope} \\
\bottomrule
\end{tabular}}
\caption{\textbf{Where to find what the paper points to.} Left: section of the paper; right: the section, table, or figure of this document that contains the material.}
\label{tab:pointers}
\end{table}

\section{Benchmark Construction}
\label{sec:construction}

Sec.~3 of the paper describes \bench in brief.
This section gives the full procedure in the order of \cref{fig:pipeline}: a level is generated, solved, rendered as two equivalent inputs, and used to replay model plans.
\FloatFigPipeline
\FloatFigAtlas
\FloatFigViews
\FloatFigEndings

\subsection{Level Generation and Reference Plans}
\label{sec:generation}

\para{Rooms.} A board of size $n$ is an $n{\times}n$ grid whose outer ring is wall, so a $6{\times}6$ board has a $4{\times}4$ interior.
Interior walls are sampled at a fixed ratio, and a room is kept if all of its floor cells are connected.

\para{Reverse generation.} Goals are sampled on floor cells and a box is placed on every goal, which is a solved state.
A breadth-first search then applies \emph{pulls}, the reverse of pushes, and records every state it reaches together with its number of pulls.
One state within the requested range of pulls becomes the start of the level, and the player is placed on a cell it can reach in that state (\cref{fig:pipeline}a).
Every level is therefore solvable, and the range of pulls controls its difficulty.

\para{Reference plan.} A breadth-first solver over pushes finds a solution with the minimum number of pushes; between two pushes the player takes a shortest walk (\cref{fig:pipeline}b).
A level is kept if this minimum lies in the range set for its task (for the main task, between twice the number of boxes and 20), and the reference plan is replayed as a check.
The solver skips states in which a box rests on a cell from which no goal can be reached.
The length $L$ of this plan defines the move budget of the level (\cref{sec:termination}), and the same solver decides during replay whether a position can still be solved.

\para{Selection and balance.} A level is identified by a hash of its wall layout, and each task keeps one level per layout.
The main task has 225 levels, 25 for each combination of board size (6, 8, 10) and number of boxes (1, 2, 3).
Their reference plans have 2 to 67 moves (median 19) and need 2 to 19 pushes (median 6).
\Cref{fig:atlas}a shows one level per cell, and \cref{fig:atlas}b the two difficulty distributions.
The probes draw their boards from the same generator, and P1 and P2 also include $12{\times}12$ boards (item counts in \cref{tab:stats}).

\para{Maze.} The maze keeps the rooms and removes the boxes.
Walls are sampled in the same way, the player and one exit are placed on floor cells, and a board is kept if the exit can be reached.
The 225 mazes of the main task are balanced over board size (6, 8, 10) and over three groups of shortest-path length (short, medium, long), with 25 mazes per cell and shortest paths of 2 to 17 moves (\cref{fig:atlas}c).
Every move in a maze can be undone, so a maze has no deadlock, and planning is easier than in Sokoban (Sec.~4.5 of the paper).

\FloatFigCardsA
\FloatFigCardsB
\FloatFigMazecards
\subsection{Rendering the Two Inputs}
\label{sec:render}

\para{Image input.} Every board is drawn with square tiles on a 768$\times$768 canvas with a gray background.
P1 and P2 use four visual styles: two sprite sets (flat colors and textured tiles), each with and without grid lines (\cref{fig:views}a).
The textured set with grid lines is the \emph{main style}; P3, P4, and the main task use it.
Two overlays can be added.
The \emph{coordinate ruler} prints row numbers to the left of the board and column numbers above it and leaves the cells unchanged.
The \emph{cell mark} is a magenta dot with a white border on the cell that P1 and P2 ask about.
In the prompt, one sentence describes the sprites of the style in use (\cref{sec:prompts}).

\para{Text input.} The same board is written as one line of characters per row (\cref{fig:views}b).
Floor is written as an underscore, so every cell is one visible character.
The marked cell is given in a second grid of the same shape, the \emph{mark layer}, and with the ruler, row and column numbers are added to both grids.
The image and the text thus carry the same information: every cell, the marked cell, and the optional coordinates.

\para{Ten input conditions.} For P1 and P2, the four styles with and without the ruler give eight image conditions, and text with and without the ruler gives two more.
\Cref{tab:conditions} reports all ten.

\FloatFigPromptPair
\FloatFigPromptTasks
\subsection{Move Budgets and Termination Rules}
\label{sec:termination}

\para{Single-turn.} The model returns its whole plan in one response.
The plan is truncated at the move budget of $3L$ moves, where $L$ is the length of the reference plan, and is replayed by the engine.
Any plan that puts every box on a goal within the budget counts as solved; the reference plan is one of many.
After every push, the solver checks whether the position can still be solved.
A replay ends in one of five ways (\cref{fig:endings}a): \emph{solved}; \emph{deadlock}, when a push leads to a position that the solver proves unsolvable; \emph{unfinished}, when the moves run out after at least one push; \emph{no push}, when no box is ever moved; and \emph{no valid plan}, when the response contains no parsable action sequence.
Deadlocks are further labeled by their cause: a box in a corner, a box against a wall on a cell from which no goal can be reached, boxes that block each other so that none can move, and positions that the full search alone proves unsolvable.

\para{Interactive.} A level is played in turns (\cref{fig:endings}b).
In each turn the model sees the current board and the list of moves carried out so far and returns 1 to $K$ moves, with $K{=}3$ unless stated otherwise.
The engine carries them out, and the next turn shows the new board.
An episode ends when the level is solved, when the solver proves a deadlock, when $3L$ moves have been made, or after $\lceil 3L/K \rceil + 3$ turns.
A turn without a valid answer carries out no move and counts toward the turn limit.

\para{Maze.} The maze uses the same budget of $3L$ moves, with $L$ the shortest-path length, and the same turn limit.
A maze plan ends with the exit reached, with the moves used up, or without a valid plan.

\subsection{Verification}
\label{sec:verification}

All answers are computed by the engine: the content and position of every cell, the outcome of every move, the position after every step, and the solvability of every position.
For each task, the reference answers are passed through the scorer and score 100\%, so every lost point is a model error.
Items are balanced over board size, number of boxes, and answer type (\cref{tab:stats}).

\begin{table}[t]
\centering
\footnotesize
\setlength{\tabcolsep}{3.6pt}
\fittab{%
\begin{tabular}{@{}l l l r@{}}
\toprule
Model & Answer & Decoding & Max.\ new tokens \\
\midrule
GPT-6.1 & reasoning & default & -- \\
Gemini 3.8 Flash & reasoning & default & -- \\
\midrule
Qwen3-VL-32B-I & direct & greedy & 1,024 \\
Qwen3-VL-8B-T & reasoning & $T{=}1.0$, $p{=}0.95$, $k{=}20$ & 16,384 / 32,768 \\
Qwen3-VL-8B-I & direct & greedy & 1,024 \\
InternVL3.5-8B & reasoning$^{\S}$ & $T{=}0.6$, $p{=}0.95$, $k{=}50$ & 16,384 / 32,768 \\
MiniCPM-V 4.6 & direct & greedy & 1,024 \\
\bottomrule
\end{tabular}
}
\caption{\textbf{Models and decoding.} -T: Thinking; -I: Instruct. $T$: temperature; $p$: top-$p$; $k$: top-$k$. The two token limits are for the probes and for the main task (the text version of P4 also uses 32,768). $^{\S}$For P1 and P2, InternVL3.5-8B answers directly (greedy, 1,024 tokens).}
\label{tab:models}
\end{table}

\begin{table*}[t]
\centering
\footnotesize
\setlength{\tabcolsep}{5pt}
\fittab{%
\begin{tabular}{@{}l l r r l r@{}}
\toprule
Task & Boards & Items & Sizes & Input conditions & Queries \\
\midrule
\multicolumn{6}{@{}l}{\textit{Sokoban}} \\
P1 + P2 (one query) & 120 & 240 & 6, 8, 10, 12 & 4 styles $\times$ ruler on/off; text $\times$ ruler on/off & 2,400 \\
P3 Single-step & 47 & 300 & 6, 8, 10 & main style; text & 600 \\
P4 Tracking & 120 & 480 & 6, 8, 10 & main style + ruler; text + ruler & 960 \\
Main task, single-turn & 225 & 225 & 6, 8, 10 & 4 styles; text & 1,125 \\
Main task, interactive & 225 & 225 & 6, 8, 10 & main style; text; $K \in \{3, 5, 7\}$ & episodes \\
\midrule
\multicolumn{6}{@{}l}{\textit{Maze}} \\
P1 + P2 (one query) & 120 & 240 & 6, 8, 10, 12 & 4 styles $\times$ ruler on/off; text $\times$ ruler on/off & 2,400 \\
P3 Single-step & -- & 300 & 6, 8, 10 & main style; text & 600 \\
P4 Tracking & 120 & 480 & 6, 8, 10 & main style + ruler; text + ruler & 960 \\
Main task, single-turn & 225 & 225 & 6, 8, 10 & 4 styles; text & 1,125 \\
Main task, interactive & 225 & 225 & 6, 8, 10 & main style; text; $K{=}3$ & episodes \\
\bottomrule
\end{tabular}
}
\caption{\textbf{Composition of \bench.} \emph{Items}: distinct questions; \emph{Queries}: items $\times$ input conditions for one model. P3 has 60 items per outcome type on Sokoban (five types) and 100 per type on the maze (three types); P4 has 120 items for each length $H \in \{5, 10, 15, 20\}$. The main style is the textured sprite set with grid lines.}
\label{tab:stats}
\end{table*}
\begin{table*}[t]
\centering
\footnotesize
\setlength{\tabcolsep}{4.2pt}
\fittab{%
\begin{tabular}{@{}l cccc cccc cc@{}}
\toprule
 & \multicolumn{4}{c}{\Img, no ruler} & \multicolumn{4}{c}{\Img, with ruler} & \multicolumn{2}{c}{\Txt} \\
\cmidrule(lr){2-5}\cmidrule(lr){6-9}\cmidrule(l){10-11}
Model & Flat+G & Flat & Tex+G & Tex & Flat+G & Flat & Tex+G & Tex & no ruler & ruler \\
\midrule
\multicolumn{11}{@{}l}{\textit{P1 Recognition}} \\
GPT-6.1 & \multicolumn{4}{c}{94.3 (pooled)} & \multicolumn{4}{c}{92.8 (pooled)} & 98.3 & 95.4 \\
Gemini 3.8 Flash & \multicolumn{4}{c}{100.0 (pooled)} & \multicolumn{4}{c}{99.9 (pooled)} & \best{99.6} & \best{99.2} \\
\cmidrule(l){1-11}
Qwen3-VL-32B-Instruct & \best{57.5} & \best{63.7} & \best{62.1} & \best{71.2} & \best{60.4} & \best{67.5} & \best{64.6} & \best{68.8} & \second{27.5} & \second{53.8} \\
Qwen3-VL-8B-Thinking & \second{48.8} & \second{51.7} & 44.2 & \second{49.2} & \second{49.6} & \second{55.4} & \second{50.4} & \second{59.2} & \best{75.8} & \best{84.6} \\
Qwen3-VL-8B-Instruct & 27.1 & 29.2 & 24.2 & 31.7 & 29.6 & 26.7 & 32.1 & 34.6 & 24.2 & 44.6 \\
InternVL3.5-8B & 20.0 & 20.0 & 20.0 & 20.0 & 20.0 & 20.0 & 20.0 & 20.0 & 26.7 & 35.0 \\
MiniCPM-V 4.6 & 38.8 & 35.4 & \second{45.4} & 40.4 & 40.0 & 37.1 & 46.2 & 42.5 & 19.2 & 25.4 \\
\midrule
\multicolumn{11}{@{}l}{\textit{P2 Localization}} \\
GPT-6.1 & \multicolumn{4}{c}{89.7 (pooled)} & \multicolumn{4}{c}{97.9 (pooled)} & 99.2 & \best{100.0} \\
Gemini 3.8 Flash & \multicolumn{4}{c}{97.0 (pooled)} & \multicolumn{4}{c}{99.9 (pooled)} & \best{100.0} & \best{100.0} \\
\cmidrule(l){1-11}
Qwen3-VL-32B-Instruct & \second{27.9} & \best{18.8} & \best{29.6} & \best{28.7} & \best{87.1} & \best{89.2} & \best{88.8} & \best{88.3} & \second{81.2} & \best{89.6} \\
Qwen3-VL-8B-Thinking & \best{30.0} & \second{17.9} & \second{23.3} & \second{13.8} & 76.2 & \second{60.8} & 75.4 & \second{67.9} & \best{83.3} & \best{89.6} \\
Qwen3-VL-8B-Instruct & 10.4 & 8.3 & 11.2 & 7.1 & \second{85.4} & 60.0 & \second{82.9} & 62.9 & 24.6 & \second{54.2} \\
InternVL3.5-8B & 7.1 & 7.1 & 5.8 & 6.7 & 35.8 & 31.7 & 28.7 & 32.9 & 24.6 & 44.2 \\
MiniCPM-V 4.6 & 1.7 & 1.7 & 0.8 & 1.7 & 11.7 & 7.9 & 13.3 & 7.5 & 4.2 & 13.8 \\
\bottomrule
\end{tabular}
}
\caption{\textbf{P1 and P2 accuracy (\%) under all ten input conditions.} Flat / Tex: flat-color or textured sprites; +G: with grid lines; Tex+G is the main style of the paper. Each cell has 240 items; proprietary models are pooled over the four styles (960 items). \tablemarks}
\label{tab:conditions}
\end{table*}

\FloatFigFactors
\section{Tasks and Prompts}
\label{sec:prompts}

\Cref{fig:cards1,fig:cards2} show one real item for every Sokoban task with its reference answer and, for the probes, the answers of \qT.
\Cref{fig:mazecards} shows the same tasks on the maze.
All prompts are zero-shot.
They are assembled from a small set of shared sentences, so tasks differ in the question, and the two inputs differ in the sentences that describe the board.

\para{Paired prompts.} \Cref{fig:prompt-pair} prints the complete P3 prompt for both inputs.
The rules, the question, the definitions of the five outcome types, and the answer format are the same word for word.
Three parts differ: the legend (sprites or characters), one clause of the coordinate convention (top row of the board or first line of the text), and the board itself, which is the attached image or a text block.

\para{Task questions.} \Cref{fig:prompt-tasks} lists the question and the answer format of every task and the order in which the shared sentences are combined.
P1 and P2 are asked in one query and scored separately.
P4 includes the ruler for both inputs, so that it measures tracking and leaves indexing to P2.
The maze uses the same templates with its own rules, legend, and outcome types.

\para{Answers.} Every task asks for one JSON object.
For reasoning models the reasoning segment is removed first.
The last parsable JSON object of the response is the answer, and a response without one is scored as wrong.
Positions are \texttt{[row, column]} pairs counted from 1 and include the outer wall.
P3 is scored on the outcome type (Tab.~2 of the paper) and separately on the start and end positions; P4 is scored on the final position and, step by step, for the new-error rate.

\begin{table*}[t]
\centering
\footnotesize
\setlength{\tabcolsep}{3.6pt}
\fittab{%
\begin{tabular}{@{}l cccc cccc cccc c@{}}
\toprule
 & \multicolumn{4}{c}{\Img} & \multicolumn{4}{c}{\Img{} + ruler} & \multicolumn{4}{c}{\Txt} & One-cell \\
\cmidrule(lr){2-5}\cmidrule(lr){6-9}\cmidrule(lr){10-13}
Model & 6 & 8 & 10 & 12 & 6 & 8 & 10 & 12 & 6 & 8 & 10 & 12 & errors \\
\midrule
GPT-6.1 & 95.0 & 95.8 & 88.3 & 79.6 & 98.3 & 98.8 & 97.9 & 96.7 & \best{100.0} & 98.3 & \best{100.0} & \best{98.3} & -- \\
Gemini 3.8 Flash & \best{100.0} & \best{100.0} & \best{89.2} & \best{90.4} & \best{100.0} & \best{100.0} & \best{100.0} & \best{99.2} & \best{100.0} & \best{100.0} & \best{100.0} & \best{98.3} & -- \\
\midrule
Qwen3-VL-32B-Instruct & \best{42.9} & \best{27.9} & \best{20.0} & \best{14.2} & \second{85.8} & \best{94.6} & \best{87.1} & \best{85.8} & \second{83.3} & \second{85.0} & \second{80.0} & \best{76.7} & 75\% \\
Qwen3-VL-8B-Thinking & \second{40.4} & \second{22.9} & \second{12.9} & \second{8.8} & 82.5 & 67.5 & \second{71.2} & 59.2 & \best{96.7} & \best{90.0} & \best{85.0} & \second{61.7} & 47\% \\
Qwen3-VL-8B-Instruct & 18.3 & 7.1 & 3.8 & 7.9 & \best{87.1} & \second{73.8} & 70.0 & \second{60.4} & 41.7 & 30.0 & 16.7 & 10.0 & 33\% \\
InternVL3.5-8B & 19.2 & 2.9 & 4.2 & 0.4 & 40.0 & 21.7 & 32.1 & 35.4 & 43.3 & 25.0 & 20.0 & 10.0 & 21\% \\
MiniCPM-V 4.6 & 3.3 & 1.7 & 0.8 & 0.0 & 18.3 & 12.5 & 3.3 & 6.2 & 6.7 & 3.3 & 3.3 & 3.3 & 7\% \\
\bottomrule
\end{tabular}
}
\caption{\textbf{P2 accuracy (\%) by board size (number of cells per side).} Image and Image + ruler: four styles pooled, 240 items per cell; Text: 60 items per cell. \emph{One-cell errors}: share of the wrong image answers (main style, no ruler) that name a cell directly above, below, left, or right of the target. \tablemarks}
\label{tab:p2size}
\end{table*}
\begin{table*}[t]
\centering
\scriptsize
\setlength{\tabcolsep}{3.4pt}
\fittab{%
\begin{tabular}{@{}l c cc c cc c l@{}}
\toprule
 & P3 text & \multicolumn{3}{c}{P2 Localization} & \multicolumn{3}{c}{Interactive main task} & \\
\cmidrule(lr){3-5}\cmidrule(lr){6-8}
Model & [95\% CI] & \Img & \Txt & $\dvis$ & \Img & \Txt & $\dvis$ [95\% CI] & Bottleneck \\
\midrule
\multicolumn{9}{@{}l}{\textit{Sokoban (rule threshold 40\%)}} \\
GPT-6.1 & 98.2 [96.1, 100.0]$^{\dagger}$ & 89.7 [87.7, 91.6]$^{*}$ & 99.2 [97.9, 100.0] & $+$9.5 & 81.3 [76.4, 86.2] & 85.3 [80.4, 89.8] & $+$4.0 [$-$2.2, 10.2] & Planning \\
Gemini 3.8 Flash & 100.0 [100.0, 100.0]$^{\dagger}$ & 97.0 [95.5, 98.2]$^{*}$ & 100.0 [100.0, 100.0] & $+$3.0 & 87.1 [82.7, 91.6] & 84.0 [79.1, 88.9] & $-$3.1 [$-$8.0, 1.8] & Planning \\
\cmidrule(l){1-9}
Qwen3-VL-32B-Instruct & 28.7 [24.0, 32.8] & 29.6 [23.3, 36.2] & 81.2 [76.2, 85.8] & $+$51.6 & 0.4 [0.0, 1.3] & 0.4 [0.0, 1.3] & $+$0.0 & Rules \\
Qwen3-VL-8B-Thinking & 81.0 [76.5, 85.3] & 23.3 [17.5, 29.6] & 83.3 [78.3, 87.9] & $+$60.0 & 5.8 [3.1, 8.9] & 32.4 [26.2, 38.7] & $+$26.6 & Perception \\
Qwen3-VL-8B-Instruct & 22.0 [19.3, 24.4] & 11.2 [7.1, 15.8] & 24.6 [19.2, 30.4] & $+$13.4 & 0.0 [0.0, 0.0] & 0.0 [0.0, 0.0] & $+$0.0 & Rules \\
InternVL3.5-8B & 30.0 [24.3, 36.2] & 5.8 [2.9, 9.2] & 24.6 [19.2, 30.0] & $+$18.8 & 0.4 [0.0, 1.3] & 5.3 [2.7, 8.4] & $+$4.9 & Rules \\
MiniCPM-V 4.6 & 20.3 [17.5, 23.0] & 0.8 [0.0, 2.1] & 4.2 [2.1, 6.7] & $+$3.4 & 0.0 [0.0, 0.0] & 0.0 [0.0, 0.0] & $+$0.0 & Rules \\
\midrule
\multicolumn{9}{@{}l}{\textit{Maze (rule threshold 66.7\%)}} \\
GPT-6.1 & 87.0$^{\ddagger}$ & 97.5$^{\ddagger}$ & 97.1$^{\ddagger}$ & $-$0.4 & 99.6 [98.7, 100.0] & 96.4 [93.8, 98.7] & $-$3.2 & Planning \\
Gemini 3.8 Flash & 100.0 [100.0, 100.0] & 98.3 [96.7, 99.6] & 100.0 [100.0, 100.0] & $+$1.7 & 100.0 [100.0, 100.0] & 100.0 [100.0, 100.0] & $+$0.0 & Planning \\
\cmidrule(l){1-9}
Qwen3-VL-32B-Instruct & 39.7 [34.0, 45.3] & 34.2 [27.9, 40.0] & 82.5 [77.5, 87.1] & $+$48.3 & 32.0 [26.2, 38.2] & 14.2 [9.8, 18.7] & $-$17.8 & Rules \\
Qwen3-VL-8B-Thinking & 88.0 [84.3, 91.7] & 27.1 [21.7, 32.9] & 77.5 [72.1, 82.5] & $+$50.4 & 46.7 [40.4, 53.3] & 84.4 [79.6, 88.9] & $+$37.7 & Perception \\
Qwen3-VL-8B-Instruct & 35.7 [30.3, 41.0] & 11.7 [7.5, 15.8] & 26.2 [20.8, 32.1] & $+$14.5 & 9.3 [5.8, 13.3] & 5.8 [3.1, 8.9] & $-$3.5 & Rules \\
InternVL3.5-8B & 39.0 [33.7, 44.7] & 9.2 [5.8, 12.9] & 31.2 [25.4, 37.1] & $+$22.0 & 8.9 [5.3, 12.9] & 65.8 [59.6, 71.6] & $+$56.9 & Rules \\
MiniCPM-V 4.6 & 20.0 [15.3, 24.7] & 2.9 [0.8, 5.0] & 6.2 [3.3, 9.6] & $+$3.3 & 2.7 [0.9, 4.9] & 4.0 [1.8, 6.7] & $+$1.3 & Rules \\
\bottomrule
\end{tabular}
}
\caption{\textbf{Quantities used by the diagnostic criteria, with 95\% confidence intervals.} Accuracy and success rate in \%, visual cost $\dvis$ (text $-$ image) in points. A model is assigned \emph{Rules} if its P3 text interval does not lie above the threshold, \emph{Perception} if it passes this check and its visual cost is significant on P2 and on the interactive main task, and \emph{Planning} otherwise. Paired intervals are given for the proprietary models; for the other models the image and text intervals are listed side by side. $^{*}$Four styles pooled. $^{\dagger}$6$\times$6 and 8$\times$8 boards. $^{\ddagger}$Point estimate; the lower bound of the P3 interval is 83.0\%.}
\label{tab:diag}
\end{table*}

\FloatFigDiag
\section{Models, Decoding, and Item Counts}
\label{sec:suppsetup}

\para{Models and decoding.} \Cref{tab:models} lists the decoding settings.
The open-source models are served with vLLM~\cite{vllm} in bfloat16 and receive every image at its native 768$\times$768 resolution.
Direct-answer models are decoded greedily with up to 1,024 new tokens.
Reasoning models use the sampling settings recommended with their checkpoints and a larger token limit, 16,384 for the probes and 32,768 for the main task, so that the reasoning segment has room to finish.
The proprietary models, GPT-6.1~\cite{gpt61sol} and Gemini 3.8 Flash~\cite{gemini38flash}, are queried through an API with default decoding settings.
All models are evaluated zero-shot with the prompts of \cref{sec:prompts}.

\para{Item counts.} \Cref{tab:stats} lists boards, items, and queries for every task.
For the probes and the single-turn task, one open-source model answers $2{,}400 + 600 + 960 + 1{,}125 = 5{,}085$ Sokoban queries, the number given in Sec.~4.1 of the paper.
The maze adds the same number, and the interactive episodes are evaluated in addition.
For P3 and P4, the proprietary models are evaluated on the first half of the boards in index order (23 of 47 boards for P3, 60 of 120 for P4) with all items on these boards, 163 and 240 items.
Boards are indexed by size, so these items cover $6{\times}6$ and $8{\times}8$ boards, and the items are the same for both inputs.

\para{Confidence intervals.} All intervals are 95\% percentile bootstrap intervals from 10,000 resamples~\cite{bootstrap}.
The resampling unit is the board, which keeps items that share a board together~\cite{clusterboot}.
Differences between two conditions on the same boards (image and text, with and without the ruler, before and after training) are paired: both conditions are computed on the same resampled boards, and the interval is taken over their differences.
A difference is called significant when its interval excludes zero, and a score is above a threshold when its whole interval lies above it.

\para{Random-guess rows of Tab.~2.} P1 and P3 have five answer classes on Sokoban (20\%); on the maze they have four and three (25\% and 33\%).
In P2 and P4 a random cell is correct in fewer than 3\% of cases, since the smallest board has 36 cells.
For the main task we replay uniformly random action sequences of the budget length, 200 per Sokoban level and 2,000 per maze, which gives 0.5\% and 7.4\% in the single-turn setting.

\section{Extended Results}
\label{sec:results}

This section reports the complete numbers behind Secs.~4.3 to 4.6 of the paper.

\subsection{P1 and P2 under All Ten Input Conditions}
\label{sec:conditions}

\Cref{tab:conditions} reports recognition and localization for the eight image conditions and the two text conditions.

\para{The ruler is the factor that matters.} For the five open-source models, the ruler raises image localization in all four styles, by 5.8 to 75.0 points.
\Cref{fig:factors} isolates the three rendering factors with paired differences: the ruler adds 12.5 to 71.7 points, grid lines change accuracy by $-$0.8 to $+$9.6 points, and the sprite set by $-$6.7 to $+$1.7 points.
Printing the coordinates removes most of the difficulty, and changing how the cells look removes little of it.
The difficulty lies in counting rows and columns.

\para{Accuracy falls with board size.} \Cref{tab:p2size} splits localization by board size.
Without the ruler, accuracy is lower on $12{\times}12$ than on $6{\times}6$ boards for all seven models; for \qT it goes from 40.4\% to 8.8\%.
Three quarters of the wrong answers of \qL (75\%) and about half of those of \qT (47\%) name a cell directly next to the target.
Both the lower accuracy on larger boards and the one-cell offsets are signatures of miscounting.

\para{The same factor affects the proprietary models.} GPT-6.1 localizes 95.0\% of the marked cells on $6{\times}6$ boards and 79.6\% on $12{\times}12$ boards, and the ruler brings the latter to 96.7\%.
On maze boards drawn with flat colors and no grid lines, where cell borders are least visible, it localizes 43.3\% of the marked cells, and 88.8\% with the ruler.

\subsection{The Coordinate Ruler on the Main Task}
\label{sec:rulermain}

\begin{table}[t]
\centering
\footnotesize
\setlength{\tabcolsep}{3.2pt}
\fittab{%
\begin{tabular}{@{}l c c c c c@{}}
\toprule
Setting & \Img & \Img{} + ruler & \Txt & Gain from ruler & Gap closed \\
\midrule
Maze, single-turn & 13.8 & 38.7 & 48.9 & \gup{$+$24.9 [18.2, 31.6]} & 71\% \\
Sokoban, single-turn & 1.3 & 7.6 & 8.9 & \gup{$+$6.2 [3.1, 9.8]} & 82\% \\
Sokoban, interactive & 5.8 & 18.7 & 32.4 & \gup{$+$12.9 [8.4, 17.8]} & 48\% \\
\bottomrule
\end{tabular}
}
\caption{\textbf{Coordinate ruler on the main task (Qwen3-VL-8B-Thinking, success rate, \%).} Same 225 boards per setting; the gain is paired by board with a 95\% interval. \emph{Gap closed} $=$ (ruler $-$ image) / (text $-$ image). Gains are shown in red.}
\label{tab:rulermain}
\end{table}

The ruler test of P2 is repeated on the main task for \qT: the same 225 boards are solved from the plain image, from the image with the ruler, and from text (\cref{tab:rulermain}).
The ruler raises success in all three settings, and each paired gain is significant: 24.9 points on the single-turn maze, 6.2 on single-turn Sokoban, and 12.9 on interactive Sokoban.
Measured against the gap between image and text input, the ruler closes 71\%, 82\%, and 48\% of the gap.
No model weights are updated in this test, so it isolates grid indexing as the main source of the visual cost on the main task.
The fine-tuning result of the paper agrees with it: \readset brings interactive success to 20.4\%, and the ruler brings it to 18.7\%.

\subsection{Interactive Solving with $K = 3$, 5, and 7}
\label{sec:ksteps}

\begin{table}[t]
\centering
\footnotesize
\setlength{\tabcolsep}{3.0pt}
\fittab{%
\begin{tabular}{@{}l cc cc cc cc@{}}
\toprule
 & \multicolumn{2}{c}{Single-turn} & \multicolumn{2}{c}{$K{=}3$} & \multicolumn{2}{c}{$K{=}5$} & \multicolumn{2}{c}{$K{=}7$} \\
\cmidrule(lr){2-3}\cmidrule(lr){4-5}\cmidrule(lr){6-7}\cmidrule(l){8-9}
Model & \Img & \Txt & \Img & \Txt & \Img & \Txt & \Img & \Txt \\
\midrule
GPT-6.1 & 61.8 & 66.2 & 81.3 & \best{85.3} & 81.8 & 81.3 & 85.3 & \best{88.0} \\
Gemini 3.8 Flash & \best{82.2} & \best{80.0} & \best{87.1} & 84.0 & \best{91.6} & \best{88.4} & \best{88.0} & 85.8 \\
\midrule
Qwen3-VL-32B-Instruct & 0.0 & 0.0 & \second{0.4} & 0.4 & \second{0.9} & 0.0 & 0.9 & \second{1.3} \\
Qwen3-VL-8B-Thinking & \best{1.3} & \best{8.9} & \best{5.8} & \best{32.4} & \best{6.2} & \best{28.0} & \best{8.4} & \best{25.8} \\
Qwen3-VL-8B-Instruct & 0.9 & 0.0 & 0.0 & 0.0 & 0.4 & 0.9 & 0.0 & 0.4 \\
InternVL3.5-8B & 0.0 & 0.0 & \second{0.4} & \second{5.3} & -- & -- & -- & -- \\
MiniCPM-V 4.6 & 0.0 & 0.0 & 0.0 & 0.0 & 0.0 & 0.0 & 0.0 & 0.0 \\
\bottomrule
\end{tabular}
}
\caption{\textbf{Main-task success (\%) with up to $K$ moves per turn.} 225 Sokoban levels per cell. InternVL3.5-8B is evaluated with $K{=}3$. \tablemarks}
\label{tab:ksteps}
\end{table}

\Cref{tab:ksteps} repeats the interactive main task with up to 5 and up to 7 moves per turn.
The diagnosis is the same for every $K$.
The proprietary models solve 81.3--91.6\% of the levels, and their visual cost stays within $\pm$4 points with intervals that include zero.
For \qT the visual cost is 26.6, 21.8, and 17.4 points.
The rule-limited models evaluated with all three values stay at or below 1.3\%.

\subsection{P3 Control: Stating the Two Cells Ahead}
\label{sec:p3ctrl}

\begin{table}[t]
\centering
\footnotesize
\setlength{\tabcolsep}{2.6pt}
\fittab{%
\begin{tabular}{@{}l c c c c@{}}
\toprule
 & Board & Two cells, & Board + & Blocked \\
Model & (P3 text) & no board & two cells & predicted \\
\midrule
Qwen3-VL-32B-I & 28.7 & 98.3 [96.7, 99.7] & 98.3 [96.7, 99.7] & 60.0 \\
Qwen3-VL-8B-T & 81.0 & 100.0 [100.0, 100.0] & 99.3 [98.3, 100.0] & 60.0 \\
Qwen3-VL-8B-I & 22.0 & 96.3 [94.0, 98.3] & 97.3 [95.3, 99.0] & 62.0 \\
InternVL3.5-8B & 30.0 & 89.0 [85.3, 92.3] & 93.0 [90.0, 95.7] & 65.0 \\
MiniCPM-V 4.6 & 20.3 & 22.7 [18.0, 27.3] & 23.3 [18.7, 28.0] & 0.0 \\
\bottomrule
\end{tabular}
}
\caption{\textbf{P3 control: the two cells ahead of the player are stated in words (outcome accuracy, \%).} 300 text items per cell. \emph{Two cells, no board}: the board is removed and only the contents of the two cells are given. \emph{Blocked predicted}: share of blocked answers in that setting (60\% of the items are blocked).}
\label{tab:p3ctrl}
\end{table}

\FloatFigPromptCtrl
\begin{table*}[t]
\centering
\footnotesize
\setlength{\tabcolsep}{5.0pt}
\fittab{%
\begin{tabular}{@{}l ccccc c ccccc@{}}
\toprule
 & \multicolumn{5}{c}{\Img} & & \multicolumn{5}{c}{\Txt} \\
\cmidrule(lr){2-6}\cmidrule(l){8-12}
Model & Solved & Unfinished & Deadlock & No push & No valid plan & & Solved & Unfinished & Deadlock & No push & No valid plan \\
\midrule
GPT-6.1 & 139 & 34 & 37 & 8 & 7 & & 149 & 29 & 22 & 10 & 15 \\
Gemini 3.8 Flash & 185 & 23 & 11 & 3 & 3 & & 180 & 23 & 15 & 2 & 5 \\
\midrule
Qwen3-VL-32B-Instruct & 0 & 0 & 0 & 0 & 225 & & 0 & 0 & 0 & 0 & 225 \\
Qwen3-VL-8B-Thinking & 3 & 68 & 46 & 85 & 23 & & 20 & 71 & 82 & 30 & 22 \\
Qwen3-VL-8B-Instruct & 2 & 45 & 57 & 117 & 4 & & 0 & 47 & 60 & 109 & 9 \\
InternVL3.5-8B & 0 & 15 & 11 & 27 & 172 & & 0 & 42 & 43 & 37 & 103 \\
MiniCPM-V 4.6 & 0 & 17 & 12 & 61 & 135 & & 0 & 16 & 5 & 200 & 4 \\
\bottomrule
\end{tabular}
}
\caption{\textbf{How single-turn plans end (number of levels out of 225).} The counts behind the outcome figure of the paper. \emph{Unfinished}: the moves run out first; \emph{Deadlock}: a push after which the solver proves the level unsolvable; \emph{No push}: no box is ever moved; \emph{No valid plan}: no parsable action sequence.}
\label{tab:outcomes}
\end{table*}
\begin{table*}[t]
\centering
\footnotesize
\setlength{\tabcolsep}{3.6pt}
\fittab{%
\begin{tabular}{@{}l ccccc c ccccc c cc@{}}
\toprule
 & \multicolumn{5}{c}{\Img: accuracy by true outcome} & & \multicolumn{5}{c}{\Txt: accuracy by true outcome} & & \multicolumn{2}{c}{Predicts blocked} \\
\cmidrule(lr){2-6}\cmidrule(lr){8-12}\cmidrule(l){14-15}
Model & Walk & Wall & Push & Box--wall & Box--box & & Walk & Wall & Push & Box--wall & Box--box & & \Img & \Txt \\
\midrule
GPT-6.1 & \best{100.0} & \best{100.0} & \best{100.0} & 96.3 & \best{100.0} & & 100.0 & 94.1 & 97.1 & \best{100.0} & \best{100.0} & & 57.7 & 57.7 \\
Gemini 3.8 Flash & \best{100.0} & \best{100.0} & \best{100.0} & \best{100.0} & \best{100.0} & & 100.0 & \best{100.0} & \best{100.0} & \best{100.0} & \best{100.0} & & 58.3 & 58.3 \\
\midrule
Qwen3-VL-32B-Instruct & \best{76.7} & 0.0 & 66.7 & 0.0 & 0.0 & & 73.3 & 5.0 & 65.0 & 0.0 & 0.0 & & 0.0 & 2.3 \\
Qwen3-VL-8B-Thinking & \best{76.7} & \best{56.7} & 41.7 & \best{35.0} & \best{40.0} & & 90.0 & \best{93.3} & 73.3 & \best{66.7} & \best{81.7} & & 36.3 & 53.0 \\
Qwen3-VL-8B-Instruct & 13.3 & 0.0 & \best{90.0} & 0.0 & 0.0 & & 13.3 & 0.0 & \best{96.7} & 0.0 & 0.0 & & 0.0 & 0.0 \\
InternVL3.5-8B & \second{26.7} & \second{21.7} & \second{70.0} & 0.0 & 0.0 & & 36.7 & \second{26.7} & 65.0 & \second{8.3} & 13.3 & & 14.3 & 21.0 \\
MiniCPM-V 4.6 & 18.3 & \second{21.7} & 58.3 & 0.0 & 0.0 & & 0.0 & 5.0 & \second{91.7} & 5.0 & 0.0 & & 19.3 & 10.3 \\
\bottomrule
\end{tabular}
}
\caption{\textbf{P3 accuracy (\%) by outcome type.} \emph{Wall}, \emph{Box--wall}, and \emph{Box--box} are the three blocked moves and make up 60\% of the items. \emph{Predicts blocked}: share of answers that name one of these three types. Qwen3-VL-8B-Thinking handles blocked moves from text (67--93\%) and loses 32--42 points on each of them from images. Proprietary models: 6$\times$6 and 8$\times$8 boards. \tablemarks}
\label{tab:p3type}
\end{table*}

Whether a move is a walk, a push, or a blocked move depends on two cells: the cell next to the player in the direction of the move, and the cell after it.
The control states what these two cells contain and asks the P3 question in two variants: without the board (\emph{two cells, no board}), and with the board (\emph{board + two cells}).
\Cref{fig:prompt-ctrl} gives the wording.

With the two cells named, \qL, \qI, and InternVL3.5-8B reach 98.3\%, 96.3\%, and 89.0\%, compared with 28.7\%, 22.0\%, and 30.0\% on the full board (\cref{tab:p3ctrl}).
They now predict a blocked move for 60--65\% of the items, and 60\% of the items are blocked.
These three models know the rules.
What they miss on a board is the step before the rule: finding the two cells from the position of the player and the direction of the move.
MiniCPM-V 4.6 stays at 22.7\% and predicts no blocked move at all, so for this model the rules themselves are missing.
Keeping the board next to the sentence gives the same accuracy as removing it (93.0--99.3\% for the first four models): once the two cells are known, the rest of the board does not affect the answer.

\subsection{All Diagnostic Quantities}
\label{sec:diag}

\Cref{tab:diag} lists every quantity used by the criteria of Sec.~3.4 of the paper with its interval, for both games, and \cref{fig:diag} plots the two checks.

\para{Rule check.} The models fall into two groups with a wide margin (\cref{fig:diag}a,b).
On Sokoban, the four rule-limited models have upper bounds between 23.0\% and 36.2\%, below the 40\% threshold, and the other three have lower bounds of at least 76.5\%.
On the maze, the upper bounds are 24.7--45.3\% against a threshold of 66.7\%, and the lower bounds of the other three are at least 83.0\%.
No model lies near a threshold, so the labels do not depend on its exact value.

\para{Perception check.} For \qT the image and text intervals are far apart, on P2 (17.5--29.6\% and 78.3--87.9\%) and on the interactive main task (3.1--8.9\% and 26.2--38.7\%; \cref{fig:diag}c).
For the proprietary models the paired visual cost on the interactive task is $+$4.0 [$-$2.2, 10.2] and $-$3.1 [$-$8.0, 1.8] points, and both intervals include zero.

\para{What a primary label means.} \qL and InternVL3.5-8B also have visual costs on P2 (51.6 and 18.8 points on Sokoban).
Both fail P3 from text, so better board reading would leave them at the rule stage.
Sec.~5.2 of the paper tests this case with \qI: \readset brings its localization to 100\%, and P3 stays at 20.7\%.

\subsection{P1 from Text: The Format of the Marked Cell}
\label{sec:p1fmt}

\begin{table}[t]
\centering
\footnotesize
\setlength{\tabcolsep}{3.0pt}
\fittab{%
\begin{tabular}{@{}l c c c c@{}}
\toprule
 & \multicolumn{2}{c}{P1 from \Txt} &  &  \\
\cmidrule(lr){2-3}
Model & Mark layer & Inline mark & Change & P1 from \Img \\
\midrule
Qwen3-VL-32B-I & 27.5 & 85.8 [81.2, 90.0] & \gup{$+$58.3} & 62.1 \\
Qwen3-VL-8B-T & 75.8 & 95.8 [93.3, 97.9] & \gup{$+$20.0} & 44.2 \\
Qwen3-VL-8B-I & 24.2 & 68.3 [62.5, 74.2] & \gup{$+$44.1} & 24.2 \\
InternVL3.5-8B & 26.7 & 78.3 [73.3, 82.9] & \gup{$+$51.6} & 20.0 \\
MiniCPM-V 4.6 & 19.2 & 43.8 [38.8, 48.3] & \gup{$+$24.6} & 45.4 \\
\bottomrule
\end{tabular}
}
\caption{\textbf{P1 with the marked cell written inside the text grid (accuracy, \%).} \emph{Mark layer}: the format of the benchmark, a second grid that marks one cell with \texttt{X}. \emph{Inline mark}: the marked cell is written as \texttt{[c]} in the board itself. Same 240 items. Gains are shown in red.}
\label{tab:p1fmt}
\end{table}

In Tab.~2 of the paper, two open-source models read the marked cell less accurately from text than from images in P1 (\qL: 27.5\% and 62.1\%; MiniCPM-V 4.6: 19.2\% and 45.4\%).
The text format of the mark explains this.
In \bench the marked cell is given in a second grid, so the model has to align two grids before it can read the cell.
The control writes the mark into the board itself, as brackets around the character of the cell, and keeps the 240 items unchanged (\cref{tab:p1fmt}; wording in \cref{fig:prompt-ctrl}).
With the inline mark, text recognition rises for all five models, by 20.0 to 58.3 points; \qL reaches 85.8\% and \qT 95.8\%.
The low text scores of P1 thus reflect the difficulty of aligning two grids.
\bench keeps the mark layer because it leaves the board text identical across all tasks.

\section{Failure Analysis}
\label{sec:failures}

\FloatFigReplay
\subsection{How Single-Turn Plans End}
\label{sec:outcomes}

\begin{table}[t]
\centering
\footnotesize
\setlength{\tabcolsep}{3.0pt}
\fittab{%
\begin{tabular}{@{}l l ccc ccc ccc c@{}}
\toprule
 & & \multicolumn{3}{c}{6$\times$6} & \multicolumn{3}{c}{8$\times$8} & \multicolumn{3}{c}{10$\times$10} & \\
\cmidrule(lr){3-5}\cmidrule(lr){6-8}\cmidrule(lr){9-11}
Model & Input & 1 & 2 & 3 & 1 & 2 & 3 & 1 & 2 & 3 & All \\
\midrule
GPT-6.1 & \Img & 22 & 15 & 15 & 22 & 19 & 13 & 15 & 12 & 6 & 139 \\
 & \Txt & 22 & 20 & 22 & 22 & 18 & 12 & 14 & 13 & 6 & 149 \\
\midrule
Qwen3-VL-8B-T & \Img & 3 & 0 & 0 & 0 & 0 & 0 & 0 & 0 & 0 & 3 \\
 & \Txt & 11 & 1 & 0 & 3 & 4 & 0 & 1 & 0 & 0 & 20 \\
\bottomrule
\end{tabular}
}
\caption{\textbf{Single-turn success by board size and number of boxes (solved levels out of 25 per cell).} Columns 1--3: number of boxes. The 37 image-input deadlocks of GPT-6.1 are 13 boxes pushed into a corner, 11 against a wall, 3 frozen box pairs, and 10 positions that only the solver proves dead (text input: 7, 10, 1, 4).}
\label{tab:grid}
\end{table}

\Cref{tab:outcomes} gives the counts behind Fig.~3 of the paper.
Three profiles appear.
The proprietary models return valid plans that mostly succeed, and their failures are split between unfinished plans and deadlocks.
\qT changes its failure type with the input: from images it never pushes a box in 85 levels (38\%); from text this number falls to 30, and deadlocks rise from 46 to 82 (36\%).
With a board that it can read, the model reaches the boxes and then fails at planning.
\qI ends in the same way with both inputs, with 117 and 109 levels in which no box is pushed.
\qL returns no valid plan for any level, because each response repeats a short cycle of actions until the token limit.

\para{The same level with two inputs.} \qT solves 18 levels from text that it fails from the image.
\Cref{fig:replay} replays four of them, the ones in which the image plan ends in a deadlock.
From text its plan pushes every box onto a goal.
From the image, in three of the four levels, its first push moves a box against a wall with no goal along it, within the first six moves.
In the other 14 levels the image plan pushes no box (8), runs out of moves (4), or contains no actions (2).

\Cref{tab:grid} breaks success down by board size and number of boxes.
For GPT-6.1, success falls with the number of boxes with both inputs: 59, 46, and 34 of 75 levels from images and 58, 51, and 40 from text.
More boxes mean longer plans with more pushes that cannot be undone, and the same trend with both inputs places the difficulty in planning.
Most of its deadlocks are simple ones: 24 of the 37 image deadlocks are a box pushed into a corner or against a wall.

\subsection{How Interactive Episodes End}
\label{sec:iaend}

\begin{table}[t]
\centering
\footnotesize
\setlength{\tabcolsep}{3.2pt}
\fittab{%
\begin{tabular}{@{}l l cccc c@{}}
\toprule
Model & Input & Solved & Deadlock & Budget & Turns & Moves/turn \\
\midrule
GPT-6.1 & \Img & 183 & 40 & \multicolumn{2}{c}{2} & -- \\
 & \Txt & 192 & 31 & \multicolumn{2}{c}{2} & -- \\
Gemini 3.8 Flash & \Img & 196 & 27 & \multicolumn{2}{c}{2} & -- \\
 & \Txt & 189 & 33 & \multicolumn{2}{c}{3} & -- \\
\midrule
Qwen3-VL-8B-T & \Img & 13 & 157 & 25 & 30 & 2.49 \\
 & \Txt & 73 & 135 & 4 & 13 & 2.29 \\
\quad + \readset & \Img & 46 & 155 & 1 & 23 & 1.87 \\
 & \Txt & 64 & 142 & 4 & 15 & 2.29 \\
\quad + \solveset & \Img & 12 & 158 & 28 & 27 & 2.58 \\
 & \Txt & 42 & 159 & 21 & 3 & 2.76 \\
\bottomrule
\end{tabular}
}
\caption{\textbf{How interactive episodes end ($K{=}3$, number of levels out of 225).} An episode ends when the level is solved, when the solver proves a deadlock, or when the move budget (\emph{Budget}) or the turn limit (\emph{Turns}) is reached; the two are reported together for the proprietary models. \emph{Moves/turn}: average number of moves the model commits per turn.}
\label{tab:iaend}
\end{table}

\FloatFigFtdata
\begin{table*}[t]
\centering
\footnotesize
\setlength{\tabcolsep}{4.2pt}
\fittab{%
\begin{tabular}{@{}l l l l c l@{}}
\toprule
Run & Start model & Trained modules & Training data & Epochs / steps & Checkpoint \\
\midrule
\readset & Thinking & vision encoder (LoRA) + projector & 20k board-reading items, image & 1 / 625 & last \\
\readset & Instruct & language model (LoRA) + projector & 20k board-reading items, image & 2 / 1,250 & step 1,000 \\
\ruleset & Instruct & language model (LoRA) & 20k single-step items, text & 2 / 1,250 & best val. \\
\readset{} + \ruleset & Instruct + \readset & language model (LoRA) & 8k single-step items, image & 1 / 250 & last \\
\midrule
\solveset & Thinking & language model (LoRA) & 8k levels with solutions (4k image, 4k text) & 1 / 250 & step 100 \\
\solveset, matched & Thinking & vision encoder (LoRA) + projector & 20k levels with solutions, image & 1 / 625 & last \\
\midrule
\readset & Kimi-VL-A3B-Thinking & vision encoder (LoRA) + projector & 20k board-reading items, image & 1 / 625 & last \\
\bottomrule
\end{tabular}
}
\caption{\textbf{Training configurations.} All runs use low-rank adapters of rank 16, learning rate $10^{-4}$, an effective batch size of 32, and bfloat16 weights, and are trained with ms-swift. For reasoning models the loss is computed on the answer only. \emph{Best val.}: highest joint accuracy on 1,000 validation items; step 1,000 and step 100 are the checkpoints with the highest validation accuracy and validation success rate. \emph{\solveset, matched}: the same modules, data size, and schedule as \readset, with solver solutions as the target. Thinking / Instruct: Qwen3-VL-8B.}
\label{tab:ftcfg}
\end{table*}

With the board shown again after every turn, little is left to track, and failures concentrate in deadlocks (\cref{tab:iaend}).
For the proprietary models, 131 of the 140 failed episodes end in a deadlock.
For \qT with image input, 55 episodes reach the move budget or the turn limit before training.
After \readset this number is 24, and 46 levels are solved, up from 13.
The number of deadlocks stays near 155, so the failures that remain after \readset are planning failures.
\solveset changes the behavior in the other direction: the model commits more moves per turn from text (2.76 compared with 2.29) and ends 159 text episodes in a deadlock, up from 135.

\subsection{P3: Outcome Types and Confusion Matrices}
\label{sec:p3conf}

\begin{figure}[t]
\centering
\includegraphics[width=\linewidth]{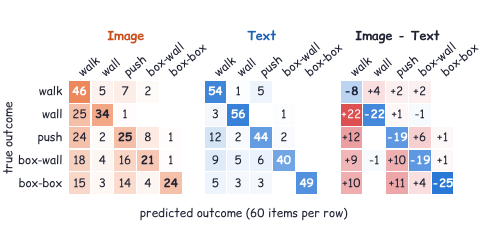}
\caption{\textbf{P3 confusion matrices of \qT.} Rows: true outcome (60 items each); columns: predicted outcome. Right: difference between the two inputs; the first column counts answers of a plain walk.}
\label{fig:conf}
\end{figure}

\Cref{tab:p3type} splits P3 accuracy by the true outcome, and \cref{fig:conf} shows the confusion matrices of \qT.

\para{\qT.} From text the diagonal dominates, with 243 of 300 answers correct.
From images the largest change is in the first column: 82 of the 240 items whose true outcome is a wall, a push, or a blocked box are answered as a plain walk, compared with 29 from text.
These 53 additional walk answers account for more than half of the 93 additional image errors.
The model reports a free cell ahead where a wall or a box stands.
A misread position produces exactly this error: if the player is placed one cell off, the cells ahead of it are different cells.

\para{Rule-limited models.} The table shows the response bias behind the near-chance scores of the four rule-limited models.
\qI answers \emph{push} for most items (90.0\% and 96.7\% correct on push items, 0\% on every blocked type), \qL answers \emph{walk} or \emph{push}, and InternVL3.5-8B answers \emph{push} for 190 of the 300 image items.
None of the four predicts a blocked move for more than 21.0\% of the items, although 60\% of the items are blocked.
For the two Qwen Instruct models the per-type accuracy is nearly the same from image and from text, which matches a failure in applying the rules.

\subsection{P4: Where Tracking Errors Start}
\label{sec:p4type}

\begin{figure}[t]
\centering
\includegraphics[width=\linewidth]{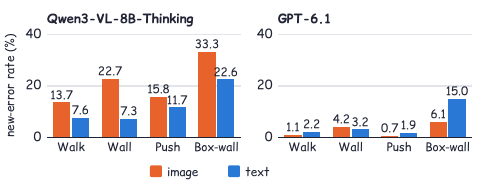}
\caption{\textbf{P4 new-error rate by type of the move.} Share of steps at which the reported position does not follow from the previous reported position.}
\label{fig:p4type}
\end{figure}

\Cref{fig:p4type} reports the new-error rate of P4 by the type of the move.
For \qT the rate is higher from images for every type, and the difference is largest for moves into a wall (22.7\% from images, 7.3\% from text).
A blocked move leaves the player in place, and recognizing it requires knowing exactly which cell lies ahead.
Moves into a box that is blocked by a wall show the highest rates with both inputs (33.3\% and 22.6\%).
GPT-6.1 stays below 5\% on walk, wall, and push moves with both inputs, and its remaining tracking errors are on box--wall moves (6.1\% from images, 15.0\% from text).

\section{Fine-Tuning Details}
\label{sec:ftdetails}

\subsection{Training Data}
\label{sec:ft-data}

All training boards are newly generated with the engine of \cref{sec:construction}, and a layout whose hash appears in the benchmark is discarded.

\para{\readset.} 20,000 items with image input of three kinds (\cref{fig:ftdata}a): \emph{locate}, which asks for the content and the position of a marked cell (60\%); \emph{look up}, which asks for the content of the cell at a given position (20\%); and \emph{transcribe}, which asks for the board as text (20\%).
Boards have 6 to 12 cells per side and are drawn with the flat and the textured sprite sets; grid lines and the ruler are each switched on in half of the items.
No item contains an action, a rule, or a plan, so \readset supervises reading the board and nothing after it.

\para{\ruleset.} 20,000 single-step items in the P3 format with text input and 1,000 validation items, plus 8,000 items with image input for the run that follows \readset.

\para{\solveset.} 8,000 levels with the reference solution of the solver as the target, half with image and half with text input.
The matched run of \cref{sec:ft-matched} uses 20,000 levels with image input.

\FloatFigIareplay
\subsection{Training Configurations}
\label{sec:ft-config}

All runs use ms-swift~\cite{msswift} with low-rank adapters~\cite{lora}; \cref{tab:ftcfg} lists the settings of every run, and \cref{fig:ftdata}b shows the modules that each run updates.
For \readset on the Thinking model, adapters are placed in the vision encoder, the projector is trained with it, and the language model receives no update.
Text inputs do not pass through either trained module, so text scores before and after \readset differ by sampling alone (\cref{tab:mazeft}).

\subsection{How \readset Changes the Later Stages}
\label{sec:ft-chain}

\begin{table}[t]
\centering
\footnotesize
\setlength{\tabcolsep}{3.0pt}
\fittab{%
\begin{tabular}{@{}l c c c@{}}
\toprule
Quantity (Qwen3-VL-8B-Thinking) & Base, \Img & + \readset, \Img & Base, \Txt \\
\midrule
\multicolumn{4}{@{}l}{\textit{P3 Single-step}} \\
Start position correct & 33.7 & 97.0 & 85.0 \\
Outcome correct & 50.0 & 89.3 & 81.0 \\
Start and outcome correct & 22.7 & 87.0 & 74.7 \\
Wrong outcomes with a wrong start & 78 & 6 & -- \\
\midrule
\multicolumn{4}{@{}l}{\textit{P4 Tracking}} \\
Final position correct & 43.1 & 67.9 & 60.4 \\
New error at the first move & 11.2 & 1.1 & 3.6 \\
New error per move, all moves & 16.9 & 12.8 & 11.0 \\
\bottomrule
\end{tabular}
}
\caption{\textbf{How \readset{} changes the stages that it does not train (\%).} Qwen3-VL-8B-Thinking. Paired gains from images: start position $+$63.3 [56.7, 70.1], outcome $+$39.3 [34.5, 44.7], P4 final position $+$24.8 [19.2, 30.4]. The first move of P4 depends only on reading the board.}
\label{tab:chain}
\end{table}

\begin{table}[t]
\centering
\footnotesize
\setlength{\tabcolsep}{3.2pt}
\fittab{%
\begin{tabular}{@{}l ccc c ccc@{}}
\toprule
 & \multicolumn{3}{c}{Sokoban, interactive} & & \multicolumn{3}{c}{Maze, single-turn} \\
\cmidrule(lr){2-4}\cmidrule(l){6-8}
Model, input & 6$\times$6 & 8$\times$8 & 10$\times$10 & & Short & Medium & Long \\
\midrule
Base, \Img & 12.0 & 5.3 & 0.0 & & 27 & 7 & 8 \\
+ \readset, \Img & 24.0 & 21.3 & 16.0 & & 91 & 60 & 51 \\
Base, \Txt & 38.7 & 38.7 & 20.0 & & -- & -- & -- \\
\bottomrule
\end{tabular}
}
\caption{\textbf{Gain from \readset{} by board size and by path length (success rate, \%; 75 levels per cell).} Qwen3-VL-8B-Thinking. Short, medium, and long: the three groups of shortest-path length of the maze.}
\label{tab:ftsize}
\end{table}

\begin{table}[t]
\centering
\footnotesize
\setlength{\tabcolsep}{4.0pt}
\fittab{%
\begin{tabular}{@{}l cc cc@{}}
\toprule
 & \multicolumn{2}{c}{\Img} & \multicolumn{2}{c}{\Txt} \\
\cmidrule(lr){2-3}\cmidrule(l){4-5}
Maze task & Base & + \readset & Base & + \readset \\
\midrule
P1 Recognition & 56.2 & 98.3 & 79.2 & 79.2 \\
P2 Localization & 27.1 & 95.4 & 77.5 & 78.3 \\
P2 Localization + ruler & 77.9 & 99.6 & 90.8 & 90.0 \\
P3 Start position & 38.7 & 99.7 & 89.3 & 89.7 \\
P3 Outcome & 71.0 & 95.7 & 88.0 & 90.3 \\
P4 Final position & 59.8 & 78.3 & 72.3 & 69.6 \\
Main task, single-turn & 13.8 & 67.1 & 48.9 & 52.0 \\
\bottomrule
\end{tabular}
}
\caption{\textbf{The maze before and after \readset{} (Qwen3-VL-8B-Thinking, \%).} The maze never appears in training. Text inputs do not pass through the trained modules, so the two text columns differ only by sampling.}
\label{tab:mazeft}
\end{table}

\para{Item-level attribution.} \Cref{tab:chain} gives the numbers behind Sec.~5.2 of the paper.
Before training, the model reads the start position correctly in 33.7\% of the P3 image items; after \readset it does so in 97.0\%, above the 85.0\% it reaches from text.
Outcome accuracy follows, from 50.0\% to 89.3\%.
Among the wrong outcomes, 78\% came with a wrong start position before training and 6\% after: the errors that remain are rule errors on a correctly read board.
P4 shows the same effect at the first move, whose result depends on reading the board alone: the new-error rate falls from 11.2\% to 1.1\% (3.6\% from text).

\para{By board size and path length.} \Cref{tab:ftsize} splits the gain on the main task.
Interactive success rises on every board size, including $10{\times}10$ boards, on which the base model solves no level from images (0.0\% to 16.0\%).
On the maze the gain holds for short, medium, and long paths (27\% to 91\%, 7\% to 60\%, and 8\% to 51\%).

\para{The same level before and after.} After \readset the model solves 37 levels that it did not solve before.
\Cref{fig:iareplay} replays four of them.
Before training the model moves for the whole episode and never pushes a box: in the $8{\times}8$ level with two boxes it makes 66 moves and visits 9 cells.
After training it walks to the boxes and solves the same levels in 20 to 40 moves.

\para{All maze tasks.} \Cref{tab:mazeft} reports every maze task before and after \readset.
The maze never appears in training, and all seven image scores rise: localization from 27.1\% to 95.4\%, and single-turn solving from 13.8\% to 67.1\%.
The text columns stay at the level of the base model; the 52.0\% quoted in Sec.~5.3 of the paper is the text score of the trained model.

\subsection{\solveset: Training on Solver Solutions}
\label{sec:ft-matched}

\begin{figure}[t]
\centering
\includegraphics[width=\linewidth]{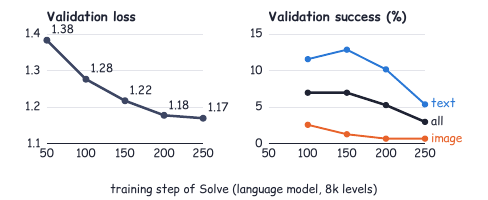}
\caption{\textbf{Training on \solveset.} Validation loss (left) and success rate on held-out validation levels (right) during training.}
\label{fig:solve}
\end{figure}

\begin{table}[t]
\centering
\footnotesize
\setlength{\tabcolsep}{3.2pt}
\fittab{%
\begin{tabular}{@{}l c c c c c@{}}
\toprule
 & \multicolumn{3}{c}{Sokoban} & & Maze \\
\cmidrule(lr){2-4}\cmidrule(l){6-6}
Training (\Img{} input) & P2 & Single-turn & Interactive & & Single-turn \\
\midrule
Base & 23.3 & 1.3 & 5.8 & & 13.8 \\
+ \readset{} (vision, 20k) & 90.8 & 10.2 & 20.4 & & 67.1 \\
+ \solveset{} (language model, 8k) & -- & 1.8 & 5.3 & & -- \\
+ \solveset{} (vision, 20k; matched) & 2.9 & 0.0 & -- & & 11.1 \\
\bottomrule
\end{tabular}
}
\caption{\textbf{Changing only the supervision target (Qwen3-VL-8B-Thinking, image input, \%).} The matched run keeps the modules, data size, and schedule of \readset{} and replaces the board-reading answers with solver solutions. --: not evaluated.}
\label{tab:controls}
\end{table}

\para{Training curve.} During training on \solveset, the validation loss falls steadily from 1.38 to 1.17, and the success rate on held-out validation levels falls from 7.0\% to 3.0\% (\cref{fig:solve}).
The model fits the solution strings better and solves fewer levels.
We evaluate the checkpoint with the highest validation success (step 100), the most favorable choice for \solveset.

\para{Matched budget.} In the paper, \solveset trains the language model on 8,000 levels, and \readset trains the vision modules on 20,000 items.
To remove these differences, we repeat \solveset with the modules, the data size, and the schedule of \readset: vision encoder and projector, 20,000 levels with image input, one epoch.
Only the supervision target changes, from board-reading answers to solver solutions.
With this budget, \readset brings localization from 23.3\% to 90.8\% and maze solving from 13.8\% to 67.1\% (\cref{tab:controls}).
The matched \solveset run ends at 2.9\% localization, 0.0\% on single-turn Sokoban, and 11.1\% on the maze, all at or below the base model.
The gain of \readset therefore comes from the supervision target, which the diagnosis identifies.

\FloatFigSprites
\subsection{Text-Twin Control}
\label{sec:ft-twin}

\begin{table}[t]
\centering
\footnotesize
\setlength{\tabcolsep}{6.0pt}
\fittab{%
\begin{tabular}{@{}l c c@{}}
\toprule
Qwen3-VL-8B-Instruct & P2 from \Img & P2 from \Txt \\
\midrule
Base & 11.2 & 24.6 \\
+ \readset{} (image items) & 100.0 & 25.8 \\
+ the same items as text & 20.8 & 100.0 \\
\bottomrule
\end{tabular}
}
\caption{\textbf{Text-twin control (localization accuracy, \%).} The same 20,000 questions are used for training with the board given as an image (\readset) or as text.}
\label{tab:twin}
\end{table}

A possible alternative explanation for the gain of \readset is that the model learns the question format and the coordinate convention, with no change in vision.
The text-twin control tests this explanation: \qI is trained on the same 20,000 questions with every board given as text (\cref{tab:twin}).
The text-trained model reaches 100.0\% localization from text and 20.8\% from images.
The image-trained model reaches 100.0\% from images and 25.8\% from text.
Each run improves the input it was trained on and leaves the other near the base model (11.2\% and 24.6\%).
The format and the convention are the same in the two runs, so the 79-point difference on images is the effect of training on images.

\FloatFigMazeds
\subsection{A Second Model Family}
\label{sec:ft-second}

\begin{table}[t]
\centering
\footnotesize
\setlength{\tabcolsep}{2.6pt}
\fittab{%
\begin{tabular}{@{}l c c c c@{}}
\toprule
 & Base, \Img & + \readset, \Img & Paired gain & Base, \Txt \\
\midrule
\multicolumn{5}{@{}l}{\textit{Kimi-VL-A3B-Thinking}} \\
P1 Recognition & 45.4 & 99.6 & \gup{$+$54.2 [48.8, 60.0]} & 48.8 \\
P2 Localization & 16.7 & 92.1 & \gup{$+$75.4 [69.6, 80.8]} & 51.2 \\
P3 Start position & 27.7 & 72.3 & \gup{$+$44.7 [36.8, 52.6]} & 49.0 \\
P3 Outcome & 27.7 & 47.3 & \gup{$+$19.7 [14.0, 25.0]} & 60.3 \\
Interactive main task & 0.9 & 1.8 & \gup{$+$0.9 [$-$1.3, 3.1]} & 3.1 \\
\bottomrule
\end{tabular}
}
\caption{\textbf{A second model family: \readset{} on Kimi-VL-A3B-Thinking (\%).} Screening values of the base model: P3 text 60.3 [54.5, 66.0]; interactive visual cost $+$2.2 [$-$0.4, 4.9]. GLM-4.1V-9B-Thinking, screened in the same way: P2 image 40.8 [33.8, 48.3], text 35.8 [30.0, 42.1]; P3 text 62.3 [57.4, 67.2]. Gains are shown in red.}
\label{tab:second}
\end{table}

The fine-tuning experiments of the paper use Qwen3-VL-8B.
To test the criteria on other families, we screen two further open-source reasoning models with the probes and the interactive main task: Kimi-VL-A3B-Thinking~\cite{kimivl} and GLM-4.1V-9B-Thinking~\cite{glm41v}.

\para{Prediction.} Both models pass the rule check (P3 text 60.3\% and 62.3\%, with lower bounds above 40\%).
GLM-4.1V-9B-Thinking shows no visual cost on P2 (image 40.8\%, text 35.8\%), so the perception criterion does not apply to it.
Kimi-VL has a large visual cost on P2 (16.7\% from images, 51.2\% from text).
On the interactive main task, its visual cost is 2.2 points with an interval that includes zero, because it solves 3.1\% of the levels even from text.
The second condition of the perception criterion is thus unmet: for this model, planning limits solving before perception does.
The criteria therefore predict that \readset will improve its board reading and leave its solving unchanged.

\para{Result.} \Cref{tab:second} confirms both parts.
\readset, with the configuration used for \qT, raises the localization of Kimi-VL from 16.7\% to 92.1\% and its recognition from 45.4\% to 99.6\%, so the training procedure carries over to a different vision encoder and language model.
The gain reaches P3: start-position accuracy rises from 27.7\% to 72.3\% and outcome accuracy from 27.7\% to 47.3\%.
Interactive success moves from 0.9\% to 1.8\%, a change whose interval includes zero, and stays below the 3.1\% of text input.
Together with \qI in the paper, this is the second case in which the criteria correctly predict that board-reading data will leave solving unchanged.

\section{Transfer and Scope of \readset}
\label{sec:transfer}

\subsection{Sprite Sets Never Seen in Training}
\label{sec:sprites}

\begin{table}[t]
\centering
\footnotesize
\setlength{\tabcolsep}{3.0pt}
\fittab{%
\begin{tabular}{@{}l l c c c@{}}
\toprule
Sprite set & Boards & Base & + \readset & Paired gain \\
\midrule
Textured (training) & Sokoban & 23.3 & 90.8 & -- \\
\midrule
Garden & Sokoban & 32.1 & 86.2 & \gup{$+$54.2 [47.1, 61.3]} \\
Blue brick & Sokoban & 22.1 & 82.1 & \gup{$+$60.0 [52.9, 66.7]} \\
Meadow & Sokoban & 27.5 & 92.1 & \gup{$+$64.6 [57.5, 71.2]} \\
Meadow & third-party maze & 17.5 & 87.5 & \gup{$+$70.0 [60.0, 79.2]} \\
\bottomrule
\end{tabular}
}
\caption{\textbf{Image localization (P2, \%) on sprite sets that never appear in training.} Qwen3-VL-8B-Thinking, no ruler. \readset{} is trained on the flat and textured sprite sets only. Third-party maze: \cref{sec:mazeds}. Gains are shown in red.}
\label{tab:sprites}
\end{table}

\readset is trained on boards drawn with two sprite sets, flat and textured.
To test whether the trained skill depends on the look of the tiles, we draw the P2 boards with three new sprite sets, Garden, Blue brick, and Meadow, in which walls, floor, goals, boxes, and the player all look different (\cref{fig:sprites}a).
The legend sentence of the prompt describes the new sprites, and nothing else changes.

Image localization after \readset reaches 86.2\%, 82.1\%, and 92.1\% on the three sets, compared with 90.8\% on the training sprites, and every paired gain is at least 54 points (\cref{tab:sprites}).
The last row of the table repeats the test on the third-party mazes of \cref{sec:mazeds}.
Counting rows and columns is independent of what the tiles look like, and this is the skill that \readset trains.

\subsection{Scope: Another Renderer and General Benchmarks}
\begin{table*}[t]
\centering
\footnotesize
\setlength{\tabcolsep}{4.0pt}
\fittab{%
\begin{tabular}{@{}l c ccc c cc l@{}}
\toprule
 & P3 text & \multicolumn{3}{c}{P2 Localization} & $\dvis$ P2 & \multicolumn{2}{c}{Single-turn} & \\
\cmidrule(lr){3-5}\cmidrule(lr){7-8}
Model & [95\% CI] & \Img & \textcolor{imgc}{+ ruler} & \Txt & [95\% CI] & \Img & \Txt & Bottleneck \\
\midrule
Qwen3-VL-32B-Instruct & 36.0 [28.7, 43.3] & 58.3 & 100.0 & 82.5 & $+$24.2 [12.5, 35.8] & -- & -- & Rules \\
Qwen3-VL-8B-Thinking & 91.3 [86.7, 95.3] & 17.5 & 86.7 & 87.5 & $+$70.0 [60.8, 78.3] & 8.0 & 64.7 & Perception \\
Qwen3-VL-8B-Instruct & 34.0 [26.7, 41.3] & 14.2 & 95.0 & 19.2 & $+$5.0 [$-$4.2, 14.2] & -- & -- & Rules \\
InternVL3.5-8B & 35.3 [28.0, 43.3] & 0.0 & 71.7 & 20.0 & $+$20.0 [13.3, 26.7] & -- & -- & Rules \\
MiniCPM-V 4.6 & 25.3 [18.7, 32.7] & 2.5 & 15.0 & 3.3 & $+$0.8 [$-$3.3, 5.0] & -- & -- & Rules \\
\bottomrule
\end{tabular}
}
\caption{\textbf{The diagnosis on third-party mazes (\%).} 150 mazes of 9$\times$9 cells generated with the \texttt{maze-dataset} library and drawn with a new sprite set (Meadow). P3: 150 text items, rule threshold 66.7\%; P2: 120 items; single-turn main task on all 150 mazes, run for the models that pass the rule check.}
\label{tab:mzddiag}
\end{table*}

\FloatFigMzcases
\label{sec:scope}

\begin{table}[t]
\centering
\footnotesize
\setlength{\tabcolsep}{5.0pt}
\fittab{%
\begin{tabular}{@{}l l c c@{}}
\toprule
Benchmark & Input & Base & + \readset \\
\midrule
VSP maze, cell perception & \Img & 73.3 & 40.0 \\
 & \Txt & 91.2 & 93.5 \\
VSP maze, main task & \Img & 49.3 & 17.5 \\
 & \Txt & 98.0 & 98.5 \\
\midrule
GSM8K & \Txt & 96.0 & 96.7 \\
MMStar & \Img & 73.2 & 56.0 \\
\bottomrule
\end{tabular}
}
\caption{\textbf{Scope of \readset{} outside \bench{} (\%).} Qwen3-VL-8B-Thinking. VSP mazes (600 items per row) are drawn by a different renderer. Text inputs do not pass through the trained modules.}
\label{tab:scope}
\end{table}

\FloatFigBlocks
\begin{table*}[t]
\centering
\footnotesize
\setlength{\tabcolsep}{4.0pt}
\fittab{%
\begin{tabular}{@{}l cc cc cc cc cc c@{}}
\toprule
 & \multicolumn{2}{c}{P1 Read cell} & \multicolumn{2}{c}{P2 Locate block} & \multicolumn{2}{c}{P3 One move} & \multicolumn{2}{c}{P4 Several moves} & \multicolumn{2}{c}{Main task} & $\dvis$ main \\
\cmidrule(lr){2-3}\cmidrule(lr){4-5}\cmidrule(lr){6-7}\cmidrule(lr){8-9}\cmidrule(lr){10-11}
Model & \Img & \Txt & \Img & \Txt & \Img & \Txt & \Img & \Txt & \Img & \Txt & [95\% CI] \\
\midrule
GPT-6.1 & 96.7 & \best{100.0} & 96.7 & \best{100.0} & 98.0 & 99.0 & 90.7 & 96.7 & 82.0 & 91.5 & $+$9.5 [3.5, 15.5] \\
Gemini 3.8 Flash & \best{99.3} & \best{100.0} & \best{100.0} & \best{100.0} & \best{100.0} & \best{100.0} & \best{100.0} & \best{100.0} & \best{100.0} & \best{99.5} & $-$0.5 [$-$1.5, 0.0] \\
\midrule
Qwen3-VL-32B-Instruct & \best{95.3} & \best{100.0} & \best{96.7} & \best{100.0} & \best{93.0} & \best{99.0} & \best{93.3} & \second{94.0} & \best{63.5} & \best{69.0} & $+$5.5 [$-$2.0, 13.0] \\
Qwen3-VL-8B-Thinking & 87.3 & \best{100.0} & 85.3 & \best{100.0} & 83.0 & \second{98.0} & \second{76.7} & \best{96.0} & \second{32.0} & \second{52.5} & $+$20.5 [14.0, 27.5] \\
Qwen3-VL-8B-Instruct & \second{91.3} & \best{100.0} & 89.3 & \best{100.0} & \second{84.0} & 88.0 & 68.0 & 75.3 & 18.0 & 29.0 & $+$11.0 [5.5, 16.5] \\
InternVL3.5-8B & 79.3 & \second{98.7} & \second{90.0} & \second{98.7} & 82.0 & 87.0 & 60.0 & 59.3 & 6.0 & 5.5 & $-$0.5 [$-$4.5, 4.0] \\
MiniCPM-V 4.6 & 61.3 & 12.0 & 50.7 & 56.0 & 18.0 & 8.0 & 5.3 & 12.7 & 4.5 & 0.0 & $-$4.5 [$-$7.5, $-$2.0] \\
\bottomrule
\end{tabular}
}
\caption{\textbf{The paired protocol on the photo-realistic Blocks World images of VSP (\%).} 750 items per model, each asked with the image and with the equivalent text state. P1: color of the block at a given stack and level (150 items); P2: stack and level of a given block (150); P3: legality of one move (100); P4: legality of a sequence of 3, 5, or 7 moves (150); main task: shortest plan of 1, 3, 5, or 7 moves, replayed by the official evaluator (200). $\dvis$: visual cost (text $-$ image) with its paired 95\% interval. \tablemarks}
\label{tab:blocks}
\end{table*}

\Cref{tab:scope} reports the maze task of VSP~\cite{vsp2025} and two general benchmarks, as announced in Sec.~5.3 of the paper.

\para{The base model shows the same diagnosis on VSP.} VSP mazes come from a different renderer and a different benchmark.
The base \qT solves 98.0\% of them from text and 49.3\% from images, and it reads single cells with 91.2\% accuracy from text and 73.3\% from images.
The visual cost that \bench measures for this model is therefore present on an independent benchmark.

\para{Scope of \readset.} \readset is a diagnostic intervention: 20,000 items from one renderer, trained for one epoch, to test whether board reading is the bottleneck.
Its effect is specific to that rendering family.
Scores that do not pass through the trained modules are unchanged: GSM8K~\cite{gsm8k}, a text-only benchmark, stays at 96.0\% and 96.7\%, and text input on VSP stays at 98.0\% and 98.5\%.
Scores that depend on the trained vision modules outside the training family fall, from 73.2\% to 56.0\% on MMStar~\cite{mmstar} and from 49.3\% to 17.5\% on VSP images.
Together with \cref{sec:sprites,sec:mazeds}, this marks the range in which the trained skill holds: new sprites, a new game, and mazes from another generator, all within the grid rendering of \bench.

\section{Third-Party Mazes}
\label{sec:mazeds}

The mazes of the paper come from our own generator.
To test the diagnosis and the transfer on boards that we did not design, we generate 150 mazes with the \texttt{maze-dataset} library~\cite{mazedataset}, a public generator built for studying maze-solving models.
The mazes have $9{\times}9$ cells and are carved by randomized depth-first search.
They consist of corridors one cell wide with dead ends and a single route to the exit, a structure that our random-wall mazes do not have (\cref{fig:mazeds}a).
We render them in two ways: with a sprite set that no model has seen (Meadow), and with the textured sprites of the paper.

\para{The diagnosis is the same.} \Cref{tab:mzddiag} reports P2, P3, and the single-turn main task on the Meadow mazes.
The rule check gives the same split as in the paper: the four models that fail it on our mazes score 25.3--36.0\% on P3 from text, with upper bounds of at most 43.3\% against the threshold of 66.7\%, and \qT passes with 91.3\%.
\qT again loses accuracy at localization (17.5\% from images, 87.5\% from text), the ruler restores it (86.7\%), and its visual cost on the main task is 56.7 points [47.3, 65.3].
All five open-source models are assigned the bottleneck that they have in Tab.~2 of the paper.

\para{Real answers.} \Cref{fig:mzcases} shows items with the answers of every model.
In the first P2 item all five models report a wrong position from the image, four of them report the correct one once the ruler is drawn, and \qT after \readset reports it without the ruler.
In the first P3 item the text board gives a wall to the right of the player; the four models that fail the rule check answer that the player walks, and \qT answers correctly.

\begin{table}[t]
\centering
\footnotesize
\setlength{\tabcolsep}{2.8pt}
\fittab{%
\begin{tabular}{@{}l c c c c@{}}
\toprule
Sprites & Base, \Img & + \readset, \Img & Paired gain & Base, \Txt \\
\midrule
Textured (paper) & 10.7 & 64.0 & \gup{$+$53.3 [45.3, 61.3]} & \multirow{2}{*}{64.7} \\
Meadow (new) & 8.0 & 16.7 & \gup{$+$8.7 [3.3, 14.7]} &  \\
\midrule
\multicolumn{5}{@{}l}{\textit{Textured sprites, by path length}} \\
\quad $\le$ 8 moves (74) & 21.6 & 85.1 & \gup{$+$63.5} & 73.0 \\
\quad 10--16 moves (53) & 0.0 & 54.7 & \gup{$+$54.7} & 52.8 \\
\quad $\ge$ 18 moves (23) & 0.0 & 17.4 & \gup{$+$17.4} & 65.2 \\
\bottomrule
\end{tabular}
}
\caption{\textbf{\readset{} on third-party mazes (single-turn success, \%).} Qwen3-VL-8B-Thinking on the same 150 mazes under two sprite sets. Number of mazes per path-length group in parentheses. Gains are shown in red.}
\label{tab:mzdread}
\end{table}

\FloatFigBlockcases

\para{\readset transfers to these mazes.} With the sprites of the paper, \readset raises single-turn success from 10.7\% to 64.0\%, a paired gain of 53.3 points [45.3, 61.3] that brings image input to the level of text input (64.7\%; \cref{tab:mzdread}).
The trained model has never seen a maze or a corridor layout.
Before training its plans walk into a wall after a few moves, and after training they follow the corridor to the exit (\cref{fig:mzcases}c).
In the three mazes shown, the plan from the image hits a wall at the first or second move, and the plan after \readset is identical to the plan from text, up to a corridor of 28 moves.
The gain holds for every path length: from 21.6\% to 85.1\% for paths of up to 8 moves, from 0.0\% to 54.7\% for 10 to 16 moves, and from 0.0\% to 17.4\% for 18 moves or more.

\para{New sprites and a new generator together.} With the Meadow sprites, localization transfers in full, from 17.5\% to 87.5\%, the level of text input.
Single-turn success rises from 8.0\% to 16.7\%, a significant gain of 8.7 points [3.3, 14.7].
The gain on the main task is smaller with these sprites because solving also requires telling wall tiles from floor tiles along the whole route.
This is a recognition step, and \readset was trained on two sprite sets.

\begin{figure}[t]
\centering
\includegraphics[width=\linewidth]{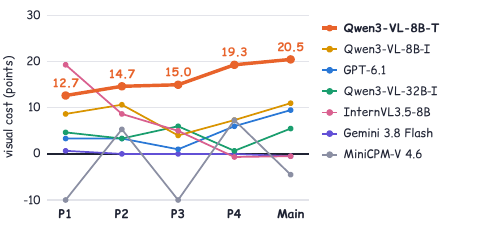}
\caption{\textbf{Visual cost (text $-$ image) along the chain on VSP Blocks World.} Values are printed for \qT. MiniCPM-V 4.6 is off the scale on P1 ($-$49.3), where it scores 12\% from text.}
\label{fig:blocks}
\end{figure}
\section{Photo-Realistic Images from VSP}
\label{sec:photos}
\FloatFigGalleries

\bench uses boards drawn from sprites.
To check that the paired, stage-wise protocol also applies beyond such boards, we build the same chain of stages on the photo-realistic Blocks World scenario of VSP~\cite{vsp2025}, in which each state is an image of colored blocks arranged in stacks.
Each state also has an exact text description, the blocks of every stack from bottom to top, which serves as the paired text input.

\para{Stages.} Five stages mirror the chain of the paper, and \cref{fig:blockcards} shows one real item for each.
P1 asks for the color of the block at a given stack and level (150 items).
P2 names a block and asks for its stack and level (150).
P3 asks whether one move is legal (100).
P4 asks whether a sequence of 3, 5, or 7 moves is legal (150).
The main task asks for the shortest plan of 1, 3, 5, or 7 moves from a start state to a goal state, and the official evaluator replays the answer (200).
Each of the 750 items is asked twice, with the image and with the text state, for all seven models.

\para{Results.} For \qT the visual cost is significant at every stage and grows along the chain: 12.7 points on P1, 14.7 on P2, 15.0 on P3, 19.3 on P4, and 20.5 [14.0, 27.5] on the main task (\cref{tab:blocks,fig:blocks}).
From text the model answers 96--100\% of the probe items correctly, so the loss comes from reading the image, and it accumulates as the task requires more of the scene to be read correctly.
Gemini 3.8 Flash shows no visual cost at any stage (at most 0.7 points).
GPT-6.1 also shows a visual cost on the main task (9.5 points [3.5, 15.5]), which is not significant on Sokoban: the protocol measures the cost separately for each kind of image.
The protocol thus carries over to photo-realistic images with its three elements unchanged: an equivalent text state, one probe per stage, and paired comparison on the same items.

\para{Items.} \Cref{fig:blockcases} shows two real items per stage with the result of every model.
The items in panel (b) are ones that \qT solves from text and misses from the image.
On these five items, the six models other than MiniCPM-V 4.6 answer all 30 text questions correctly and 15 of the 30 image questions, and Gemini 3.8 Flash answers all five from the image.

\para{Rules without grid lookup.} The P3 column of \cref{tab:blocks} agrees with the control of \cref{sec:p3ctrl}.
\qL, \qI, and InternVL3.5-8B judge single moves correctly from the Blocks World text state in 99.0\%, 88.0\%, and 87.0\% of the items.
These are the three models that reach 89.0--98.3\% on Sokoban once the two cells ahead are named.
The text state of Blocks World lists every stack from bottom to top, so the blocks that a move involves are given directly and no cell has to be located in a grid.
MiniCPM-V 4.6 stays low in both tests (8.0\% here, 22.7\% in the control).

\section{Gallery}
\label{sec:gallery}

\Cref{fig:gallery-soko} shows 18 Sokoban levels of the main task, two for each combination of board size and number of boxes.
\Cref{fig:gallery-maze} shows mazes of the paper with their shortest paths, third-party mazes in two sprite sets, and two mazes under four renderings each.

\end{document}